\documentclass{article}

\usepackage{jmlr2arxiv}

\usepackage{amsmath}

\usepackage{amsfonts}
\usepackage{subfigure}
\usepackage[boxed,noend]{algorithm2e}
\usepackage{xspace}
\usepackage{ifthen}
\usepackage{microtype}
\usepackage{times}
\usepackage{color}

\usepackage{bbm}
\usepackage{stmaryrd}
\usepackage{varioref}

\usepackage{xtab}

\usepackage[linkbordercolor={0 1 0},pdftex]{hyperref}

\jmlrheading{Volume??}{2012}{pages??}{??/??}{??/??}{Daniel Golovin
  and Andreas Krause}

\newcommand{\ourtitle}{\Term Submodularity: Theory and Applications in
  Active Learning and Stochastic Optimization}
\newcommand{\condensedtitle}{\Term Submodularity: Theory and Applications}

\ShortHeadings{\condensedtitle}{Golovin and Krause}
\firstpageno{1}

\newboolean{istechrpt}
\setboolean{istechrpt}{true}  %
\newboolean{isArxiv}          %
\setboolean{isArxiv}{true}   %
\newcommand{\TechReportOnly}[1]{\ifthenelse{\boolean{istechrpt}}{#1}{}}
\newcommand{\COLTonly}[1]{\ifthenelse{\boolean{istechrpt}}{}{#1}}
\newcommand{\ArxivOnly}[1]{\ifthenelse{\boolean{isArxiv}}{#1}{}}
\newboolean{showcomments}
\setboolean{showcomments}{false}
\newcommand{\term}[0]{adaptive\xspace}  %
\newcommand{\Term}[0]{Adaptive\xspace} %

\newcommand{\certifying}{self--certifying\xspace}
\newcommand{\Certifying}{Self--Certifying\xspace}

\newcommand{\probname}[0]{Stochastic Submodular Maximization} %

\newcommand{\AppendixA}[0]{The Appendix\xspace}
\newcommand{\appendixA}[0]{the Appendix\xspace}
\newcommand{\substitute}[3]{ \ensuremath{{#1}\!\left[{#2}/{#3}\right]} }

\newcommand{\averaged}[1]{\ensuremath{#1_{\text{avg}}}} %
\newcommand{\entropy}[1]{\ensuremath{\mathbb{H}\paren{#1}}}
\newcommand{\infogain}[2]{\ensuremath{\mathbb{I}\paren{{#1} ; {#2}}}}
\newcommand{\support}{\op{support}}
\newcommand{\perm}{\operatorname{perm}}   %
  
\newcommand{\indicator}[1]{\ensuremath{\one_{#1}}}

\newcommand {\costminsum}[1]{\ensuremath{c_{\Sigma}\!\paren{#1}}}
\newcommand {\costminsumsym}[0]{\ensuremath{c_{\Sigma}}}
\newcommand{\avgf}[0]{\averaged{f}} %
\newcommand {\util}[2]{\ensuremath{u\!\paren{{#1}, {#2}}}}

\newcommand{\cavg}[1]{{c_{\text{avg}}{(#1)}}}
\newcommand{\acst}[1]{{c_{\text{avg}}{(#1)}}}
\newcommand{\acstsym}[0]{{c_{\text{avg}}}}

\newcommand{\wcst}[1]{{c_{\text{wc}}{(#1)}}}
\newcommand{\awstsym}[0]{{c_{\text{wc}}}}
\newcommand{\condcost}[2]{\ensuremath{c\paren{ {#2} \! \mid \! {#1} }  }}
\newcommand{\diff}[2]{\ensuremath{\Delta \hspace{-0.3mm}\paren{ {#2}\! \mid \! {#1} }  }
}  
\newcommand{\sequence}{\sigma}

\newcommand{\policy}[0]{\ensuremath{\pi}}
\newcommand{\greedypolicy}[0]{\ensuremath{\pi^{\text{greedy}}}}
\newcommand{\piavg}{\policy_{\text{avg}}^*}
\newcommand{\policycover}{\policy_{\text{avg}}^*}

\newcommand{\strictprune}[2]{\ensuremath{  #1_{[\gets{#2}] }}} %
\newcommand{\laxprune}[2]{\ensuremath{  #1_{[{#2}\to]}}} %
\newcommand{\prune}[2]{\ensuremath{  #1_{[{#2}]} }  } %
\newcommand{\append}[2]{\ensuremath{{#1}@{#2}}} %

\newcommand{\treeSmush}[3]{\ensuremath{ #1_{[#2] \cup \set{#3}}   }} %
\newcommand{\node}[0]{\ensuremath{u}}
\newcommand{\elem}[0]{\ensuremath{e}}
\newcommand {\groundset}{\ensuremath{E}} %
\newcommand {\groundsubset}{\ensuremath{A}}%
\newcommand{\outcome}[0]{\ensuremath{o}} %
\newcommand{\outcomes}[0]{\ensuremath{O}} %
\newcommand{\mass}[1]{{\color{black}{\ensuremath{p\paren{#1}}}}}
\newcommand{\rlzmass}[1]{{\color{black}{\ensuremath{p\paren{#1}}}}}
\newcommand{\rlzmasssym}[0]{{\color{black}{\ensuremath{p}}}}
\newcommand{\rlzmassover}[2]{{\color{black}{\ensuremath{p_{#1}\!\paren{#2}}}}}
\newcommand{\rlzprior}[0]{{\color{black}{\ensuremath{p\paren{\rlz}}}}}
\newcommand{\rlz}[0]{{\color{black}{\ensuremath{\phi}}}} %
\newcommand{\rvrlz}[0]{{\color{black}{\ensuremath{\Phi}}}} %
\newcommand{\rvprlz}[0]{{\color{black}{\ensuremath{\Psi}}}} %
\newcommand{\prlz}[0]{{\color{black}{\ensuremath{\psi}}}} %

\newcommand{\prlzsub}[2]{\substitute{\prlz}{#1}{#2}}

\newcommand{\prlzxo}[0]{\prlzsub{x}{\outcome}}
\newcommand{\vselim}[0]{\ensuremath{\vs_{\text{elim}}}}

\newcommand{\prlzset}[0]{\mathcal{P}}

\newcommand{\distrib}[0]{\ensuremath{\mathcal{D}}}

\newcommand{\quota}[0]{\ensuremath{Q}} %
\newcommand{\budget}[0]{\ensuremath{k}} %
\newcommand{\price}[0]{\ensuremath{\theta}} %
\newcommand{\charge}[1]{\ensuremath{C_{#1}}} %

\newcommand{\progress}[1]{\ensuremath{  \expct{f(\dom({#1}), \rvrlz) \mid
  \rvrlz \sim {#1}}  }}

\newcommand{\hypotheses}[0]{\ensuremath{H}} 
\newcommand{\data}[0]{\ensuremath{X}} 
\newcommand{\labels}[0]{\ensuremath{L}} 
\newcommand{\target}[0]{\ensuremath{h^*}} 
\newcommand{\error}[0]{\operatorname{error}} 
\newcommand{\prior}[0]{\ensuremath{p_{\hypotheses}}}
\newcommand{\vs}[0]{\ensuremath{V}\xspace} %
 \newcommand{\gbs}[0]{GBS\xspace} %
\newcommand{\played}[2]{\groundset({#1}, {#2})}

\newcommand{\range}[0]{\operatorname{range}}

\newcommand{\prt}[2]{{#1}\!\paren{{#2}}}

\newcommand{\alive}[1]{{A({#1})}}

\newcommand{\numsensors}{\ensuremath{n}}

\newcommand{\sensor}{\ensuremath{v}}
\newcommand{\sensors}{V}

\newcommand{\pfail}[1]{p_{\text{fail}}\!\paren{#1} }

\newcommand{\numberof}[0]{\operatorname{rep}}
\newcommand{\invnumberof}[0]{\operatorname{rep}^{-1}}
\newcommand{\nmaps}{m}
\newcommand{\ntreasure}{t}
\newcommand{\nmatrix}{n}
\newcommand{\mapsize}{s}

\newcommand{\trace}{\tau}
\newcommand{\ptrace}[2]{\prlz\paren{{#1}, {#2}}}

\newcommand{\Xvec}{\mathbf{X}}

\newcommand{\bound}{\beta}
\newcommand{\avgbound}{\beta_{\text{avg}}}
\newcommand{\daniel}[1]{\ifthenelse{\boolean{showcomments}}{\textcolor{red}{Daniel: #1}}{}}
\newcommand{\andreas}[1]{\ifthenelse{\boolean{showcomments}}{\textcolor{red}{Andreas: #1}}{}}

\newcommand{\cf}{\emph{c.f.}\xspace}

\newcommand{\commentout}[1]{}

\newcommand{\collection}{\ensuremath{\mathcal{C}}}

\newcommand{\cS}{{\mathcal{S}}}
\newcommand{\cA}{{\mathcal{A}}}

\newcommand{\cX}{{\mathcal{X}}}
\newcommand{\cY}{{\mathcal{Y}}}

\newcommand{\cO}{{\mathcal{O}}}

\newcommand{\bx}{{\mathbf{x}}}
\newcommand{\by}{{\mathbf{y}}}

\newlength{\presec}
\newlength{\postsec}
\newlength{\presubsec}
\newlength{\postsubsec}
\newlength{\prepara}
\newlength{\postpara}
\newcommand{\initOneLiners}{%
    \setlength{\itemsep}{0pt}
    \setlength{\parsep }{0pt}
    \setlength{\topsep }{0pt}
}
\newenvironment{OneLiners}[1][\ensuremath{\bullet}]
    {\begin{list}
        {#1}
        {\initOneLiners}}
    {\end{list}}

\newenvironment{proofof}[1]{
\noindent{\bf Proof of {#1}:}}
{\hfill$\blacksquare$
}

\newcommand{\ignore}[1]{}

\def \etal {et al.\ }
\def \argmax {\mathop{\rm arg\,max}}
\def \argmin {\mathop{\rm arg\,min}}

\newcommand{\ith}{\ensuremath{i^{\mathrm{th}}}\xspace}

\newcommand{\op}[1]{\operatorname{#1}}
\newcommand{\one}{\mathbf{1}}

\newcommand{\paren} [1] {\ensuremath{ \left( {#1} \right) }}

\renewcommand{\Pr}[1]{\ensuremath{\mathbb{P}\left[#1\right] }}
\newcommand{\prob}[1]{\ensuremath{\mathbb{P}\left[#1\right] }}

\newcommand{\probover}[2]{\ensuremath{\mathbb{P}_{#1}\left[#2\right]
  }}
\newcommand{\proboverrlz}[2]{\ensuremath{\mathbb{P}\left[#2\right] }}

\newcommand{\floor}[1]{\ensuremath{\left\lfloor#1\right\rfloor}}
\newcommand{\tuple}[1]{\ensuremath{\langle #1 \rangle}}

\renewcommand{\implies}[0]{\ensuremath{\Rightarrow}}

\newcommand{\integers}{\ensuremath{\mathbb{Z}}}
\newcommand{\nats}{\ensuremath{\mathbb{N}}}
\newcommand{\reals}{\ensuremath{\mathbb{R}}}

\newcommand{\class} [1] {\textrm{#1}} %
\renewcommand{\P} {\class{P}}
\newcommand{\NP} {\class{NP}}

\renewcommand{\c}{c}

\newcommand{\OPT}{\textsf{OPT}}
\newcommand{\NULL}{\textsc{null}}

\newcommand{\obj}{\ensuremath{f}} %

\newcommand{\omitproof}[1]{}

\newcommand{\Alg}{\ensuremath{\mathcal{A}}}

\newcommand{\NonNegativeReals}{\ensuremath{\mathbb{R}_{\ge 0}}}
\newcommand{\NonNegativeIntegers}{\ensuremath{\mathbb{Z}_{\ge 0}}}

\DeclareMathOperator{\dom}{dom}
\newcommand{\expct}[1]{\mathbb{E}\left[#1\right]}
\newcommand{\expctover}[2]{\mathbb{E}_{#1}\!\left[#2\right]}
\newcommand{\expctoverrlz}[2]{\mathbb{E}\left[#2\right]}

\newcommand{\set}[1]{\left\{#1\right\}}
\newcommand{\Real}{\mathbb R}
\newcommand{\eps}{\varepsilon}

\newcommand{\defref}[1]{Definition~\ref{#1}}

\newcommand{\figref}[1]{Fig.~\ref{#1}}
\newcommand{\eqnref}[1]{Eq.~(\ref{#1})}
\newcommand{\secref}[1]{\S\ref{#1}}
\newcommand{\thmref}[1]{Theorem~\ref{#1}}

\newcommand{\lemref}[1]{Lemma~\ref{#1}}
\newcommand{\algref}[1]{Algorithm~\ref{#1}}

\begin{document}

  \title{\ourtitle}
  \author{
     \name \!Daniel Golovin  
     \email {\textcolor{white}{antispam.}golovin@gmail.com}\\
     \addr California Institute of Technology\\
     Pasadena, CA 91125, USA
    \AND
     \name Andreas Krause 
     \email {\textcolor{white}{antispam.}krausea@ethz.ch}\\                    
     \addr ETH Zurich\\
     8092 Zurich, Switzerland 
     }

\editor{???}

\maketitle

\footnotetext{This work appeared in the Journal of Artificial Intelligence Research~\citep{golovin11jair}, and an
  eariler extended abstract appeared in the International Conference on Learning Theory~\citep{golovin10colt}.}

\begin{abstract}
Many problems in artificial intelligence require adaptively making a sequence of decisions with uncertain outcomes under partial observability. Solving such stochastic optimization problems is a fundamental but notoriously difficult challenge.  In this paper, we introduce the concept of \emph{\term submodularity}, generalizing submodular set functions to adaptive policies.  We prove that if a problem satisfies this property, a simple adaptive greedy algorithm is guaranteed to be competitive with the optimal policy. In addition to providing  performance guarantees for both stochastic maximization and coverage, \term submodularity can be exploited to drastically speed up the greedy algorithm by using lazy evaluations. We illustrate the usefulness of the concept by giving several examples of \term submodular objectives arising in diverse AI applications including management of sensing resources, viral marketing and active learning. Proving \term submodularity for these problems allows us to recover existing results in these applications as special cases, improve approximation guarantees and handle natural generalizations.
\end{abstract}


\section{Introduction} \label{sec:intro}
\looseness -1 In many  problems arising in artificial intelligence one
needs to adaptively make a sequence of decisions, taking into account
observations about the outcomes of past decisions.  Often these
outcomes are uncertain, and one may only know a probability
distribution over them.  Finding optimal policies for decision making
in such partially observable stochastic optimization problems is
notoriously intractable (see, e.g., \citet{littman98computational}).  A fundamental challenge is to identify classes of planning problems for which simple solutions obtain (near-) optimal performance.

In this paper, we
introduce the concept of \emph{\term submodularity}, and prove that
if a partially observable stochastic optimization problem satisfies this property, a simple adaptive greedy
algorithm is guaranteed to obtain near-optimal solutions. In fact, under reasonable complexity-theoretic assumptions, no polynomial time algorithm is able to obtain better solutions in general.
\Term
  submodularity generalizes the classical notion of submodularity\footnote{For
  an extensive treatment of submodularity, see the books
  of~\citet{fujishige05} and~\citet{SchrijverB}.}, which has
been successfully used to develop approximation algorithms for a
variety of non-adaptive optimization problems.
 Submodularity, informally, is an intuitive notion of diminishing
 returns, which states that adding an element to a small set helps
 more than adding that same element to a larger (super-)\ set.  A
 celebrated result of the work of~\citet{nemhauser78} guarantees that for such
 submodular functions, a simple greedy algorithm, which adds the element that maximally increases the objective value, selects a near-optimal set of $k$ elements.
Similarly, it is guaranteed to find a set of near-minimal cost that
achieves a desired quota of utility~\citep{wolsey82}, using
near-minimum average time to do so~\citep{streeter08}.
Besides guaranteeing theoretical performance bounds, submodularity allows us
to speed up algorithms without loss of solution quality by using lazy evaluations \citep{minoux78},
often leading to performance improvements of several orders of
magnitude \citep{leskovec07}.  Submodularity has been shown to be very useful in a variety of problems in artificial intelligence \citep{krause09ijcaitut}.

The challenge in generalizing
submodularity to adaptive planning --- where the action taken in each
step depends on information obtained in the previous steps  --- is that feasible solutions are now
policies (decision trees or conditional plans) instead of subsets. We propose a natural
generalization of the diminishing returns property for adaptive
problems, which reduces to the classical characterization of submodular set
functions for deterministic distributions. We show how these results
of~\citet{nemhauser78},~\citet{wolsey82},~\citet{streeter08},
and~\citet{minoux78} generalize to the adaptive setting.  Hence,
we demonstrate how \term submodular optimization problems enjoy
theoretical and practical benefits similar to those of classical,
nonadaptive submodular problems. We further demonstrate the usefulness
and generality of the concept by showing how it captures known results
in stochastic optimization and active learning as special cases,
admits tighter performance bounds,  leads to natural generalizations
and allows us to solve new problems for which no performance guarantees were known.

As a first example, consider the problem of deploying (or controlling) a collection of
sensors to monitor some spatial phenomenon. Each sensor can cover a
region depending on its sensing range. Suppose we would like to find
the best subset of $k$ locations to place the sensors.  In this
application, intuitively, adding a sensor helps more if we have placed
few sensors so far and helps less if we have already placed many sensors. We can formalize this diminishing returns property using the notion of submodularity --
the total area covered by the sensors is a submodular function defined
over all sets of locations.  \citet{krause07nearoptimal} show that
many more realistic utility functions in sensor placement (such as the
improvement in prediction accuracy w.r.t.~some probabilistic model)
are submodular as well. Now consider the following stochastic variant:
Instead of deploying a fixed set of sensors, we deploy one sensor at a
time.  With a certain probability, deployed sensors can fail, and our
goal is to maximize the area covered by the functioning sensors.
Thus, when deploying the next sensor, we need to take into account
which of the sensors we deployed in the past failed.  This problem has
been studied by \citet{AsadpourNS08} for the case where each sensor
fails independently at random.  In this paper, we show that the
coverage objective is \term submodular, and use this concept to handle
much more general settings (where, e.g., rather than all-or-nothing
failures there are different types of sensor failures of varying severity).
We also consider a related problem where the goal is to place the minimum number of
sensors to achieve the maximum possible sensor coverage (i.e., the coverage
obtained by deploying sensors everywhere), or more generally the
goal may be to achieve any fixed percentage of the maximum possible
sensor coverage.
Under the first goal, the problem is equivalent to one studied by~\citet{goemans06stochastic}, and
generalizes a problem studied by \citet{liu08near}.

As another example, consider a viral marketing problem, where we are
given a social network, and we want to influence as many people as
possible in the network to buy some product. We do that by giving the
product for free to a subset of the people, and hope that they
convince their friends to buy the product as well.  Formally, we have
a graph, and each edge $e$ is labeled by a number $0\leq p_e\leq 1$.  We ``influence'' a subset of nodes in the graph, and for each influenced node, their neighbors get randomly influenced according to the probability annotated on the edge connecting the nodes.  This process repeats until no further node gets influenced.  \citet{kempe03} show that the set function which quantifies the expected number of nodes influenced is submodular.  A natural stochastic variant of the problem is where we pick a node, get to see which nodes it influenced, then adaptively pick the next node based on these observations and so on.  We show that a large class of such adaptive influence maximization problems satisfies \term submodularity.

Our third application is in active learning,
where we are given an unlabeled data set, and we would like to
adaptively pick a small set of examples whose labels imply all other
labels.  The same problem arises in automated diagnosis, where we have
hypotheses about the state of the system (e.g., what illness a patient has), and would like to perform tests to identify the correct hypothesis.
In both domains we want to pick examples / tests to shrink the remaining
version space (the set of consistent hypotheses) as quickly as
possible. Here, we show that the reduction in version space
probability mass is
\term submodular, and use that observation to prove that the adaptive
greedy algorithm is a near-optimal querying policy.
Our results for active learning and automated diagnosis are also related to recent results of \citet{guillory10interactive,guillory2011-noisy-interactive-submod-cover} who study generalizations of submodular set cover to an interactive setting.  In contrast to our approach however, \citeauthor{guillory10interactive} analyze worst-case costs, and use rather different technical definitions and proof techniques.

We summarize our main contributions below, and provide a more
technical summary in  Table~\ref{table:results}.
At a high level, our main contributions are:
\begin{itemize}
\item We consider a particular class of partially observable adaptive stochastic optimization problems, which we prove to be hard to approximate in general.
\item We introduce the concept of \emph{\term submodularity}, and
  prove that if a problem instance satisfies this property, a simple
  adaptive greedy policy performs near-optimally, for both adaptive
  stochastic maximization and coverage, and also a natural min-sum objective.
\item We show how \term submodularity can be exploited by allowing the
  use of an accelerated adaptive greedy algorithm using lazy
  evaluations, and how we can obtain data-dependent bounds on the optimum.
\item We illustrate \term submodularity on several realistic problems,
  including Stochastic Maximum Coverage, Stochastic Submodular Coverage,
  Adaptive Viral Marketing, and
  Active Learning.  For these applications, \term submodularity allows
  us to recover known results and prove natural generalizations.
\end{itemize}

\newcommand{\AS}{A.S.\xspace}
\newcommand{\AMS}{A.M.S.\xspace}

\begin{table}[t]
  \begin{tabular}{|p{1.5in}|p{3.4in}|p{0.8in}|}
\hline
\hspace{0.5in} {\bf Name} & \hspace{1.15in} {\bf New Results} &
\hspace{3mm} {\bf Location} \\ %
\hline
\AS Maximization & Tight $(1-1/e)$-approx.\ for \AMS objectives &
\secref{ssec:max-cover-objective}, page~\pageref{ssec:max-cover-objective}\\
\hline
\AS Min Cost Coverage & Squared logarithmic approx.\ for \AMS objectives &
\secref{sec:min-cost-cover}, page~\pageref{sec:min-cost-cover}\\
\hline
\AS Min Sum Cover & Tight $4$-approx.\ for \AMS objectives &
\secref{sec:min-sum-cover}, page~\pageref{sec:min-sum-cover}\\
\hline
Data Dependent Bounds & Generalization to
\AMS functions & \secref{ssec:max-cover-objective}, page~\pageref{ssec:max-cover-objective}\\
\hline
Accelerated Greedy & Generalization of lazy evaluations to the
adaptive setting & \secref{sec:the-greedy-policy}, page~\pageref{sec:the-greedy-policy}\\
\hline
Stochastic Submodular Maximization &  Generalization of the previous $(1-1/e)$-approx.\ to \mbox{arbitrary}
per--item set distributions, and to item costs &
\secref{sec:stochastic-maximization}, page~\pageref{sec:stochastic-maximization}\\
\hline
Stochastic Set Cover &
$(\ln(n)+1)^2$-approx.\ for \mbox{arbitrary} per-item set distributions, with
item costs & \secref{sec:stochastic-set-cover}, page~\pageref{sec:stochastic-set-cover}\\
\hline
\mbox{Adaptive Viral} \mbox{Marketing} & Adaptive analog of previous
$(1-1/e)$-approx.\ for non-adaptive viral marketing, under more
general reward functions; squared logarithmic approx.\ for the adaptive
min cost cover version & \secref{sec:viral-marketing}, page~\pageref{sec:viral-marketing}\\
\hline
Active Learning & New analysis for generalized binary search and its approximate versions
with and without item costs & \secref{sec:active-learning}, page~\pageref{sec:active-learning}\\
\hline
Hardness in the absence of Adapt.\ Submodularity &
$\Omega(|\groundset|^{1-\epsilon})$-approximation hardness for \AS
Maximization, Min Cost Coverage, and Min-Sum Cover, if $f$ is not
\term submodular.  & \secref{sec:hardness}, page~\pageref{sec:hardness}\\
\hline
 \end{tabular}
  \caption{Summary of our theoretical results.  \AS is shorthand for
    ``adaptive stochastic'', and \AMS is shorthand for ``adaptive
    monotone submodular.'' \vspace{-5mm}}  \label{table:results}
\end{table}

\subsection{Organization} %

\looseness -1 This article is organized as follows.
In \secref{sec:problem-statment} (page~\pageref{sec:problem-statment}) we set up notation and formally define
the relevant adaptive optimization problems for general objective
functions.
\emph{For the reader's convenience, we have also provided a reference table of
important symbols on page~\pageref{table:symbol-table}.}
In~\secref{sec:term-submodularity} (page~\pageref{sec:term-submodularity}) we review the classical notion of submodularity and introduce the novel
\term submodularity property.  In~\secref{sec:the-greedy-policy}
(page~\pageref{sec:the-greedy-policy}) we introduce
the adaptive greedy policy, as well as an accelerated variant.
In~\secref{sec:greedy} (page~\pageref{sec:greedy}) we discuss the theoretical guarantees
that the adaptive greedy policy enjoys when applied to
problems with \term submodular objectives.
Sections~\ref{sec:stochastic-maximization} through~\ref{sec:active-learning}
provide examples on how to apply the \term submodular framework to
various applications, namely Stochastic Submodular Maximization
(\secref{sec:stochastic-maximization}, page~\pageref{sec:stochastic-maximization}), Stochastic Submodular Coverage
(\secref{sec:stochastic-set-cover}, page~\pageref{sec:stochastic-set-cover}),
Adaptive Viral Marketing (\secref{sec:viral-marketing}, page~\pageref{sec:viral-marketing}), and
Active Learning (\secref{sec:active-learning}, page~\pageref{sec:active-learning}).
In~\secref{sec:experiments} (page~\pageref{sec:experiments}) we report empirical results on two sensor
selection problems.
In~\secref{sec:adaptgap} (page~\pageref{sec:adaptgap}) we discuss the adaptivity gap of the problems we
consider, and in~\secref{sec:hardness} (page~\pageref{sec:hardness}) we prove hardness results
indicating that problems which are not \term submodular  can be extremely
inapproximable under reasonable complexity assumptions.
We review related work in~\secref{sec:related-work} (page~\pageref{sec:related-work}) and provide concluding remarks in
\secref{sec:conclusions} (page~\pageref{sec:conclusions}).  \AppendixA (page~\pageref{sec:proofs}) gives details of how to incorporate item
costs and includes all of the proofs omitted from the main text.

\section{Adaptive Stochastic Optimization} \label{sec:problem-statment}
We start by introducing notation and defining the general class of adaptive optimization problems that we address in this paper. For sake of clarity, we will illustrate our notation using the sensor placement application mentioned in \secref{sec:intro}. We give examples of other applications in~\secref{sec:stochastic-maximization},~\secref{sec:stochastic-set-cover},~\secref{sec:viral-marketing},
and~\secref{sec:active-learning}. 

\subsection{Items and Realizations} Let $\groundset$ be a finite set
of items (e.g., sensor locations).  Each item $\elem\in\groundset$ is
in a particular (initially unknown) state from a set $\outcomes$ of
possible states 
(describing whether a sensor
placed at location $\elem$ would malfunction or not). 
We represent the item states using a
function $\rlz:\groundset\rightarrow\outcomes$, called a 
\emph{realization} (of the states of all items in the ground set).
Thus, we say that $\rlz(\elem)$ is the state of $\elem$ under
realization $\rlz$.
We use $\rvrlz$ to denote a random realization. 
We take a Bayesian approach and assume that there is a known prior
probability distribution $\rlzprior := \Pr{\rvrlz = \rlz}$ over realizations (e.g., modeling
that sensors fail independently with failure probability), so that 
we can compute posterior distributions\footnote{In some situations, we
may not have exact knowledge of the prior $\rlzprior$.  Obtaining
algorithms that are robust to incorrect priors remains an interesting
source of open problems.  We briefly discuss some robustness
guarantees of our algorithm in~\secref{sec:the-greedy-policy} on page~\pageref{sec:robustness}.}.
We will consider problems where we sequentially pick an item
$\elem\in\groundset$, get to see its state $\rvrlz(\elem)$, pick the
next item, get to see its state, and so on (e.g., place a sensor, see whether it fails, and so on).  After each pick, our
observations so far can be represented as a \emph{partial realization}
$\prlz$, a function from some
subset of $\groundset$ (i.e., the set of items that we already picked)
to their states (e.g., $\prlz$ encodes where we placed sensors and which of them failed).  
For notational convenience, we sometimes represent  $\prlz$ as a relation, so that $\prlz \subseteq \groundset \times \outcomes$ equals 
$\set{(\elem, \outcome) : \prlz(\elem) = \outcome }$.
We use the notation $\dom(\prlz)=\set{\elem: \exists \outcome. (\elem,\outcome)\in\prlz}$ to refer to the domain of $\prlz$ (i.e., the set of items observed in $\prlz$).
A partial realization $\prlz$ is \emph{consistent}
with a realization $\rlz$ if they are equal everywhere in the domain
of $\prlz$.  
In this case we write $\rlz\sim\prlz$.  
If $\prlz$ and $\prlz'$ are both consistent with some $\rlz$, and 
$\dom(\prlz) \subseteq \dom(\prlz')$, we say $\prlz$ is a \emph{subrealization}
of $\prlz'$.  Equivalently, $\prlz$ is a subrealization
of $\prlz'$ if and only if, when viewed as relations, $\prlz \subseteq \prlz'$.

Partial realizations are similar to the notion of ``belief states'' in
Partially Observable Markov Decision Problems (POMDPs), as they encode
the effect of all actions taken (items selected) and observations made, and determine our
posterior belief about the state of the world (i.e., the state of all
items $e$ not yet selected, $\rlzmass{\rlz \mid \prlz } := \Pr{\rvrlz =
  \rlz \mid \rvrlz \sim \prlz}$).

\subsection{Policies} 
\begin{figure} 
 \centering 
 \includegraphics[width=0.85\textwidth]{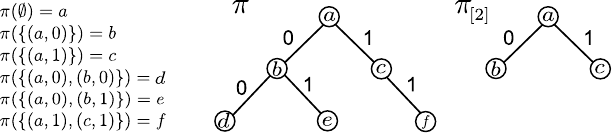}
 \caption{Illustration of a policy $\policy$, its corresponding
   decision tree representation, and the decision tree representation
   of $\prune{\policy}{2}$, the level $2$ truncation of $\policy$
   (as defined in \secref{ssec:max-cover-objective}).}
\label{fig:policyastree}
 \end{figure}

We encode our adaptive strategy for picking items as a \emph{policy}
$\policy$, which is a function from a set of partial realizations to
$\groundset$, specifying which item to pick next under a particular
set of observations (e.g., $\policy$ chooses the next sensor location
given where we have placed sensors so far, and whether they failed or
not). 
We also allow randomized policies that are functions 
from a set of partial realizations to distributions on $\groundset$,
though our emphasis will primarily be on deterministic policies.
If $\prlz$ is not in the domain of $\policy$, the policy
terminates (stops picking items) upon observation of $\prlz$.
We use $\dom(\policy)$ to denote the domain of $\policy$.
Technically, we require that $\dom(\policy)$ be closed under
subrealizations.  That is, if $\prlz' \in \dom(\policy)$ and $\prlz$
is a subrealization of $\prlz'$ then $\prlz \in \dom(\policy)$.  
We use the notation $\played{\policy}{\rlz}$ to refer to the set of items selected by $\policy$ under realization $\rlz$.  Each deterministic policy $\policy$ can be associated with a decision
tree $T^{\policy}$ in a natural way (see \figref{fig:policyastree} for
an illustration).  
Here, we adopt a policy-centric view that admits concise
notation, though we find the decision tree view to be valuable conceptually.

Since partial realizations are similar to POMDP belief states, our definition of policies is similar to the notion of policies in POMDPs, which are usually defined as functions from belief states to actions. We will further discuss the relationship between the stochastic optimization problems considered in this paper and POMDPs in Section~\ref{sec:related-work}.

\subsection{Adaptive Stochastic Maximization, Coverage, and Min-Sum Coverage} We wish to maximize, subject to some constraints, a utility function $f :2^{\groundset} \times \outcomes^\groundset \to \NonNegativeReals$ that depends on which items we pick and which state each item is in (e.g., modeling the total area covered by the working sensors).   
Based on this notation, the expected utility of a policy $\policy$ is
$\avgf(\policy) := \expctoverrlz{\rvrlz}{f(\played{\policy}{\rvrlz},
  \rvrlz)}$ where the expectation is taken with respect to $\rlzprior$.
The goal of the \emph{Adaptive Stochastic Maximization} problem is to 
find a policy $\policy^{*}$ such that 
\begin{equation}\policy^{*}\in\argmax_{\policy} \avgf(\policy) \text{
    subject to } |\played{\policy}{\rlz}|\leq \budget\text{ for all }\rlz, 
\label{eq:stochmax}\end{equation} where $\budget$ is a budget on how many items can be picked (e.g., we would like to adaptively choose $k$ sensor locations such that the working sensors provide as much information as possible in expectation).

\ifthenelse{\boolean{istechrpt}}{
Alternatively, we can specify a quota $\quota$ of utility that we
would like to obtain, and try to find the cheapest policy achieving
that quota (e.g., we would like to achieve a certain amount of information, as cheaply as possible in expectation). Formally, we define the average cost $\acst{\policy}$ of a
policy as the expected number of items it picks, so that
$\acst{\policy}:=\expctoverrlz{\rvrlz}{|\played{\policy}{\rvrlz}|}$. Our goal is then to find
\begin{equation}
\policy^{*}\in\argmin_{\policy} \acst{\policy}\text{ such that } f(\played{\policy}{\rlz},\rlz)\geq \quota\text{ for all }\rlz,\label{eq:stochcover}
\end{equation}
i.e., the policy $\policy^{*}$ that minimizes the expected number of
items picked such that under all possible realizations, at least
utility $Q$ is achieved. We call Problem~\ref{eq:stochcover} the
\emph{Adaptive Stochastic Minimum Cost Cover} problem.  We will also
consider the problem where we want to minimize the worst-case cost
$\wcst{\policy}:=\max_{\rlz}{|\played{\policy}{\rlz}|}$.  This
worst-case cost $\wcst{\policy}$ is the cost
incurred under adversarially chosen realizations, or  equivalently the
depth of the deepest leaf in $T^{\policy}$, the decision tree
associated with $\policy$.

Yet another important variant is to minimize the average time required
by a policy to obtain its utility.  Formally, let $\util{\policy}{t}$ be
the expected utility obtained by $\policy$ after $t$
steps\footnote{For a more formal definition of $\util{\policy}{t}$,
  see \secref{sec:proofs-min-sum-cover} on page~\pageref{sec:proofs-min-sum-cover}.},
let $\quota = \expctoverrlz{\rlz}{f(\groundset, \rvrlz)}$ be the maximum
possible expected utility, 
and define the \emph{min-sum cost} $\costminsum{\policy}$ of a policy as 
$\costminsum{\policy} := \sum_{t = 0}^{\infty} \paren{
\quota -  \util{\policy}{t}}$.  We then define the  \emph{Adaptive
Stochastic Min-Sum Cover} problem as the search for
\begin{equation}
\policy^{*} \in \argmin_{\policy} \costminsum{\policy}. \label{eq:minsumcover}
\end{equation}

Unfortunately, as we will show in \secref{sec:hardness}, even for
linear functions $f$, i.e., those where $f(A,\rlz)=\sum_{\elem\in
  A}w_{\elem,\rlz}$ is simply the sum of weights (depending on the
realization $\rlz$), Problems~(\ref{eq:stochmax}),  (\ref{eq:stochcover}),
and (\ref{eq:minsumcover})  
are hard to approximate 
under reasonable complexity theoretic assumptions. 
Despite the hardness of the general problems, 
in the following sections we will identify conditions that are sufficient to 
allow us to approximately solve them.
}{
Unfortunately, as we will show in \secref{sec:hardness}, even for
linear functions $f$, i.e., those where $f(A,\rlz)=\sum_{\elem\in
  A}w_{\elem,\rlz}$ is simply the sum of weights (depending on the
realization $\Phi$), Problems~(\ref{eq:stochmax})
is hard to approximate under reasonable complexity theoretic
assumptions.
Despite the hardness of the general problem, 
in the following sections we will identify conditions that are sufficient to 
allow us to approximately solve it.
}

\subsection{Incorporating Item Costs}  
Instead of quantifying the cost of a set $\played{\policy}{\rlz}$ by
the number of elements $|\played{\policy}{\rlz}|$, we can also
consider the case where each item $\elem\in\groundset$ has a 
cost $c(\elem)$, and the cost of a set
$S\subseteq\groundset$ is $c(S)=\sum_{\elem\in S}c(\elem)$. We can
then consider variants of
Problems~\eqref{eq:stochmax},~\eqref{eq:stochcover},
and~\eqref{eq:minsumcover} with the $|\played{\policy}{\rlz}|$
replaced by 
$c(\played{\policy}{\rlz})$. For clarity of presentation, we will focus on the unit cost case, i.e., $c(\elem)=1$ for all $e$, 
and explain how our results generalize to the non-uniform case in \appendixA.

\section{\Term Submodularity} \label{sec:term-submodularity}

\looseness -1 We first review the classical notion of submodular set functions, and then introduce the novel notion of \term submodularity.

\subsection{Background on Submodularity} 
\label{sec:submod-background}
Let us first consider the very special case where $\rlzprior$ is
deterministic or, equivalently, $|\outcomes|=1$ (e.g., in our sensor
placement applications, sensors never fail).  In this case, the
realization $\rlz$ is known to the decision maker in advance, and thus
there is no benefit in adaptive selection.  
Given the realization $\rlz$, Problem~\eqref{eq:stochmax} is equivalent to finding a set $A^{*}\subseteq\groundset$ such that
\begin{equation}A^{*}\in\argmax_{A\subseteq\groundset} f(A,\rlz)\text{ such that }|A|\leq \budget.\label{eq:detmax}\end{equation}
For most interesting classes of utility functions $f$, this is an NP-hard optimization problem.  However, in many practical problems, such as those mentioned in \secref{sec:intro},  $f(A)=f(A,\rlz)$ satisfies \emph{submodularity}.  A set function $f:2^{\groundset}\rightarrow\Real$ is called submodular if, whenever $A\subseteq B\subseteq \groundset$ and $\elem\in \groundset\setminus B$ it holds that
\begin{equation}
f(A\cup\{\elem\})-f(A)\geq
f(B\cup\{\elem\})-f(B),\label{eq:dimreturns}
\end{equation}
\looseness -1 i.e., adding $\elem$ to the smaller set $A$ increases $f$ by at least as
much as adding $\elem$ to the superset $B$.  Furthermore, $f$ is
called \emph{monotone}, if, whenever $A\subseteq B$ it holds that
$f(A)\leq f(B)$ (e.g., adding a sensor can never reduce the amount of information obtained). A celebrated result by~\citet{nemhauser78} states
that for monotone submodular functions with $f(\emptyset)=0$, a
simple greedy algorithm that starts with the empty set,
$A_{0}=\emptyset$ and chooses
\begin{equation}A_{i+1}=A_{i}\cup\{\argmax_{\elem\in\groundset\setminus A_{i}}
f(A_{i}\cup\{\elem\})\}\label{eq:greedyrule}\end{equation} guarantees that $f(A_{\budget})\geq (1-1/e)
\max_{|A|\leq \budget} f(A)$. Thus, the greedy set $A_{\budget}$
obtains at least a $(1-1/e)$ fraction of the optimal value
achievable using $\budget$ elements.  
Furthermore,
\citet{feige98threshold} shows that this result is tight if
$\class{P} \neq \NP$; under this assumption \emph{no polynomial time algorithm} can do strictly better than the greedy algorithm, i.e., achieve a
$(1 - 1/e + \epsilon)$-approximation for any constant $\epsilon > 0$, 
even for
the special case of Maximum $k$-Cover where $f(A)$ is the
cardinality of the union of sets indexed by $A$. 
Similarly,  \citet{wolsey82} shows that the same greedy algorithm also
near-optimally solves the
deterministic case of Problem~\eqref{eq:stochcover}, called  the \emph{Minimum Submodular Cover} problem:
\begin{equation}
A^{*}\in\argmin_{A\subseteq\groundset} |A|\text{ such that }f(A)\geq
\quota
\label{eq:detcover}.
\end{equation}
\looseness -1 Pick the first set $A_{\ell}$ constructed by the greedy
algorithm such that $f(A_{\ell})\geq \quota$. Then, for integer-valued
submodular functions, $\ell$ is at most $|A^{*}|(1+\log \max_{\elem}
f(\elem))$, i.e., the greedy set is at most a logarithmic factor
larger than the smallest set achieving quota $\quota$.  For the 
special case of Set Cover, where $f(A)$ is the cardinality
of a union of sets indexed by $A$, this result matches a lower bound by \citet{feige98threshold}: Unless $\NP\subseteq \text{DTIME}(n^{\cO(\log\log n)})$, Set Cover is hard to approximate by a factor better than $(1-\varepsilon)\ln Q$, where $Q$ is the number of elements to be covered.

Now let us relax the assumption that $\rlzprior$ is
deterministic. In this case, we may still want to find a non-adaptive
solution (i.e., a constant policy $\policy_{A}$ that always picks set
$A$ independently of $\rvrlz$) maximizing $\avgf(\policy_{A})$. If $f$
is \emph{pointwise} submodular, i.e., $f(A,\rlz)$ is  submodular in
$A$ for any fixed $\rlz$, the function $f(A)=\avgf(\policy_{A})$ is
submodular, since nonnegative linear combinations of submodular
functions remain submodular.  Thus, the greedy algorithm allows us to
find a near-optimal \emph{non-adaptive} policy. That is, in our sensor placement example, if we are willing to commit to all locations \emph{before} finding out whether the sensors fail or not, the greedy algorithm can provide a good solution to this non-adaptive problem.

However, in practice, we may be more interested in obtaining a
non-constant policy $\policy$, that \emph{adaptively} chooses items
based on previous observations (e.g., takes into account which sensors
are working before placing the next sensor).  In many settings,
selecting items adaptively offers huge advantages, analogous to the
advantage of binary search over sequential (linear)
search\footnote{We provide a well--known example in active learning that illustrates this phenomenon
  crisply in~\secref{sec:active-learning}; see~\figref{fig:activelearning} on
  page~\pageref{sec:active-learning}.  We consider the
  general question of the magnitude of the potential benefits of
  adaptivity in~\secref{sec:adaptgap} on page~\pageref{sec:adaptgap} .}.  
Thus, the question is whether there is a natural extension of submodularity to policies. In the following, we will develop such a notion -- \emph{\term submodularity}.

\subsection{\Term Monotonicity and Submodularity} 
The key challenge is to find 
appropriate generalizations of monotonicity and 
of the diminishing returns condition
\eqref{eq:dimreturns}.  We begin by again considering the very special
case where $\rlzprior$ is deterministic, so that the policies are
non-adaptive.  In this case a policy $\policy$ simply specifies a
sequence of items $(e_1, e_2, \ldots, e_r)$ which it selects in order.
Monotonicity in this context can be characterized as the property that
``the marginal benefit of selecting an item is always nonnegative,'' meaning that for all such sequences  $(e_1, e_2, \ldots, e_r)$, items $e$ and $1\leq i \leq r$ it holds that $f(\set{e_{j}:j\leq i}\cup\{e\})-f(\set{e_{j}:j\leq i})\geq 0$. 
Similarly, submodularity can be viewed as the property that
``selecting an item later never increases its marginal benefit,''
meaning that for all sequences $(e_1, e_2, \ldots, e_r)$, items
$e$, and all $i \le r$, 
$f(\set{e_j : j \le i} \cup \set{e}) - f(\set{e_j : j \le i}) \ge f(\set{e_j : j \le r} \cup \set{e}) - f(\set{e_j : j \le r}) $.

We take these views of monotonicity and submodularity when defining
their \term analogues, by using an appropriate generalization of the marginal benefit.
When moving to the general adaptive setting, the challenge is that
the items' states are now random and only revealed upon selection. A
natural approach is thus to condition on  observations (i.e., partial
realizations of selected items), and take the expectation with respect to the items that we consider selecting. 
Hence, we define our \term monotonicity and submodularity properties in terms of the
\emph{conditional expected marginal benefit} of an item.

\begin{definition}[Conditional Expected Marginal Benefit] \label{def:conditional-marginal-gain}
Given a partial realization $\prlz$ and an item $e$, the 
  \emph{conditional expected marginal benefit} of $e$ 
 conditioned on having observed $\prlz$, denoted $\diff{\prlz}{\elem}$, is 
\begin{equation}
   \label{eq:adaptive-derivative-def}
   \diff{\prlz}{\elem} :=
 \expctoverrlz{\rlz}{\frac{}{}\!  f(\dom(\prlz) \cup \set{\elem}, \rvrlz) -
   f(\dom(\prlz), \rvrlz)\ \bigg| \ \rvrlz\sim\prlz}%
 \end{equation}
where the expectation is computed with respect to $\rlzmass{\rlz \mid
  \prlz} = \Pr{\rvrlz = \rlz
  \mid \rvrlz \sim \prlz}$. 
Similarly, the \emph{conditional expected marginal benefit of a
  policy} $\policy$ is 
\begin{equation}
   \label{eq:adaptive-derivative-policy-def}
   \diff{\prlz}{\policy} :=
 \expctoverrlz{\rlz}{\frac{}{}\!  f(\dom(\prlz) \cup \played{\policy}{\rvrlz}, \rvrlz) -
   f(\dom(\prlz), \rvrlz)\ \bigg| \ \rvrlz\sim\prlz}\mbox{.}  
 \end{equation}
\end{definition}
 
\noindent 
In our sensor placement example, $\diff{\prlz}{\elem}$
quantifies the expected amount of additional area covered by placing a
sensor at location $\elem$, in expectation over the posterior
distribution 
$\rlzmassover{\rvrlz(\elem)}{\outcome} := \Pr{\rvrlz(\elem) = \outcome \mid \rvrlz \sim \prlz}$
of whether the sensor will fail or not, and taking into account the
area covered by the placed working sensors as encoded by $\prlz$.
Note that the benefit we have accrued upon observing $\prlz$ (and
hence after having selected the items in $\dom(\prlz)$) is $\expct{
  f(\dom(\prlz), \rvrlz)\ \mid \ \rvrlz\sim\prlz}$, which is the
benefit term subtracted out in \eqnref{eq:adaptive-derivative-def} and
\eqnref{eq:adaptive-derivative-policy-def}. 
Similarly, the expected total benefit obtained after observing $\prlz$
and then selecting $\elem$ is $\expct{f(\dom(\prlz) \cup \set{\elem},
  \rvrlz)\ \mid \ \rvrlz\sim\prlz}$.  The corresponding benefit for
running $\policy$ after observing $\prlz$ is slightly more complex.
Under realization $\rlz \sim \prlz$, the final cumulative benefit will
be $f(\dom(\prlz) \cup \played{\policy}{\rlz}, \rlz)$.  Taking the
expectation with respect to $\rlzmass{\rlz \mid \prlz}$ and
subtracting out the benefit already obtained by $\dom(\prlz)$ then yields
the conditional expected marginal benefit of $\policy$.

We are now ready to introduce our generalizations of monotonicity and submodularity to the adaptive setting:

\begin{definition}[\Term Monotonicity] \label{def:adapt-monotonicity}
A function $f:2^{\groundset} \times O^{\groundset} \to \NonNegativeReals$
is \emph{\term monotone} with respect to distribution $\rlzprior$ 
if the conditional expected marginal benefit of any item is
nonnegative, i.e., for all $\prlz$ with $\Pr{\rvrlz \sim \prlz} > 0$ and all $e \in \groundset$ 
we have 
\begin{equation}\diff{\prlz}{\elem} \ge 0.\label{eq:adapt-monotone}\end{equation}
\ignore{
Given two policies $\policy_1, \policy_2$ 
define $\append{\policy_1}{\policy_2}$ as the policy obtained by running $\policy_1$ to completion, and then running policy 
 $\policy_2$ as if from a fresh start, ignoring the information
 gathered\footnote{Technically, if under any realization $\rlz$ policy
   $\policy_2$ selects an item that
   $\policy_1$ previously selected, then $\append{\policy_1}{\policy_2}$
   cannot be written as a function from a set of partial realizations
   to $\groundset$, i.e., it is not a policy.  This can be amended by
   allowing partial realizations to be multisets over elements of
   $\groundset \times \outcomes$, so that, e.g., if $e$ is played
   twice
   then $(e, \rlz(e))$ appears twice in
   $\prlz$.  However, in the interest of readability we will avoid this more cumbersome multiset
   formalism, and abuse notation slightly by calling
   $\append{\policy_1}{\policy_2}$ a policy.  This issue arises
   whenever we run some policy and then run another from a fresh start.}
 during the running of $\policy_1$. 
A function $f:2^{\groundset} \times O^{\groundset} \to \NonNegativeReals$
is \emph{\term monotone} 
with respect to distribution $\prob{\rlz}$ 
if for all policies $\policy,\policy'$  it holds that 
$\avgf(\pi)\leq \avgf( \append{\pi'}{\pi})$, where 
$\avgf(\policy) := \expctoverrlz{\rlz}{f(\played{\policy}{\rlz}, \rlz)}$
is defined w.r.t.~$\rlzprior$. 
} %
\end{definition}

\begin{definition}[\Term Submodularity] \label{def:adapt-submod}
A function $f:2^{\groundset} \times O^{\groundset} \to \NonNegativeReals$
is \emph{\term submodular} with respect to distribution $\rlzprior$
if the conditional expected marginal benefit of any fixed item does not
increase as more items are selected and their states are observed.
Formally, $f$ is \term submodular w.r.t. $\rlzprior$
if for all $\prlz$ and $\prlz'$ such that $\prlz$ is a subrealization of $\prlz'$ (i.e., $\prlz
  \subseteq \prlz'$), and for all $\elem \in \groundset \setminus \dom(\prlz')$, we have 
\begin{equation} \label{eqn:adapt-submod}
\diff{\prlz}{\elem} \ge \diff{\prlz'}{\elem}.
\end{equation}
\end{definition}

From the decision tree perspective, the condition 
$\diff{\prlz}{\elem} \ge \diff{\prlz'}{\elem}$ amounts to saying that
for any decision tree $T$, if we are at a node $v$ in $T$ which selects an item $\elem$, and compare the
expected marginal benefit of $\elem$ selected at $v$ with the expected marginal benefit
$\elem$ would have obtained if it were selected at an ancestor of
$v$ in $T$, then the latter must be no smaller than the former.
Note that when comparing the two expected marginal benefits, there is
a difference in both the set of items previously selected
(i.e., $\dom(\prlz)$ vs. $\dom(\prlz')$) and in the distribution over
realizations (i.e., $\rlzmass{\rlz \mid \prlz}$ vs. $\rlzmass{\rlz \mid \prlz'}$).  
It is also worth emphasizing that \term
submodularity is defined relative to the distribution $\rlzprior$ over
realizations; it is possible that $f$ is
\term submodular with respect to one distribution, but not with
respect to another.

We will give concrete examples of \term monotone and \term
submodular functions that arise in the applications introduced in
\secref{sec:intro}  
in~\secref{sec:stochastic-maximization},~\secref{sec:stochastic-set-cover},~\secref{sec:viral-marketing},
and~\secref{sec:active-learning}. 
In \appendixA, we will explain how the notion of \term submodularity can be extended to handle non-uniform costs (since, e.g., the cost of placing a sensor at an easily accessible location may be smaller than at a location that is hard to get to).

\subsection{Properties of \Term Submodular Functions} It can be seen that \term monotonicity and \term
submodularity enjoy similar closure properties as monotone submodular
functions.  In particular, if $w_{1},\dots,w_{m}\geq 0$ and
$f_{1},\dots,f_{m}$ are \term monotone submodular w.r.t.~distribution
$\rlzprior$, then $f(A,\rlz) = \sum_{i=1}^{m}w_{i} f_{i}(A,\rlz)$ is
\term monotone submodular w.r.t. $\rlzprior$.  
Similarly, for a fixed constant $c\geq 0$ and \term monotone
submodular function $f$, the function $g(E,\rlz)=\min(f(E,\rlz),c)$ is
\term monotone submodular.
Thus, \term monotone submodularity is preserved by nonnegative linear combinations and by truncation.
\Term monotone submodularity is also preserved by 
restriction, so that if 
$f:2^{\groundset} \times \outcomes^{\groundset} \to \NonNegativeReals$
is \term monotone submodular w.r.t. $\rlzprior$, then for any $\elem \in \groundset$, the function 
$g : 2^{\groundset \setminus \set{\elem}} \times \outcomes^{\groundset} \to \NonNegativeReals$
defined by $g(A, \rlz) := f(A, \rlz)$ for all $A \subseteq \groundset
\setminus \set{\elem}$ and all $\rlz$ is also \term submodular w.r.t. $\rlzprior$.
Finally, if 
$f:2^{\groundset} \times \outcomes^{\groundset} \to \NonNegativeReals$
is \term monotone submodular w.r.t. $\rlzprior$ then for each partial
realization $\prlz$ the conditional function  
$g(A, \rlz) := f(A \cup \dom(\prlz), \rlz)$
is \term monotone
submodular w.r.t. $\rlzmass{\rlz \mid \prlz} := \Pr{\rvrlz = \rlz \mid
  \rvrlz \sim \prlz}$.

\subsection{What Problem Characteristics Suggest Adaptive
  Submodularity?}
\Term submodularity is a diminishing returns property for policies.
Speaking informally, it can be applied in situations where there is an objective function
to be optimized does not feature synergies in the benefits of items
conditioned on observations.   In some cases, the primary objective
might not have this property, but a suitably chosen proxy of it does, as is
the case with active learning with persistent noise~\citep{golovin10nips,bellala10modified}.
We give example applications in \secref{sec:stochastic-maximization}
through \secref{sec:active-learning}.   It is also worth mentioning
where \term submodularity is \emph{not} directly applicable.
An extreme example of synergistic effects between items conditioned on
observations is the class of ``treasure hunting'' instances used to prove \thmref{thm:hardness}
on page~\pageref{thm:hardness}, where the (binary) state of certain
groups of items encode the treasure's location in a complex manner.  Another 
problem feature which \term submodularity does not directly address is
the possibility that items selection can alter the underlying realization $\rlz$, as
is the case for the problem of optimizing policies for general POMDPs.

\section{The Adaptive Greedy Policy} \label{sec:the-greedy-policy}

The classical non-adaptive greedy algorithm~\eqref{eq:greedyrule} has a natural generalization to the adaptive setting.
The \emph{greedy policy} $\greedypolicy$ tries, at each iteration, to
myopically increase the expected objective value, given its current
observations.
That is, suppose
$f:2^{\groundset} \times \outcomes^{\groundset} \to \NonNegativeReals$ is the objective, and
 $\prlz$ is the partial realization indicating the states of items selected so
far.
Then the greedy policy will select the item $\elem$ maximizing the
expected increase in value, conditioned on the observed states of
items it has
already selected (i.e., conditioned on $\rvrlz\sim\prlz$).  That is, it
will select $\elem$ to maximize the conditional expected
marginal benefit $\diff{\prlz}{\elem}$ as defined in \eqnref{eq:adaptive-derivative-def}.
\looseness -1
Pseudocode of the adaptive greedy algorithm is given in
Algorithm~\ref{alg:greedy}.  The only difference to the classic,
non-adaptive greedy algorithm studied by \citet{nemhauser78}, is
Line~\ref{ln:adapt}, where an observation $\rvrlz(\elem^{*})$ of the
selected item $\elem^{*}$ is obtained. Note that the algorithms
in this section are presented for Adaptive Stochastic Maximization.
For the coverage objectives, we simply keep selecting items as prescribed by
$\greedypolicy$ until achieving the quota on objective value (for the
min-cost objective) or until we have selected every item (for the
min-sum objective).

\subsection{Incorporating Item Costs}  The adaptive greedy algorithm can be naturally modified
to handle non-uniform item costs by replacing its selection rule by
$$\elem^{*}\in\argmax_{e} \frac{\diff{\prlz}{\elem}}{c(\elem)}.$$
In the following, we will focus on the uniform cost case
($c \equiv 1$), and defer the analysis with costs to \appendixA.

\looseness -1 \subsection{Approximate Greedy Selection} In some applications, finding an item maximizing
$\diff{\prlz}{\elem}$ may be computationally
intractable, and the best we can do is find an $\alpha$-approximation
to the best greedy selection.  This means we find an $\elem'$ such that
$$\diff{\prlz}{\elem'} \ge \frac{1}{\alpha} \max_{\elem}
\diff{\prlz}{\elem}.$$  We call a policy which always
selects such an item an \emph{$\alpha$-approximate greedy policy}.

\begin{algorithm}
 \KwIn{Budget $k$; ground set $\groundset$; distribution $\rlzprior$;
   function $f$.}
\KwOut{Set $A \subseteq \groundset$ of size $k$}
\Begin{
\lnl{ln:ag1}   $\groundsubset\leftarrow\emptyset$; $\prlz\leftarrow\emptyset$\;
\lnl{ln:ag2}   \For{$i=1$ \KwTo $k$}{
\lnl{ln:ag3}       \lForEach{ $\elem\in\groundset\setminus \groundsubset$}{
      	compute $\diff{\prlz}{\elem} =
\expctoverrlz{\rvrlz}{f(\groundsubset \cup \set{\elem}, \rvrlz) - f(\groundsubset, \rvrlz)\ | \
\rvrlz\sim\prlz
} $
       }\;
\lnl{ln:ag5}       Select $\elem^{*}\in\argmax_{e}
\diff{\prlz}{\elem}$\;
\lnl{ln:ag6}    Set $\groundsubset\leftarrow \groundsubset\cup\{\elem^{*}\}$\;
\lnl{ln:adapt} Observe $\rvrlz(\elem^{*})$;    Set $\prlz\leftarrow\prlz\cup\set{(\elem^{*},\rvrlz(\elem^{*}))}$\;
   }
 } \label{alg:greedy} \caption{The adaptive greedy
algorithm, which implements the greedy policy.}
\end{algorithm}

\subsection{Robustness \& Approximate Greedy Selection}
\label{sec:robustness}
As we will show, $\alpha$-approximate greedy policies have
performance guarantees on several problems.
The fact that these performance guarantees of greedy policies are
robust to approximate greedy selection suggests a particular
robustness guarantee against incorrect priors $\rlzprior$.
Specifically, if our incorrect prior $\rlzmasssym'$ is such that when we evaluate
$\diff{\prlz}{\elem}$ we err by a multiplicative factor of at most
$\alpha$, then when we compute the greedy policy with respect to
$\rlzmasssym'$ we are actually implementing an $\alpha$-approximate
greedy policy (with respect to the true prior), and hence obtain the
corresponding guarantees.  For example, a sufficient condition for
erring by at most a multiplicative factor of $\alpha$ is
that there exists $c \le 1$ and $d \ge 1$ with $\alpha = d/c$ such that
 $c\,\rlzmass{\rlz} \le \rlzmasssym'\paren{\rlz} \le
 d\,\rlzmass{\rlz}$ for all $\rlz$, where $\rlzmasssym$ is the true prior.

\subsection{Lazy Evaluations and the Accelerated Adaptive Greedy
  Algorithm}
The definition of \term submodularity
allows us to implement an
``accelerated'' version of the adaptive greedy algorithm using lazy evaluations
of marginal benefits as originally suggested for the
non-adaptive case by~\citet{minoux78}.
The idea is as follows.
Suppose we run $\greedypolicy$ under some fixed realization $\rlz$,
and select items $\elem_1, \elem_2, \ldots, \elem_k$.
Let $\prlz_i := \set{(\elem_j, \rlz(\elem_j) : j \le i)}$ be the
partial realizations observed during the run of $\greedypolicy$.
The adaptive greedy algorithm computes $\diff{\prlz_i}{\elem}$ for all $\elem \in \groundset$ and
$0 \le i < k$, unless $\elem \in \dom(\prlz_i)$.
Naively, the algorithm thus needs to compute $\Theta(|\groundset| k)$ marginal benefits (which can be expensive to compute).
The key insight is that
$i \mapsto \diff{\prlz_i}{\elem}$ is nonincreasing for all $\elem \in
\groundset$, because of the \term submodularity of the
objective.
Hence, if when deciding which item to select as $e_i$ we know
$ \diff{\prlz_j}{\elem'} \le  \diff{\prlz_i}{\elem}$ for some items $\elem'$ and
$\elem$ and $j < i$, then we may conclude $\diff{\prlz_i}{\elem'} \le \diff{\prlz_i}{\elem}$ and hence
eliminate the need to compute $\diff{\prlz_i}{\elem'}$.
The accelerated version of the adaptive greedy algorithm exploits this
observation in a principled manner, by computing $\diff{\prlz}{\elem}$
for items $\elem$ in decreasing order of the upper bounds known on
them, until it finds an item whose value is at least as great as the upper bounds of
all other items.  Pseudocode of this version of the adaptive greedy algorithm is given in Algorithm~\ref{alg:acc-greedy}.

In the non-adaptive setting, the use of lazy evaluations has been shown
to significantly reduce running times in practice~\citep{leskovec07}.
We evaluated the naive and accelerated implementations of the
adaptive greedy algorithm on two sensor selection problems, and obtained
speedup factors that range from roughly $4$ to $40$ for those problems.
See~\secref{sec:experiments} on page~\pageref{sec:experiments} for details.

\begin{algorithm}
 \KwIn{Budget $k$; ground set $\groundset$; distribution $\rlzprior$;
   function $f$.}
\KwOut{Set $A \subseteq \groundset$ of size $k$}
\Begin{
\lnl{ln:acc1}   $\groundsubset\leftarrow\emptyset$; $\prlz\leftarrow\emptyset$;
  Priority Queue $Q \leftarrow \textsc{empty\_queue}$\;
\lnl{ln:acc3}   \lForEach{ $\elem\in\groundset$}{$Q.\op{insert}(\elem, +\infty)$}\;
\lnl{ln:acc4}   \For{$i=1$ \KwTo $k$}{
\lnl{ln:acc5}        $\delta_{\max} \leftarrow -\infty$; $\elem_{\max} \leftarrow \NULL$\;
\lnl{ln:acc6}        \While{ $\delta_{\max} < Q.\op{maxPriority}(\, )$ }{
\lnl{ln:acc7}          $\elem \leftarrow Q.\op{pop}(\, )$\;
\lnl{ln:acc8}          $\delta \leftarrow \diff{\prlz}{\elem}  =
          \expctoverrlz{\rvrlz}{f(\groundsubset \cup \set{\elem}, \rvrlz) -
            f(\groundsubset, \rvrlz)\ | \ \rvrlz\sim\prlz}  $\;
\lnl{ln:acc9}          $Q.\op{insert}(\elem, \delta)$\;
\lnl{ln:acc9b}          \If{ $\delta_{\max} < \delta$ }{
\lnl{ln:acc10}            $\delta_{\max} \leftarrow \delta$; $\elem_{\max} \leftarrow \elem$\;
          }
        }
\lnl{ln:acc11}        $A \leftarrow A \cup \set{\elem_{\max}}$; $Q.\op{remove}(\elem_{\max})$\;
        \lnl{ln:acc-adapt} Observe $\rvrlz(\elem_{\max})$;    Set $\prlz\leftarrow\prlz\cup\set{(\elem_{\max},\rvrlz(\elem_{\max}))}$\;
   }
 }
 \label{alg:acc-greedy}
\caption{The accelerated version of the adaptive greedy
algorithm.  Here, $Q.\op{insert}(\elem, \delta)$ inserts $\elem$ with
priority $\delta$,  $Q.\op{pop}(\, )$ removes and returns the
item with greatest priority, $Q.\op{maxPriority}(\, )$ returns the
maximum priority of the elements in $Q$, and $Q.\op{remove}(\elem)$
deletes $\elem$ from $Q$.}
\end{algorithm}

\section{Guarantees for the Greedy Policy} \label{sec:greedy}
\looseness -1 In this section we show that if the objective function is \term
submodular with respect to the probabilistic model of the
environment in which we operate, then the greedy policy inherits
precisely the performance guarantees of the greedy
algorithm for classic (non-adaptive) submodular maximization and submodular coverage
problems, such as Maximum $k$-Cover and Minimum
Set Cover, as well as min-sum submodular
coverage problems, such as Min-Sum Set Cover.
In fact, we will show that this holds true more generally:
$\alpha$--approximate greedy policies inherit precisely the performance guarantees of
$\alpha$--approximate greedy algorithms for these classic problems. These guarantees suggest that \term submodularity is an
appropriate generalization of submodularity to policies.  In this section
we focus on the unit cost case (i.e., every item has the same cost).  In \appendixA we
provide the proofs omitted in this section, and
show how our results extend to non-uniform item costs if we greedily
maximize the expected benefit/cost ratio.

\subsection{The Maximum Coverage Objective} \label{ssec:max-cover-objective}

In this section we consider the maximum coverage objective, where the
goal is to select $k$ items adaptively to maximize their expected
value.  The task of maximizing expected value subject to more complex
constraints, such as matroid constraints and intersections of matroid
constraints, is considered in the work of~\citet{golovin11matroid_arxiv}.
Before stating our result, we require the following definition.

\begin{definition}[Policy Truncation] \label{def:policy-truncation}
  For a policy $\policy$, define the \emph{level-$k$-truncation}
  $\prune{\policy}{k}$ of $\policy$ to be the policy obtained by
  running $\policy$ until it terminates or until it selects $k$ items,
  and then terminating.  Formally,
  $\dom(\prune{\policy}{k}) = \set{\prlz \in \dom(\policy)\  : \
    |\prlz| < k} $, and $\prune{\policy}{k}(\prlz) = \policy(\prlz)$
  for all $\prlz \in \dom(\prune{\policy}{k})$.
\end{definition}

\looseness -1 We have the following result, which generalizes the
classic result of the work of~\citet{nemhauser78} that the greedy algorithm
achieves a $(1-1/e)$-approximation to the problem of
maximizing monotone submodular functions under a
cardinality constraint.  By setting $\ell = k$ and $\alpha = 1$ in
\thmref{thm:max-cover}, we see
that the greedy policy which selects $k$ items adaptively obtains at
least $(1-1/e)$ of the value of the optimal policy that selects $k$ items adaptively,
measured with respect to $\avgf$.
For a proof see
\thmref{thm:max-cover-with-costs} in
Appendix~\ref{sec:proofs-max-cover},
which generalizes \thmref{thm:max-cover} to nonuniform item costs.

\begin{theorem} \label{thm:max-cover}
Fix any $\alpha \ge 1$.
If $f$ is \term monotone and \term submodular with respect to the
distribution
$\rlzprior$, and $\policy$ is an
$\alpha$-approximate greedy policy, then for all policies $\policy^*$ and positive
integers $\ell$ and $k$,
\[
\avgf(\prune{\policy}{\ell}) > \paren{1 - e^{-\ell/\alpha k}}
\avgf(\prune{\policy^*}{k})
\mbox{.}
\]
In particular, with $\ell = k$ this implies any $\alpha$-approximate
greedy policy achieves a $\paren{1 - e^{-1/\alpha}}$ approximation to
the expected reward of the best
policy, if both are terminated after running for an equal number of steps.
\end{theorem}

If the greedy rule can be implemented only with small \emph{absolute} error rather than small \emph{relative} error, i.e., $\diff{\prlz}{\elem'} \ge  \max_{\elem}
\diff{\prlz}{\elem}-\varepsilon$, an argument similar to that used to
prove~\thmref{thm:max-cover}
shows that
\[
\avgf(\prune{\policy}{\ell}) \ge \paren{1 - e^{-\ell/k}}
\avgf(\prune{\policy^*}{k}) - \ell\varepsilon
\mbox{.}
\]
This is important, since small absolute error can always be achieved
(with high probability) whenever $f$ can be evaluated efficiently, and
sampling $\rlzmass{\rlz\mid\prlz}$ is efficient. In this case, we can approximate
\[
\diff{\prlz}{\elem}
\approx \frac{1}{N}\sum_{i=1}^{N}\bigl[f(\dom(\prlz) \cup \set{\elem}, \rlz_{i}) - f(\dom(\prlz), \rlz_{i})\bigr]\mbox{,}
\]
where $\rlz_{i}$ are sampled i.i.d.~from $\rlzmass{\rlz\mid\prlz}$.

\subsubsection{Data Dependent Bounds} For the maximum coverage objective, \term submodular functions have
another attractive feature: they allow us to obtain data dependent
bounds on the optimum, in a manner similar to the bounds for the non-adaptive case \citep{minoux78}.
Consider the non-adaptive problem of maximizing a monotone submodular
function $f:2^{\groundsubset} \to \NonNegativeReals$ subject to the
constraint $|A| \le k$.  Let $A^*$ be an optimal solution,
 and fix any $A \subseteq \groundset$.  Then
\begin{equation}
  \label{eq:data-dependent-bound}
   f(A^*) \le f(A) + \max_{B : |B| \le k} \sum_{e \in B} \paren{ f(A
     \cup \set{e}) - f(A)  }
\end{equation}
because setting $B = A^*$ we have
$f(A^*) \le f(A \cup B) \le  f(A) + \sum_{e \in B} \paren{ f(A
     \cup \set{e}) - f(A)  }$.
Note that unlike the original objective, we can easily compute
$\max_{B : |B| \le k} \sum_{e \in B} \paren{ f(A \cup \set{e}) - f(A)
} $ by computing $\delta(e) := f(A \cup \set{e}) - f(A)$ for each
$e$, and summing the $k$ largest values.  Hence we can quickly compute
an upper bound on our distance from the optimal value, $f(A^*) - f(A)$. In practice, such data-dependent bounds can be much tighter than the problem-independent performance guarantees of \citet{nemhauser78} for the greedy algorithm \citep{leskovec07}. Further note that these bounds hold for \emph{any} set $A$, not just sets selected by the greedy algorithm.

These data dependent bounds have the following analogue for \term monotone
submodular functions.  See
Appendix~\ref{sec:costs-data-dependent-bounds} for a proof.

\begin{lemma}[The Adaptive Data Dependent Bound] \label{lem:rate-equation}
Suppose we have made observations $\prlz$ after
selecting $\dom(\prlz)$.  Let $\policy^*$ be any policy such that
$|\played{\policy^*}{\rlz}| \le k$ for all $\rlz$.
Then for \term monotone submodular $f$
\begin{equation}
  \label{eq:data-dependent-bound-adaptive}
\diff{\prlz}{\policy^*} \ \le \ \max_{A \subseteq \groundset, |A| \le
  k} \ \sum_{e \in A} \diff{\prlz}{e}.
\end{equation}
\end{lemma}
Thus, after running \emph{any} policy $\policy$, we can efficiently compute a bound on the additional benefit that the optimal solution $\policy^{*}$ could obtain beyond the reward of $\policy$. We do that by computing the conditional expected marginal benefits for all elements $e$, and summing the $k$ largest of them. Note that these bounds can be computed on the fly when running the greedy algorithm, in a similar manner as discussed by \citet{leskovec07} for the non-adaptive setting.

\subsection{The Min Cost Cover Objective} \label{sec:min-cost-cover}
Another natural objective is to minimize the number of items selected
while ensuring that a sufficient level of value is obtained.  This
leads to the \emph{Adaptive Stochastic Minimum Cost Coverage} problem
described in~\secref{sec:problem-statment}, namely
$\policy^{*}\in\argmin_{\policy} \acst{\policy}\text{ such that }
f(\played{\policy}{\rlz},\rlz)\geq \quota\text{ for all }\rlz$.  Recall that
$\acst{\policy}$ is the expected cost of $\policy$, which in the unit
cost case equals the expected number of items
selected by $\policy$, i.e., $\acst{\policy}:=\expctoverrlz{\rvrlz}{|\played{\policy}{\rvrlz}|}$.
If the objective is \term monotone submodular,
this is an adaptive version of the Minimum Submodular Cover problem
(described on line~(\ref{eq:detcover}) in
\secref{sec:submod-background}).  Recall that the greedy algorithm is known to give a
$(\ln(\quota)+1)$-approximation for Minimum Submodular Cover
assuming the coverage function is integer-valued in addition to being monotone submodular~\citep{wolsey82}.
Adaptive Stochastic Minimum Cost Coverage is also related
to the \emph{(Noisy) Interactive Submodular Set Cover} problem studied by
\citet{guillory10interactive,guillory2011-noisy-interactive-submod-cover},
which considers the worst-case setting
(i.e., there is no distribution over states; instead states are
realized in an adversarial manner).
Similar results for active learning have been proved
by~\citet{kosaraju99} and~\citet{dasgupta04}, as we discuss in more detail in~\secref{sec:active-learning}.

We assume throughout this section that there exists a quality
threshold $Q$ such that $f(\groundset, \rlz) = Q$ for all $\rlz$, and
for all $S \subseteq \groundset$ and all $\rlz$, $f(S, \rlz) \le Q$. Note that, as discussed in Section~\ref{sec:term-submodularity}, if we replace $f(S,\rlz)$ by a new function $g(S,\rlz)=\min(f(S,\rlz),Q')$ for some constant $Q'$, $g$ will be \term submodular if $f$ is. Thus, if $f(\groundset,\rlz)$ varies across realizations, we can instead use the greedy algorithm on the function truncated at some threshold $Q'\leq \min_{\rlz}f(\groundset,\rlz)$ achievable by all realizations.

In contrast to Adaptive Stochastic Maximization, for the coverage
problem additional subtleties arise.
In particular,  it is not enough
that a policy $\policy$ achieves value $Q$ for the true realization;
in order for $\policy$ to terminate, it also requires a proof of this
fact.  Formally, we require that $\policy$ \emph{covers} $f$:

\begin{definition}[Coverage]\label{def:coverage}
Let $\prlz = \prlz(\policy, \rlz)$ be the partial realization encoding all
 states observed during the execution of $\policy$ under realization
 $\rlz$.
Given $f:2^{\groundset} \times \outcomes^{\groundset} \to
\reals$, we say a policy
$\policy$ \emph{covers} $\rlz$ \emph{with respect to} $f$
if
$f(\dom(\prlz), \rlz') = f(\groundset, \rlz')$
for all $\rlz' \sim \prlz$.
We say that $\policy$ \emph{covers} $f$ if it covers every realization
with respect to $f$.
\end{definition}

Coverage is defined in such a way that upon terminating,
$\policy$ might not know which realization
is the true one, but has
guaranteed that it has achieved the maximum reward in every possible
case (i.e., for every realization consistent with its observations).  We obtain results for both the average and worst-case cost objectives.

\subsubsection{Minimizing the Average Cost}

Before presenting our approximation guarantee for the Adaptive Stochastic Minimum Cost Coverage,
we introduce a special class of instances, called \emph{\certifying} instances.
We make this distinction because
the greedy policy has stronger performance guarantees for
\certifying instances, and such instances arise naturally in
applications.
For example, the Stochastic Submodular Cover and Stochastic Set Cover
instances in \secref{sec:stochastic-set-cover}, the Adaptive Viral Marketing instances in
\secref{sec:viral-marketing}, and the Pool-Based Active Learning instances
in \secref{sec:active-learning} are all \certifying.

\begin{definition}[\Certifying Instances]
An instance of Adaptive Stochastic Minimum Cost Coverage is
\emph{\certifying} if whenever a policy achieves the maximum possible
value for the true realization it immediately has a proof of this
fact.
Formally, an instance $(f, \rlzprior)$ is \certifying if
for all $\rlz, \rlz'$, and $\prlz$ such that $\rlz \sim \prlz$ and $\rlz'
\sim \prlz$, we have
$f(\dom(\prlz), \rlz) = f(\groundset, \rlz)$ if and only if
$f(\dom(\prlz), \rlz') = f(\groundset, \rlz')$.
\end{definition}

One class of \certifying instances which commonly arise are those in which
$f(A, \rlz)$ depends only on the state of items in $A$, and in which
there is a uniform maximum amount of reward that can be obtained
across realizations.
Formally, we have the following observation.

\begin{proposition}
\label{prop:certifying}
Fix an instance $(f, \rlzprior)$.
If there exists $\quota$ such that $f(\groundset, \rlz) = \quota$ for
all $\rlz$ and there exists some $g:2^{\groundset \times \outcomes} \to
\NonNegativeReals$ such that $f(A, \rlz) = g\paren{ \set{(\elem,
    \rlz(\elem)) : \elem \in A}}$ for all $A$ and $\rlz$,
then $(f, \rlzprior)$ is \certifying.
\end{proposition}

\begin{proof}
Fix $\rlz, \rlz'$, and $\prlz$ such that $\rlz \sim \prlz$ and $\rlz'
\sim \prlz$.  Assuming the existence of $g$ and treating $\prlz$ as a relation, we have
$f(\dom(\prlz), \rlz) = g(\prlz) = f(\dom(\prlz), \rlz')$.
Hence $f(\dom(\prlz), \rlz) = \quota = f(\groundset, \rlz)$ if and only if
$f(\dom(\prlz), \rlz') = \quota = f(\groundset, \rlz') $.
\end{proof}

\noindent
For our results on minimum cost coverage, we also need a stronger monotonicity condition and a stronger submodularity condition:

\begin{definition}[Strong \Term Monotonicity] \label{def:SSM}
A function $f:2^{\groundset} \times \outcomes^{\groundset} \to
\reals$ is \emph{strongly adaptive monotone} with respect to
$\rlzprior$ if, informally ``selecting more items never hurts'' with respect to the expected reward.
Formally, for all $\prlz$, all $\elem \notin \dom(\prlz)$, and all
possible outcomes $\outcome \in \outcomes$ such that
$\prob{\rvrlz(\elem) = \outcome \  \mid \ \rvrlz \sim \prlz  } > 0$,
we require
\begin{equation} \label{eqn:strong-adapt-monotonicity}
\expctoverrlz{\rvrlz}{f(\dom(\prlz), \rvrlz) \ \mid \ \rvrlz \sim \prlz} \le
\expctoverrlz{\rvrlz}{f(\dom(\prlz) \cup \set{\elem}, \rvrlz) \ \mid \ \rvrlz \sim
  \prlz, \rvrlz(\elem) = \outcome}\mbox{.}
\end{equation}
\end{definition}
Strong \term monotonicity implies
\term monotonicity, as the latter means that ``selecting more items
never hurts in expectation,'' i.e.,
$$\expctoverrlz{\rvrlz}{f(\dom(\prlz), \rvrlz) \ \mid \ \rvrlz \sim \prlz} \le
\expctoverrlz{\rvrlz}{f(\dom(\prlz) \cup \set{\elem}, \rvrlz) \ \mid \ \rvrlz \sim
  \prlz}.$$

\noindent
To define strong adaptive submodularity, we first need the following extension
of $\diff{\prlz}{\elem}$:

\begin{definition}[Conditional Expected Marginal Benefit (Extended version)]
  Given partial realizations\\ $\prlz \subseteq \prlz'$, let
\begin{equation} \label{eqn:diff-ext}
\diff{\prlz ; \prlz'}{\elem} :=
 \expctoverrlz{\rlz}{f(\dom(\prlz) \cup \set{\elem}, \rvrlz) -
   f(\dom(\prlz), \rvrlz)\ \bigg| \ \rvrlz\sim\prlz'}
\end{equation}
\end{definition}

\begin{definition}[Strong Adaptive Submodularity] \label{def:strong-adapt-submod}
  A function $f:2^{\groundset} \times O^{\groundset} \to \NonNegativeReals$ is
  \emph{strongly adaptive submodular} with respect to distribution $\rlzprior$
  if it is adaptive submodular and moreover the expected marginal benefit of any
  fixed item does not increase as more items are selected and their states are
  observed, conditioned on the (item, observation) pairs.  Formally, $f$ is
  \term submodular w.r.t. $\rlzprior$ if for all $\prlz$ and $\prlz'$ such that
  $\prlz$ is a subrealization of $\prlz'$ (i.e., $\prlz \subseteq \prlz'$), and
  for all $\elem \in \groundset \setminus \dom(\prlz')$, we have
\begin{equation} \label{eqn:strong-adapt-submod}
\diff{\prlz ; \prlz'}{\elem} \ge \diff{\prlz'}{\elem}.
\end{equation}
In other words, conditioning on $\prlz'$, adding items
$\dom(\prlz') \setminus \dom(\prlz)$
cannot increase the expected marginal benefit of $\elem$.
\end{definition}

A sufficient condition for strong adaptive submodularity with respect to
$\rlzprior$ is that the function be adaptive submodular and
\emph{pointwise submodular}
(i.e., $f(A,\rlz)$ is submodular in $A$ for any fixed $\rlz$), as we prove in
Appendix~\ref{sec:proofs-min-cost-cover}.  It is worth noting that pointwise
submodularity is not sufficient to establish adaptive submodularity.
A simple counterexample is
$f(S, \rlz) = |\set{\elem : \elem \in S, \rlz(e) = 1}|$, with
$\rlzprior = 1/2$ if $\forall \elem \in \groundset,\ \rlz(\elem) = 1$ and
$\rlzprior = 1/2$ if $\forall \elem \in \groundset,\ \rlz(\elem) = 0$.
In that case, $\diff{\emptyset}{\elem} = 1/2$
yet $\diff{\set{(\elem', 1)}}{\elem} = 1$ for any $\elem' \neq \elem$.


We now state our main result for the  average case cost $\acst{\pi}$:

\begin{theorem} \label{thm:min-set-cover-avg-generalized}
Suppose $f:2^{\groundset} \times \outcomes^{\groundset} \to
\NonNegativeReals$ is strongly adaptive submodular and strongly adaptive
monotone with respect to $\rlzprior$ and there exists $Q$ such that
$f(\groundset, \rlz) = Q$ for all $\rlz$.
Let $\eta$ be any value such that
$f(S, \rlz) > Q - \eta$ implies $f(S, \rlz) = Q$ for all $S$ and
$\rlz$.
Let $\delta = \min_{\rlz} \rlzprior$ be the minimum probability of
any realization.
Let $\policycover$ be an optimal policy
minimizing the expected number of items selected
to guarantee every realization is covered.
Let $\policy$ be an $\alpha$-approximate
greedy policy with respect to the item costs.
Then in general
\[
\acst{\policy} \le  \alpha\,
\acst{\policycover}\paren{\ln \paren{\frac{Q}{\delta \eta}} + 1}^2
\]
and for \certifying instances
\[
\acst{\policy} \le  \alpha\,
\acst{\policycover}\paren{\ln \paren{\frac{Q}{\eta}} + 1}^2
\mbox{.}
\]
Note that if $\range(f) \subset \integers$, then $\eta = 1$ is a valid
choice, so for general and
\certifying instances we have $\acst{\policy} \le  \alpha\,\acst{\policycover}\paren{\ln (Q/\delta) + 1}^2$
and $\acst{\policy} \le
\alpha\,\acst{\policycover}\paren{\ln (Q) + 1}^2$, respectively.
%
\end{theorem}

\paragraph{Historical Note:}
An earlier version of \thmref{thm:min-set-cover-avg-generalized} claimed
logarithmic approximation factors
rather than the squared--logarithmic factors
present here. Unfortunately, the proof was flawed as pointed out by~\citet{nan17}.
Determining whether the logarithmic bounds hold remains an interesting open
problem.
In particular, it remains open whether
$\acst{\policy} \le  \alpha\,\acst{\policycover}\paren{\ln \paren{\frac{Q}{\delta \eta}} + 1}$ for general instances and $\acst{\policy} \le  \alpha\,\acst{\policycover}\paren{\ln \paren{\frac{Q}{\eta}} + 1}$ for \certifying instances under the conditions specified by \thmref{thm:min-set-cover-avg-generalized}. It also remains open whether the strong adaptive submodularity condition is required.

\subsubsection{Minimizing the Worst-Case Cost}

For the worst-case cost $\wcst{\policy} := \max_{\rlz}{|\played{\policy}{\rlz}|}$, strong \term monotonicity and strong submodularity are not required;
\term monotonicity and \term submodularity suffice.  We obtain the following result.

\begin{theorem} \label{thm:min-set-cover-wc-generalized}
Suppose $f:2^{\groundset} \times \outcomes^{\groundset} \to
\NonNegativeReals$
is \term monotone and \term submodular
with respect to $\rlzprior$, and
let $\eta$ be any value such that
$f(S, \rlz) > f(\groundset, \rlz) - \eta$ implies $f(S, \rlz) =
f(\groundset, \rlz)$ for all $S$ and $\rlz$.
Let $\delta = \min_{\rlz} \rlzprior$ be the minimum probability of
any realization.
Let $\policy_{wc}^{*}$ be the optimal policy
minimizing the worst-case number of queries
to guarantee every realization is covered.  Let $\policy$ be an $\alpha$-approximate
greedy policy.
Finally, let $Q := \expctoverrlz{\rlz}{f(\groundset, \rlz)}$ be the
maximum possible expected reward.
Then
\[
 \wcst{\policy} \le \alpha \, \wcst{\policy_{wc}^*}\,
 \paren{\ln \paren{\frac{Q}{\delta \eta}} +1}
\mbox{.}
\]
\end{theorem}
\noindent
The proofs of
Theorems~\ref{thm:min-set-cover-avg-generalized}
and~\ref{thm:min-set-cover-wc-generalized} are given in
Appendix~\ref{sec:proofs-min-cost-cover}.

Thus, even though \term submodularity is defined w.r.t.~a particular
distribution, perhaps surprisingly, the adaptive greedy algorithm is
competitive even in the case of  adversarially chosen realizations,
against a policy optimized to minimize the worst-case
cost. \thmref{thm:min-set-cover-wc-generalized} therefore suggests
that if we do not have a strong prior, we can obtain the strongest
guarantees if we choose a distribution that is ``as uniform as
possible'' (i.e., maximizes $\delta$) while still guaranteeing
adaptive submodularity.

\subsubsection{Discussion} Note that the approximation factor for \certifying instances
in~\thmref{thm:min-set-cover-wc-generalized}
reduces to the $(\ln(Q)+1)$-approximation
guarantee for the greedy algorithm for Set Cover instances with $Q$
elements, in the case of a deterministic distribution $\rlzprior$.
Moreover,  with a deterministic distribution $\rlzprior$
there is no distinction between average-case and
worst-case cost.
Hence, an immediate corollary of the result of
\citet{feige98threshold} mentioned in \secref{sec:term-submodularity}
is that for every constant $\epsilon > 0$
there is
no polynomial time
$(1-\epsilon)\ln \paren{Q/\eta}$ approximation algorithm for
\certifying instances of
Adaptive Stochastic Min Cost Cover, under either the $\acst{\cdot}$ or
the $\wcst{\cdot}$ objective,
unless $\NP\subseteq
\text{DTIME}(n^{\cO(\log\log n)})$.
It remains open to determine whether or not
Adaptive Stochastic Min Cost Cover with the worst-case cost objective
admits a $\ln \paren{Q/\eta}+1$ approximation for \certifying instances
via a polynomial time algorithm,
and in particular whether the greedy policy has such an approximation guarantee.
However, in \lemref{lem:approx-hardness-for-c_avg} we show that Feige's result also implies
there is no $(1-\epsilon)\ln \paren{Q/\delta\eta}$ polynomial time approximation
algorithm for general (non self-certifying) instances of
Adaptive Stochastic Min Cost Cover under either objective, unless $\NP\subseteq
\text{DTIME}(n^{\cO(\log\log n)})$.  In that sense,
\thmref{thm:min-set-cover-wc-generalized} is best-possible
and \thmref{thm:min-set-cover-avg-generalized} cannot be improved by more than
a logarithmic factor and under
reasonable complexity-theoretic assumptions.
%

%
\subsection{The Min-Sum Cover Objective} \label{sec:min-sum-cover}
\newcommand{\msgs}{U} %
\newcommand{\mse}{u} %

Yet another natural objective is the \emph{min-sum} objective, in
which an unrealized reward of $x$ incurs a cost of $x$ in each time
step, and the goal is to minimize the total cost incurred.
\subsubsection{Background on the Non-adaptive Min-Sum Cover Problem}
In the non-adaptive setting, perhaps the simplest form of a coverage
problem with this objective is the \emph{Min-Sum Set Cover}
problem~\citep{feige04} in which the input is a set system $(\msgs, \cS)$, the output
is a permutation of the sets $\tuple{S_1, S_2, \ldots, S_m}$, and the
goal is to minimize the sum of element \emph{coverage times}, where the
coverage time of $\mse$ is the index of the first set that contains
it (e.g., it is $j$ if $\mse \in S_j$ and $\mse \notin S_{i}$ for all $i <
j$).  In this problem and its generalizations
the min-sum objective is useful in modeling processing costs in
certain applications, for example in ordering diagnostic tests to
identify a disease cheaply~\citep{kaplan05}, in ordering multiple filters to be applied to
database records while processing a query~\citep{munagala05}, or in ordering multiple
heuristics to run on boolean satisfiability instances as a means to
solve them faster in practice~\citep{streeter08}.
A particularly expressive generalization of min-sum set cover has been
studied under the names \emph{Min-Sum Submodular Cover}~\citep{streeter08} and
$L_1$-\emph{Submodular Set Cover}~\citep{golovin08}.
The former paper extends the greedy algorithm to a natural online variant of the
problem, while the latter studies a parameterized family of
$L_p$-\emph{Submodular Set Cover} problems in which the objective is
analogous to minimizing the $L_p$ norm of the coverage times for
Min-Sum Set Cover instances.
In the Min-Sum Submodular Cover problem, there is a monotone submodular function $f:2^{\groundset}
\to \NonNegativeReals$ defining the reward obtained from a collection
of elements\footnote{To encode Min-Sum Set Cover instance
  $(\msgs, \cS)$, let $\groundset := \cS$ and $f(A) := |\cup_{e \in A} e|$, where each $e \in \groundset$ is a subset of
elements in $\msgs$.}.
There is an integral cost $c(e)$ for each element, and the output is a
sequence of all of the elements $\sequence = \tuple{e_1, e_2, \ldots, e_n}$.
For each $t \in \NonNegativeReals$, we define the set of elements in
the sequence $\sequence$ within a budget of $t$:
\[
\prune{\sequence}{t} := \set{e_i : \sum_{j \le i} c(e_j) \le t }.
\]
The cost we wish to minimize is then
\begin{equation}
  \label{eqn:min-sum-obj1}
  \costminsum{\sequence} := \sum_{t = 0}^{\infty} \paren{f(\groundset) - f(\prune{\sequence}{t})}.
\end{equation}
\citet{feige04} proved that for Min-Sum Set cover, the greedy
algorithm achieves a $4$-approximation to the minimum cost, and also
that this is optimal in the sense that no polynomial time algorithm can achieve a
$(4-\epsilon)$-approximation, for any $\epsilon > 0$, unless $\P = \NP$.
Interestingly, the greedy algorithm also achieves a $4$-approximation
for the more general Min-Sum Submodular Cover problem as
well~\citep{streeter08,golovin08}.

\subsubsection{The Adaptive Stochastic Min-Sum Cover Problem} In this
article, we extend the result of \citet{streeter08} and \citet{golovin08} to an
adaptive version of Min-Sum Submodular Cover.
For clarity's sake we will consider the unit-cost case here (i.e., $c(e) = 1$ for all
$e$);  we show how to extend \term submodularity to handle general
costs in \appendixA.
In the adaptive version of the problem, $\prune{\policy}{t}$ plays the role of
$\prune{\sequence}{t}$, and $\avgf$ plays the role of $f$.  The goal is to
find a policy $\policy$ minimizing
\begin{equation}
  \label{eqn:min-sum-obj2}
  \costminsum{\policy} := \sum_{t =
    0}^{\infty} \paren{\expctoverrlz{\rvrlz}{f(\groundset, \rvrlz)} -
      \avgf(\prune{\policy}{t})} = \sum_{\rlz} \rlzprior \sum_{t =
    0}^{\infty} \paren{f(\groundset, \rlz) - f(\groundset(\prune{\policy}{t},
  \rlz), \rlz)}.
\end{equation}
We call this problem the \emph{Adaptive Stochastic Min-Sum Cover} problem.
The key difference between this objective and the minimum cost
cover objective is that here, the cost at each step is only the fractional
extent that we have not covered the true realization, whereas in the
minimum cost cover objective we are charged in full in each step until
we have completely covered the true realization (according to
\defref{def:coverage}).
We prove the following result for the Adaptive Stochastic Min-Sum Cover problem with arbitrary item costs in Appendix~\ref{sec:proofs-min-sum-cover}.

\begin{theorem}\label{thm:min-sum-set-cover}
Fix any $\alpha \ge 1$.
If $f$ is \term monotone and \term submodular with respect to the
distribution $\rlzprior$, $\policy$ is an
$\alpha$-approximate greedy policy with respect to the item costs, and $\policy^*$ is any policy,
then $\costminsum{\policy} \le 4 \alpha \, \costminsum{\policy^*}$.
\end{theorem}

\section{Application: Stochastic Submodular Maximization} 
\label{sec:stochastic-maximization}

As our first application, consider the sensor placement problem
introduced in \secref{sec:intro}.  Suppose we would like to monitor a
spatial phenomenon such as temperature in a building.  We discretize
the environment into a set $\groundset$ of locations.  We would like
to pick a subset $A\subseteq\groundset$ of $k$ locations that is most
``informative'', where we use a set function $\hat{f}(A)$ 
to quantify the informativeness of placement $A$. \citet{krause07nearoptimal} show that many natural objective functions (such as reduction in predictive uncertainty measured in terms of Shannon entropy with conditionally independent observations) are monotone submodular. 

Now consider the problem,
where the informativeness of a sensor is unknown before deployment (e.g., when deploying cameras for surveillance, the location of objects and their associated occlusions may not be known in advance, or varying amounts of noise may reduce the sensing range). We can model this extension
by assigning a state $\rlz(\elem)\in\outcomes$ to each
possible location, indicating the extent to which a sensor placed at location
$\elem$ is working.  To quantify the value of a set of sensor deployments
under a realization $\rlz$ indicating to what extent the various
sensors are working, we first define $(\elem,{\outcome})$ for each $\elem \in \groundset$ and $\outcome
\in \outcomes$, which represents the placement of a sensor 
at location $\elem$ which is in state 
$\outcome$.  We then suppose there is a function 
$\hat{f}:2^{\groundset\times\outcomes} \to \NonNegativeReals$ which
quantifies the informativeness of a set of sensor deployments in arbitrary
states.  (Note $\hat{f}$ is a set function taking a set of (sensor
deployment, state) pairs as input.) 
The utility $f(A,\rlz)$ of placing sensors at the locations in $A$
under realization $\rlz$ is then
$$f(A,\rlz) := \hat{f}(\set{(\elem,\rlz(\elem)) : \elem \in A}).$$ 
\ignore{
 \begin{figure}%
 \centering 
 \includegraphics[height=4.5cm]{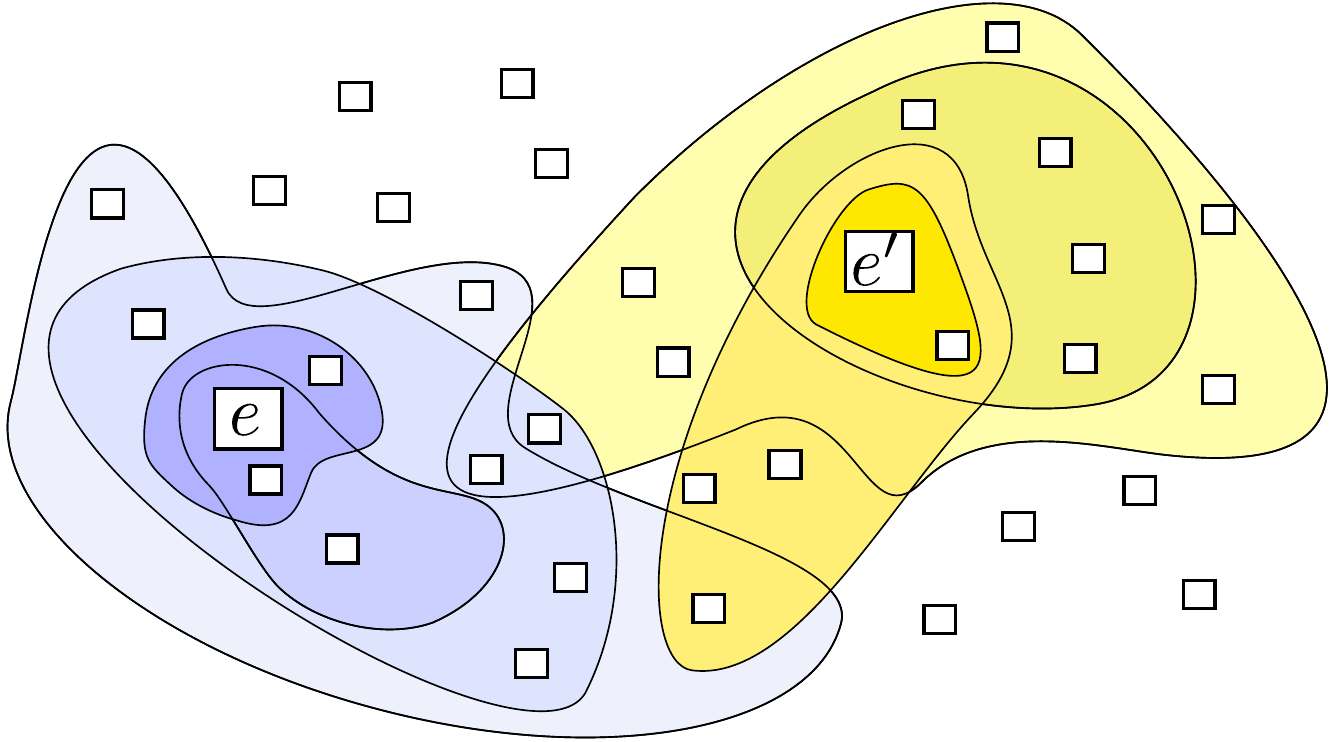}
 \caption{Illustration of part of a Stochastic Set Cover instance.  Shown are
   the supports of two distributions over sets, indexed by items $e$ (marked in blue) and $e'$ (yellow).   \label{fig:stocsetcover}}
 \end{figure}
}
\looseness -1 We aim to adaptively place $k$ sensors to maximize our expected utility.
We assume that sensor failures at each location are independent of each other, i.e., 
$\prob{\rvrlz = \rlz}=\prod_{\elem \in \groundset}\prob{\rvrlz(\elem) = \rlz(\elem)},$
where $\prob{\rlz(\elem)=\outcome}$ is the probability that a sensor
placed at location $\elem$ will be in state $\outcome$. 
\citet{AsadpourNS08} studied 
a special case of our problem,  
in which sensors either fail completely (in which case they
contribute no value at all) or work perfectly,
under the name \emph{\probname}. They proved that the adaptive greedy
algorithm obtains a 
$(1-1/e)$ approximation to the
optimal adaptive policy, provided $\hat{f}$ is monotone submodular.  
We extend their result to multiple types of failures
by showing that $f(A,\rlz)$ is \term submodular with respect to distribution $\rlzprior$ and then invoking Theorem~\ref{thm:max-cover}. \figref{fig:stocsetcover} illustrates an instance of Stochastic Submodular Maximization where $f(A,\rlz)$ is the cardinality of union of sets index by $A$ and parameterized by $\rlz$.

\begin{theorem}  \label{thm:WINE}
Fix a prior such that 
$\prob{\rvrlz = \rlz}=\prod_{\elem \in \groundset}\prob{\rvrlz(\elem) = \rlz(\elem)}$
and an integer $\budget$,
and let the objective function 
$\hat{f}:2^{\groundset\times\outcomes} \to \NonNegativeReals$ 
be monotone submodular.  
Let $\policy$ be any $\alpha$-approximate greedy policy attempting to maximize $f$, 
and let $\policy^*$ be any policy.  Then 
for all positive
integers $\ell$,
\[
\avgf(\prune{\policy}{\ell}) \ge \paren{1 - e^{-\ell/\alpha k}}
\avgf(\prune{\policy^*}{k}) 
\mbox{.}
\]
In particular, if $\policy$ is the greedy policy (i.e., $\alpha = 1$) and $\ell = k$, then
$\avgf(\prune{\policy}{k}) \ge \paren{1 - \frac{1}{e}} \avgf(\prune{\policy^*}{k}).$%
\end{theorem}

\begin{proof}
We prove Theorem~\ref{thm:WINE} by first proving $f$ is \term monotone
and \term submodular in
this model, and then applying Theorem~\ref{thm:max-cover}.
\Term monotonicity is readily proved after observing that 
$f(\cdot, \rlz)$ is monotone for each $\rlz$. 
Moving on to \term submodularity, fix any $\prlz, \prlz'$ such that 
$\prlz \subseteq \prlz'$ and any $\elem \notin \dom(\prlz')$.
We aim to show $\diff{\prlz'}{\elem} \le \diff{\prlz}{\elem}$.
Intuitively, this is clear, as $\diff{\prlz'}{\elem}$ is the expected marginal
benefit of adding $\elem$ to a larger base set than
is the case with $\diff{\prlz}{\elem}$, namely  $\dom(\prlz')$ as
compared to $\dom(\prlz)$, and the realizations are independent.
To prove it rigorously, 
we define a coupled distribution $\mu$ over pairs of realizations $\rlz \sim
\prlz$ and $\rlz' \sim \prlz'$ such that 
$\rlz(e') = \rlz'(e')$ for all $e' \notin \dom(\prlz')$.
Formally, 
$\mu(\rlz, \rlz') = \prod_{e \in \groundset \setminus
  \dom(\prlz)}\Pr{\rvrlz(e) = \rlz(e)}$
if  $\rlz \sim \prlz$, $\rlz' \sim \prlz'$, and $\rlz(e') =
\rlz'(e')$ for all $e' \notin \dom(\prlz')$; 
otherwise $\mu(\rlz, \rlz') = 0$.  
(Note that $\mu(\rlz, \rlz') > 0$ implies $\rlz(e') = \rlz'(e')$ for all $e' \in \dom(\prlz)$ as well,
since $\rlz \sim \prlz$, $\rlz' \sim \prlz'$, and  $\prlz \subseteq \prlz'$.)
Also note that $\rlzmass{\rlz \mid \prlz} = \sum_{\rlz'} \mu(\rlz, \rlz')$ and $\rlzmass{\rlz' \mid \prlz'} = \sum_{\rlz} \mu(\rlz, \rlz')$. 
Calculating 
$\diff{\prlz'}{\elem}$ and $\diff{\prlz}{\elem}$ using $\mu$, we see 
that for any $(\rlz, \rlz')$ in the support of $\mu$, 
\begin{eqnarray*}
  \label{eq:1mc}
f(\dom(\prlz') \cup \set{e}, \rlz') - f(\dom(\prlz'),
  \rlz')  & = & \hat{f}(\prlz' \cup \set{(\elem,\rlz'(\elem))}) -
  \hat{f}(\prlz'))\\
 & \le & \hat{f}(\prlz \cup \set{(\elem,\rlz(\elem))}) -
  \hat{f}(\prlz)) \\
 & = & f(\dom(\prlz) \cup \set{e}, \rlz) - f(\dom(\prlz),
  \rlz)
\end{eqnarray*}
from the submodularity of $\hat{f}$.
Hence 
\[
\begin{array}{lllll}
  \label{eq:2mc}
\diff{\prlz'}{\elem} & = & \sum_{(\rlz, \rlz')} \mu(\rlz, \rlz') \paren{
f(\dom(\prlz') \cup \set{\elem}, \rlz') - f(\dom(\prlz'),
  \rlz')} &  & \\[3mm]
  & \le & \sum_{(\rlz, \rlz')} \mu(\rlz, \rlz') \paren{ f(\dom(\prlz) \cup \set{\elem}, \rlz) - f(\dom(\prlz),
  \rlz)} & = & \diff{\prlz}{\elem}
\end{array}
\]
\ignore{
\begin{eqnarray*}
  \label{eq:2mc}
\diff{\prlz'}{\elem} & = & \sum_{(\rlz, \rlz')} \mu(\rlz, \rlz') \paren{
f(\dom(\prlz') \cup \set{e}, \rlz') - f(\dom(\prlz'),
  \rlz')} \\
  & \le & \sum_{(\rlz, \rlz')} \mu(\rlz, \rlz') \paren{ f(\dom(\prlz) \cup \set{e}, \rlz) - f(\dom(\prlz),
  \rlz)}\\
 & = & \diff{\prlz}{\elem}
\end{eqnarray*}
} %
which completes the proof. 
\end{proof}

\section{Application:  Stochastic Submodular Coverage}
\label{sec:stochastic-set-cover}

Suppose that instead of wishing to adaptively place $\budget$ unreliable sensors
to maximize
the utility of the information obtained, as discussed
in~\secref{sec:stochastic-maximization},
we have a quota on utility and wish to adaptively place the minimum number of
unreliable sensors to achieve this quota. This amounts to
a minimum-cost coverage version of the
Stochastic Submodular Maximization problem introduced
in~\secref{sec:stochastic-maximization}, which we call
\emph{Stochastic Submodular Coverage}.

As in~\secref{sec:stochastic-maximization}, in the Stochastic Submodular Coverage  problem we suppose there is a function
$\hat{f}:2^{\groundset\times\outcomes}\to \NonNegativeReals$ which quantifies
the utility of a set of sensors in arbitrary states.
Also, the states of each sensor are independent, so that
$\Pr{\rvrlz = \rlz} = \prod_{\elem \in \groundset} \Pr{\rvrlz(\elem) =  \rlz(\elem)}$.
The goal is to obtain a quota $Q$ of utility at minimum cost.
Thus, we define our objective as
$f(A,\rlz) := \min\set{Q,\hat{f}(\set{(\elem,\rlz(\elem)) : \elem \in A})}$, and want to
find a policy $\policy$ covering every realization
and minimizing
$\cavg{\policy} := \expct{|\played{\policy}{\rvrlz}|}$.
We additionally assume that this quota can always be obtained using
sufficiently many sensor placements; formally, this amounts to $f(E, \rlz) = Q$ for all $\rlz$.
We obtain the following result, whose proof we defer until the end of this section.

\begin{theorem} \label{thm:stochastic-submodular-cover}
Fix a prior with independent sensor states s.t.~$\Pr{\rvrlz = \rlz} = \prod_{\elem \in \groundset} \Pr{\rvrlz(\elem) =  \rlz(\elem)}$,
and let %
$\hat{f}:2^{\groundset\times\outcomes} \to \NonNegativeReals$
be a mon.~submodular function.
Fix $Q \in \NonNegativeReals$ s.t.
$f(A, \rlz) := \min \paren{Q,\ \hat{f}(\set{(\elem,\rlz(\elem)) : \elem \in A})}$
satisfies  $f(E, \rlz) = Q$ for all $\rlz$.
Let $\eta$ be any value such that
$f(S, \rlz) > Q - \eta$ implies $f(S, \rlz) = Q$ for all $S$ and $\rlz$.
Finally, let $\policy$ be an $\alpha$-approximate greedy policy for maximizing
$f$, and let $\policy^*$ be any policy.  Then
\[
\acst{\policy} \le  \alpha\,
\acst{\policy^*}\paren{\ln \paren{\frac{Q}{\eta}} + 1}^2 \mbox{.}
\]
\end{theorem}

\subsection{A Special Case: The Stochastic Set Coverage Problem}
The Stochastic Submodular Coverage problem is a generalization of the
\emph{Stochastic Set Coverage} problem~\citep{goemans06stochastic}.
In Stochastic Set Coverage the underlying
submodular objective $\hat{f}$ is the number of elements covered in
some input set system.  In other words, there is a ground set $U$ of
$n$ elements to be covered, and items $\groundset$ such that each item
$\elem$ is associated with a distribution over subsets of $U$.  When
an item is selected, a set is sampled from its distribution, as
illustrated in \figref{fig:stocsetcover}.  The
problem is to adaptively select items until all elements of $U$
are covered by sampled sets,
while minimizing the expected number of items selected.
Like us,
\citeauthor{goemans06stochastic} also assume that
the subsets are sampled independently for each item,
and every element of $U$ can
be covered in every realization, so that $f(\groundset, \rlz) =
|U|$ for all $\rlz$.

 \begin{figure}%
 \centering
 \includegraphics[height=4.5cm]{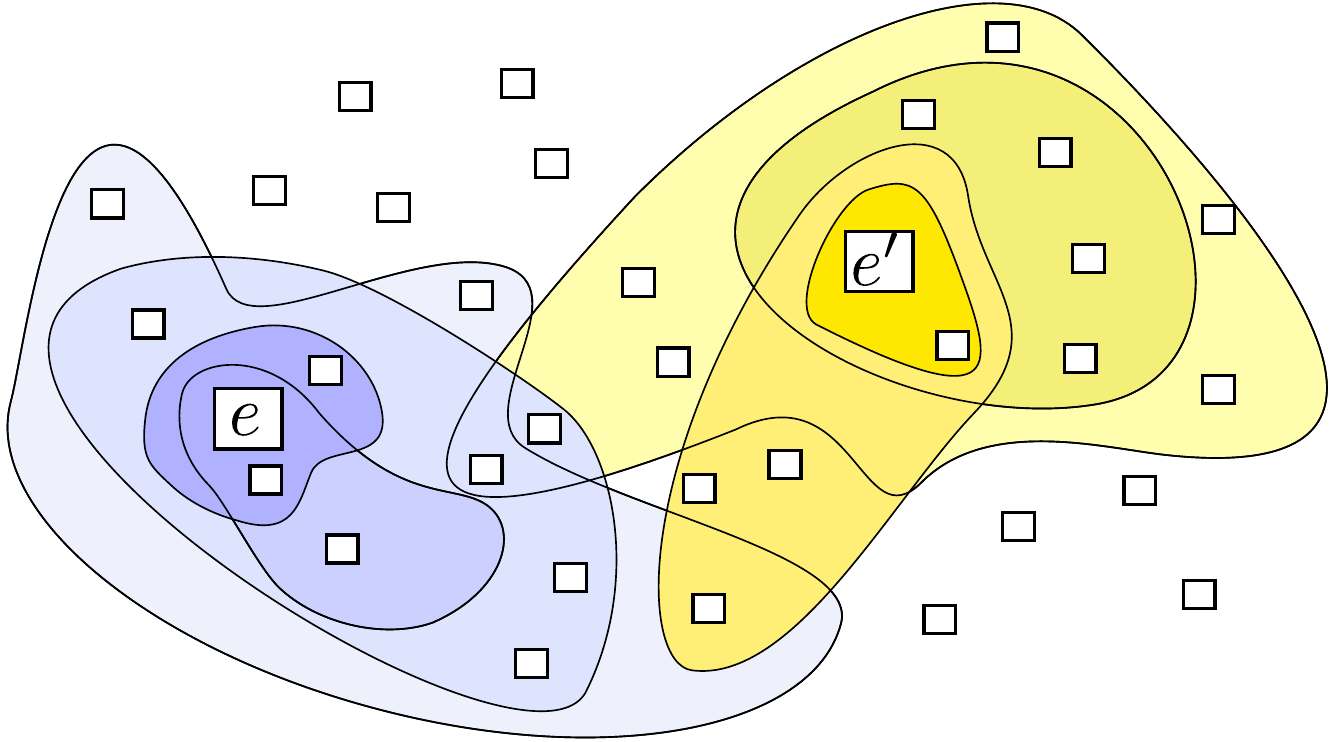}
 \caption{Illustration of part of a Stochastic Set Cover instance.  Shown are
   the supports of two distributions over sets, indexed by items $e$ (marked in blue) and $e'$ (yellow).   \label{fig:stocsetcover}}
 \end{figure}

\citeauthor{goemans06stochastic} primarily
investigated the adaptivity gap (quantifying how much adaptive policies can outperform non-adaptive policies) of Stochastic Set Coverage, for variants in which
items can be repeatedly selected or not, and prove adaptivity gaps of
$\Theta(\log n)$ in the former case, and between $\Omega(n)$ and
$\cO(n^2)$ in the latter.  They also provide an $n$-approximation
algorithm.
More recently, \citet{liu08near} considered a special case of
Stochastic Set Coverage in which each item may be in one of two
states.
They were motivated by a streaming database problem, in which a
collection of queries sharing common filters must all be evaluated on
a stream element.  They transform the problem to a Stochastic Set
Coverage instance in which (filter, query) pairs are to be covered by filter
evaluations;  which pairs are covered by a filter depends on the
(binary) outcome of evaluating it on the stream element.
The resulting instances satisfy the assumption that
every element of $U$ can be covered in every realization.
They study, among other algorithms, the adaptive greedy algorithm
specialized to this setting, and show that if
the subsets are sampled independently for each item, so that
$\prob{\rvrlz = \rlz} = \prod_{\elem} \prob{\rvrlz(\elem) = \rlz(\elem)}$,
then it is an $\mathcal{H}_{n} := \sum_{x=1}^{n} \frac{1}{x}$
approximation.
(Recall $\ln(n) \le \mathcal{H}_{n} \le \ln(n)+1$ for all $n \ge 1$.)
Moreover,  \citeauthor{liu08near} report that it
empirically outperforms a number of other algorithms in their experiments.

The \term submodularity framework allows us to prove approximate results
for richer item distributions over subsets of $U$ than considered by~\citet{liu08near} as a corollary of \thmref{thm:stochastic-submodular-cover}.
Specifically, we obtain a $(\ln(n) + 1)^2$-approximation for the Stochastic Set Coverage problem with arbitrarily many outcomes for each stochastic set,
where $n := |U|$.

We model the Stochastic Set Coverage problem by letting $\rlz(\elem)
\subseteq U$ indicate the random set sampled from $\elem$'s
distribution.  Since the sampled sets are independent we have
$\prob{\rvrlz = \rlz} = \prod_{\elem} \prob{\rvrlz(\elem) = \rlz(\elem)}$.  For any $A \subseteq \groundset$ let
$f(A, \rlz) := |\cup_{\elem \in A} \rlz(\elem) |$ be the number of elements of $U$ covered by the sets sampled from items in $A$.
As in the previous work mentioned above, we assume $f(\groundset,
\rlz) = n$ for all $\rlz$.  Therefore we may set $Q = n$.  Since the
range of $f$ includes only integers,
we may set $\eta = 1$.  Applying~\thmref{thm:stochastic-submodular-cover} then yields the following result.

\begin{corollary} \label{thm:stochastic-set-cover}
The adaptive greedy algorithm  achieves a $(\ln(n) + 1)^2$-approximation for Stochastic Set Coverage, where $n := |U|$ is the size of the ground set.
\end{corollary}

\noindent
We now provide the proof of \thmref{thm:stochastic-submodular-cover}.\\

\begin{proofof}{\thmref{thm:stochastic-submodular-cover}}
We will ultimately prove~\thmref{thm:stochastic-submodular-cover} by applying the bound from
\thmref{thm:min-set-cover-avg-generalized} for Stochastic Submodular Cover
instances.

The proof mostly consists of justifying this application.
Without loss of generality we may assume
$\hat{f}$ is truncated at $Q$, otherwise we may use
$\hat{g}(S) = \min \set{Q, \hat{f}(S)}$ in lieu of $\hat{f}$.
This removes the need to truncate $f$.
Since we established the \term submodularity of $f$ in the proof of
\thmref{thm:WINE}, and by assumption $f(E, \rlz) = Q$ for all $\rlz$,
to apply \thmref{thm:min-set-cover-avg-generalized} we need only show
that $f$ is strongly \term monotone and strongly adaptive submodular
and that the instances under consideration are \certifying.

We begin by showing the strong \term monotonicity of $f$.
Fix a partial realization $\prlz$, an item $\elem \notin \dom(\prlz)$
and a state $\outcome$.  Let
$\prlz' = \prlz \cup \set{(\elem,\outcome)}$.
Then treating $\prlz$ and $\prlz'$ as subsets of $\groundset \times \outcomes$, and
using the monotonicity of $\hat{f}$, we obtain
$$\progress{\prlz} = \hat{f}(\prlz) \le
\hat{f}(\prlz') \le \progress{\prlz'},$$
which is equivalent to the strong \term monotonicity condition.

Next we show the strong \term adaptive submodularity of $f$ by showing it is
pointwise submodular (having already proven adaptive submodularity for it).
This is clearly true, since for all $\rlz$,
$S \mapsto \hat{f}(\set{(e, \rlz(e)) : e \in S})$ is monotone submodular
by assumption.

Finally we prove that these instances are \certifying.
Consider any $\prlz$ and $\rlz, \rlz'$ consistent with $\prlz$.
Then
$$f(\dom(\prlz), \rlz) =  \hat{f}(\prlz) =  f(\dom(\prlz), \rlz').$$
Since $f(\groundset, \rlz) = f(\groundset, \rlz') = Q$ by assumption,
it follows that $f(\dom(\prlz), \rlz) = f(\groundset, \rlz)$ iff
$f(\dom(\prlz), \rlz') = f(\groundset, \rlz')$, so the instance is \certifying.

We have shown that $f$ and $\rlzprior$ satisfy the assumptions of
\thmref{thm:min-set-cover-avg-generalized} on this \certifying
instance.  Hence we may apply it
to obtain the claimed approximation guarantee.
\end{proofof}

\section{Application: Adaptive Viral Marketing} \label{sec:viral-marketing}

For our next application, consider the following scenario.
Suppose we would like to generate demand for a genuinely novel
product.  Potential customers do not realize how valuable the new
product will be to them, and conventional advertisements are
failing to convince them to try it.
In this case, we may try to spur demand by offering a special
promotional deal to a
select few people, and
hope that demand builds virally, propagating through the social
network as people recommend the product to their friends and
associates.
Supposing we know something about the structure of the
social networks people inhabit, and how ideas, innovation, and new
product adoption diffuse through them, this begs the
question: to which initial set of people should we offer the
promotional deal, in order to spur maximum demand for our product?

\looseness -1 This, broadly, is the viral marketing problem.
The same problem arises in the context of spreading technological,
cultural, and intellectual innovations, broadly construed.  In the
interest of unified terminology we
follow \citet{kempe03} and
talk of spreading \emph{influence} through the social network, where
we say people are \emph{active} if they have adopted the idea or
innovation in question, and \emph{inactive} otherwise, and that $a$ \emph{influences} $b$ if $a$
convinces $b$ to adopt the idea or innovation in question.

There are many ways to model the diffusion dynamics governing the
spread of influence in a social network.  We consider a basic and
well-studied model, the \emph{independent
  cascade model}, described in detail below.
For this model~\citet{kempe03}~obtain a very interesting result;
they show that the eventual spread of the influence ${f}$ (i.e., the
ultimate number of customers that demand the product) is a monotone
submodular function of the seed set $S$ of people initially selected.  This, in
conjunction with the results of~\citet{nemhauser78} implies that the
 greedy algorithm
obtains at least $\paren{1 - \frac1e}$ of the value of the best
feasible seed set of size at most $k$, i.e., $\argmax_{S: |S| \le k} {f}(S)$,
where we interpret $k$ as the budget for the promotional campaign.
Though~\citeauthor{kempe03} consider only the maximum coverage version of
the viral marketing problem, their result
in conjunction with that of~\citet{wolsey82} also implies that
the greedy algorithm will obtain a quota $Q$ of value at a cost of
at most $\ln(Q)+1$ times the cost of the optimal set
$\argmin_{S} \set{c(S) : f(S) \ge Q}$ if $f$ takes on only integral
values.

 \begin{figure}
 \centering
 \includegraphics[width=0.8\textwidth]{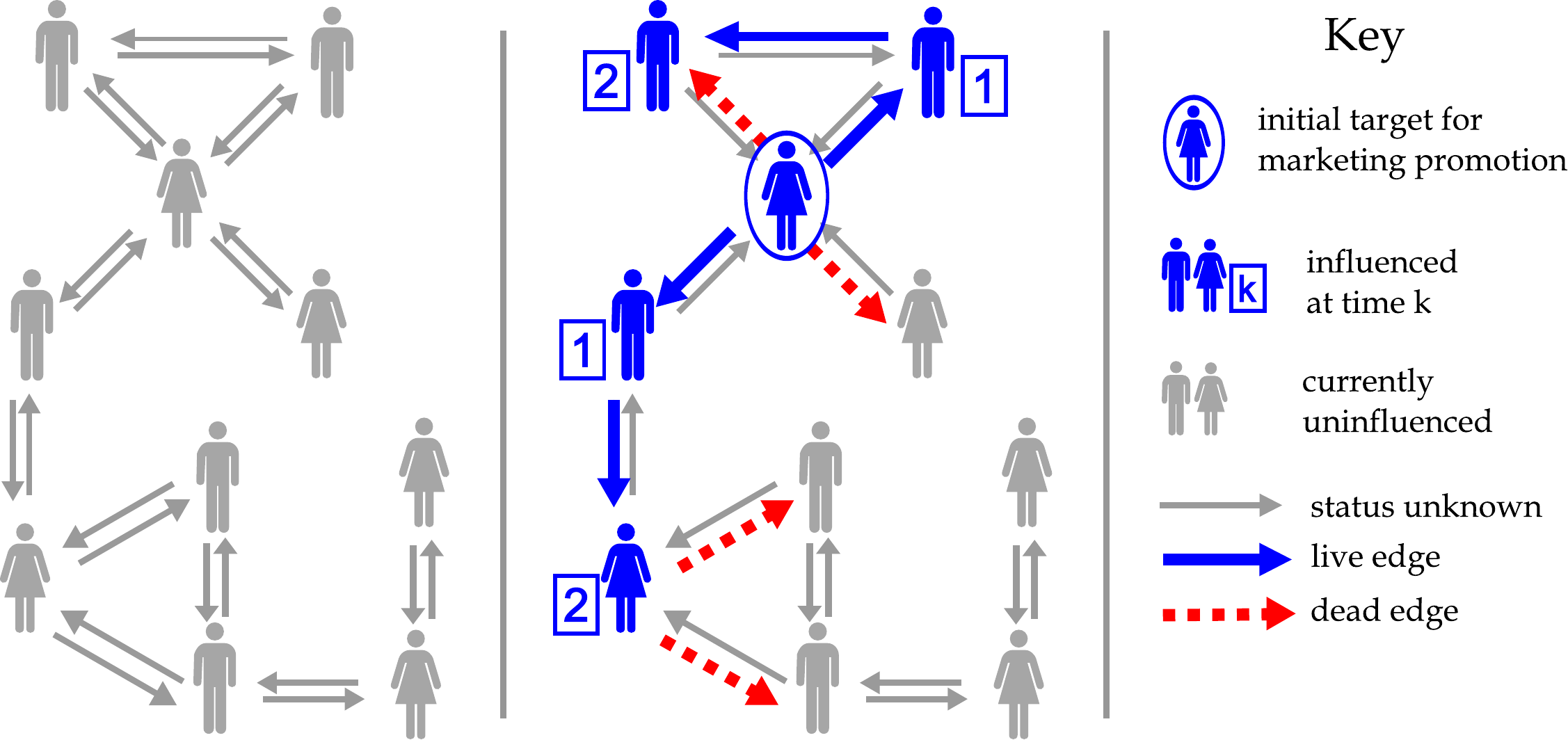}
 \caption{Illustration of the Adaptive Viral Marketing problem. Left:
   the underlying social network.
   Middle: the people influenced and
   the observations obtained after one person is selected.}
 \label{fig:viralmarketing}
 \end{figure}

\subsection{Adaptive Viral Marketing} The viral marketing problem has a very natural adaptive analog. Instead of selecting a fixed set of people in advance, we may select a person to offer the promotion to,
make some observations about the resulting spread of demand for our product, and repeat. See \figref{fig:viralmarketing} for an illustration.
In \secref{ssec:viral-max-cover},
we use the idea of \term submodularity to
obtain results analogous to
those of \citet{kempe03} in the adaptive setting.
Specifically, we show that the greedy policy obtains at least $\paren{1 - \frac1e}$ of the value of
the best \emph{policy}.  Moreover, we extend this result by
achieving that guarantee not only
for the case where our reward is simply the number of influenced
people, but also for any (nonnegative) monotone submodular function of the
\emph{set} of people influenced.  In \secref{ssec:viral-min-cost-cover}
we consider the minimum cost
cover objective, and show that the greedy policy obtains a squared logarithmic
approximation for it. To our knowledge, no approximation results for this adaptive variant of the viral marketing problem have been known.

\subsubsection{Independent Cascade Model}
In this model, the social network is a directed graph $G = (V, A)$
where each vertex in $V$ is a person, and each edge $(u, v)\in A$ has an
associated binary random variable $X_{uv}$ indicating if $u$ will
influence $v$.  That is,
$X_{uv} = 1$ if $u$ will influence
$v$ once it has been influenced, and $X_{uv} = 0$ otherwise.
The random variables $X_{uv}$ are independent, and have known means
$p_{uv} := \expct{X_{uv}}$.
We will call an edge $(u,v)$ with $X_{uv} = 1$ a \emph{live edge} and
an edge with $X_{uv} = 0$ a \emph{dead edge}.
When a node $u$ is activated, the edges $X_{uv}$ to each neighbor $v$
of $u$ are sampled, and $v$ is activated if $(u,v)$ is live.
Influence can then spread from $u$'s neighbors to their neighbors, and so on,
according to the same process.  Once active, nodes remain active
throughout the process, however~\citet{kempe03} show that this
assumption is without loss of generality, and can be removed.

\subsubsection{The Feedback Model}
In the Adaptive Viral Marketing problem under the
independent cascades model, the items correspond to people we can
activate by offering them the promotional deal.  How we define the
states $\rlz(u)$ depends on what information we obtain as a result
of activating $u$.  Given the nature of the diffusion process,
activating $u$ can have wide-ranging effects, so the state
$\rlz(u)$ has more to do with the state of the social network on the
whole than with $u$ in particular.
Specifically, we model
$\rlz(u)$ as a function $\rlz_u:A \to \set{0,1,?}$, where
$\rlz_u((u, v)) = 0$ means that activating $u$ has revealed that $(u,v)$ is dead,
$\rlz_u((u, v)) = 1$ means that activating $u$ has revealed that $(u,v)$ is live, and
$\rlz_u((u, v)) =\ ?$ means that activating $u$ has not revealed the
status of $(u,v)$ (i.e., the value of $X_{uv}$).  We require each
realization to be \emph{consistent} and \emph{complete}.
Consistency means that no edge should be declared both live and dead
by any two states.  That is, for all
$u, v \in V$ and $a \in A$, $(\rlz_u(a),\rlz_v(a)) \notin \set{(0,1), (1,0)}$.
Completeness means that the status of each edge is revealed by some
activation.  That is, for all
$a \in A$ there exists $u \in V$ such that $\rlz_u(a) \in \set{0,1}$.
A consistent and complete realization thus encodes $X_{uv}$ for each
edge $(u,v)$.  Let $\alive{\rlz}$ denote the live edges as encoded by
$\rlz$.
There are several candidates for which edge sets we
are allowed to observe when activating a node $u$.
Here we consider what we call the \emph{Full-Adoption Feedback Model}:
After activating $u$ we get to
  see the status (live or dead) of all edges exiting $v$, for all
  nodes $v$ reachable from $u$ via live edges (i.e., reachable from
  $u$ in $(V, \alive{\rlz})$, where $\rlz$ is the true realization.
We illustrate the full-adoption feedback model in~\figref{fig:viralmarketing}.

\subsubsection{The Objective Function}
In the simplest case, the reward for influencing a set $U \subseteq V$
of nodes is $\hat{f}(U) := |U|$.  \citet{kempe03} obtain an $\paren{1 -
  \frac1e}$-approximation for the slightly more general case in which
each node $\node$ has a weight $w_{\node}$ indicating its importance,
and the reward is $\hat{f}(U) := \sum_{\node \in U} w_{\node}$.  We
generalize this result further, to include arbitrary nonnegative
monotone submodular reward functions $\hat{f}$.
This allows us, for example, to encode a value associated with the
\emph{diversity} of the set of nodes influenced, such as the notion
that it is better to achieve $20\%$ market penetration in five
different (equally important) demographic segments than $100\%$ market penetration in one
and $0\%$ in the others.\\

\subsection{Guarantees for the Maximum Coverage Objective} \label{ssec:viral-max-cover}

\noindent
We are now ready to formally state our result for the maximum coverage
objective.

\begin{theorem} \label{thm:viral-marketing}
  The greedy policy $\greedypolicy$ obtains at least $\paren{1 - \frac{1}{e}}$ of the value of
the best \emph{policy} for the Adaptive Viral Marketing problem with
arbitrary monotone submodular reward functions,
in the independent cascade and
full-adoption feedback models
discussed above.
That is, if $\sigma(S, \rlz)$ is the set of all activated nodes
when $S$ is the seed set of activated nodes and $\rlz$
is the realization, $\hat{f}:2^V \to \NonNegativeReals$ is an
arbitrary monotone submodular function indicating the reward for
influencing a set, and the objective function is
$f(S, \rlz) := \hat{f}(\sigma(S, \rlz))$, then for all policies
$\policy$ and all $k \in \nats$ we have
$$\avgf(\prune{\greedypolicy}{k}) \ge \paren{1 -
  \frac{1}{e}}\avgf(\prune{\policy}{k}).$$
More generally, if $\policy$ is an $\alpha$-approximate greedy policy
then $\forall \ell \in \nats$, $\avgf(\prune{\policy}{\ell}) \ge \paren{1 - e^{-\ell/\alpha k}}
\avgf(\prune{\policy^*}{k}) $.
 \end{theorem}

\begin{proof}
Adaptive monotonicity follows immediately from the fact that $f(\cdot,\rlz)$ is monotonic for each $\rlz$.
It thus suffices to prove that $f$ is \term submodular with respect to the
probability distribution on realizations $\rlzprior$,
because then we can invoke
Theorem~\ref{thm:max-cover} to complete the proof.

\looseness -1 We will say we have \emph{observed} an edge $(u,v)$ if we know its
status, i.e., if it is live or dead.
Fix any $\prlz, \prlz'$ such that
$\prlz \subseteq \prlz'$ and any $v \notin \dom(\prlz')$.
We must show $\diff{\prlz'}{v} \le \diff{\prlz}{v}$.
To prove this rigorously,
we define a coupled distribution $\mu$ over pairs of realizations $\rlz \sim
\prlz$ and $\rlz' \sim \prlz'$.  Note that given the feedback model,
the realization $\rlz$ is a function
of the random variables $\set{X_{uw} : (u, w)
  \in A}$ indicating the status of each edge.  For conciseness we use
the notation $\Xvec = \set{X_{uw} : (u, w) \in A}$.
We  define $\mu$ implicitly in terms of a
joint distribution $\hat{\mu}$ on $\Xvec \times \Xvec'$,
where $\rlz = \rlz(\Xvec)$ and
$\rlz' = \rlz'(\Xvec')$ are the realizations
induced by the two distinct sets of random edge statuses, respectively.
Hence $\mu( \rlz(\Xvec ),  \rlz(\Xvec' )) = \hat{\mu}(\Xvec, \Xvec')$.
Next, let us say a partial realization $\prlz$ observes an edge $e$ if
some $w \in \dom(\prlz)$ has revealed its status as being live or
dead.
For edges $(u,w)$ observed by $\prlz$, the random variable
$X_{uw}$ is deterministically set to the status observed by $\prlz$.
Similarly, for edges $(u,w)$ observed by $\prlz'$, the random variable
$X'_{uw}$ is deterministically set to the status observed by $\prlz'$.
Note that since $\prlz \subseteq \prlz'$, the
state of all edges which are
observed by $\prlz$ are the same in $\rlz$ and $\rlz'$.
All $(\Xvec, \Xvec') \in \support(\hat{\mu})$ have these properties.
Additionally, we will construct $\hat{\mu}$ so that
the status of all edges which are
unobserved by both $\prlz'$ and $\prlz$ are the same in $\Xvec$ and $\Xvec'$, meaning
$X_{uw} = X'_{uw}$ for all such edges $(u,w)$, or else $\hat{\mu}(\Xvec,\Xvec') = 0$.

The above constraints leave us with the following degrees of freedom:
we may select $X_{uw}$ for all $(u,w) \in A$ which are unobserved by
$\prlz$.  We select them independently, such that $\expct{X_{uw}} =
p_{uw}$ as with the prior $\rlzprior$.  Hence for all $(\Xvec, \Xvec')$
satisfying the above constraints,
$$\hat{\mu}(\Xvec, \Xvec') = \prod_{(u,w) \text{ unobserved by }\prlz}
p_{uw}^{X_{uw} } \paren{1 - p_{uw}}^{1 - X_{uw}},$$
and otherwise $\hat{\mu}(\Xvec, \Xvec') = 0$.
Note that $\rlzmass{\rlz \mid \prlz} = \sum_{\rlz'} \mu(\rlz, \rlz')$ and $\rlzmass{\rlz' \mid \prlz'} = \sum_{\rlz} \mu(\rlz, \rlz')$.
We next claim that for all
$(\rlz, \rlz') \in \support(\mu)$
\begin{eqnarray}
  \label{eq:1vm}
f(\dom(\prlz') \cup \set{v}, \rlz') - f(\dom(\prlz'),
  \rlz')
 & \le & f(\dom(\prlz) \cup \set{v}, \rlz) - f(\dom(\prlz),
  \rlz).\quad
\end{eqnarray}
Recall $f(S, \rlz) := \hat{f}(\sigma(S, \rlz))$, where $\sigma(S, \rlz)$ is the set of all activated nodes
when $S$ is the seed set of activated nodes and $\rlz$
is the realization.
Let $B = \sigma(\dom(\prlz), \rlz)$ and $C = \sigma(\dom(\prlz) \cup
\set{v}, \rlz)$ denote the active nodes before and after selecting $v$
after $\dom(\prlz)$ under realizations
$\rlz$, and similarly define $B'$ and $C'$ with respect to $\prlz'$
and $\rlz'$.
Let $D := C \setminus B$, $D' := C' \setminus B'$.
Then \eqnref{eq:1vm} is equivalent to $\hat{f}(B' \cup D') - \hat{f}(B') \le
\hat{f}(B \cup D) - \hat{f}(B)$.
By the submodularity of $\hat{f}$, it suffices to show that $B
\subseteq B'$ and $D' \subseteq D$ to prove the above inequality,
which we will now do.

We start by proving $B \subseteq B'$.  Fix $w \in B$.  Then there
exists a path from some $u \in \dom(\prlz)$ to $w$ in
$(V, \alive{\rlz})$.  Moreover, every edge in this path is not only live
but also observed to be live,
by definition of the feedback model.  Since
$(\rlz, \rlz') \in \support(\mu)$, this implies that every edge
in this path is also live under $\rlz'$, as edges observed by $\prlz$
must have the same status under both $\rlz$ and $\rlz'$.
It follows that there is
a path from $u$ to $w$ in $(V, \alive{\rlz'})$.
Since $u$ is clearly also in  $\dom(\prlz')$, we conclude $w \in B'$,
hence $B \subseteq B'$.

Next we show $D' \subseteq D$.  Fix some $w \in D'$ and suppose by way of contradiction that $w
\notin D$.  Hence there exists a path $P$ from $v$ to $w$ in $(V,
\alive{\rlz'})$ but no such path exists in $(V, \alive{\rlz})$.  The
edges of $P$ are all live under $\rlz'$, and at least one must be
dead under $\rlz$.  Let $(u, u')$ be such an edge in $P$.  Because
the status of this edge differs in $\rlz$ and $\rlz'$, and
$(\rlz, \rlz') \in \support(\mu)$, it must be that $(u, u')$ is
observed by $\prlz'$ but not observed by $\prlz$.
Because it is observed by $\prlz'$, in our feedback model it must be that $u$ is active
after $\dom(\prlz')$ is selected, i.e., $u \in B'$.
However, this implies that all nodes reachable from $u$ via edges in
$P$ are also active after $\dom(\prlz')$ is selected, since all the
edges in $P$ are live.  Hence all such nodes, including $w$, are in
$B'$.  Since $D'$ and $B'$ are disjoint, this implies $w \notin D'$, a
contradiction.

Having proved~\eqnref{eq:1vm}, we now proceed to use it to show
$\diff{\prlz'}{v} \le \diff{\prlz}{v}$ as in
\secref{sec:stochastic-maximization}.
\[
\begin{array}{lllll}
  \label{eq:2vm}
\diff{\prlz'}{v} & = & \sum_{(\rlz, \rlz')} \mu(\rlz, \rlz') \paren{
f(\dom(\prlz') \cup \set{v}, \rlz') - f(\dom(\prlz'),
  \rlz')} &  & \\[3mm]
  & \le & \sum_{(\rlz, \rlz')} \mu(\rlz, \rlz') \paren{ f(\dom(\prlz) \cup \set{v}, \rlz) - f(\dom(\prlz),
  \rlz)} & = & \diff{\prlz}{v}
\end{array}
\]
which completes the proof.
\end{proof}

\subsubsection{Comparison with Stochastic Submodular Maximization}
It is worth contrasting the Adaptive Viral Marketing problem with the
Stochastic Submodular Maximization problem
of~\secref{sec:stochastic-maximization}.  In the latter problem, we
can think of the items as being
random independently distributed sets.
In Adaptive Viral Marketing by contrast, the random sets (of nodes
influenced when a fixed node is selected) depend on the random status
of the edges, and hence may be correlated through them.
Nevertheless, we can obtain the same $\paren{1 - \frac1e}$
approximation factor for both problems.\\

\ArxivOnly{
\paragraph{A Comment on the Myopic Feedback Model.}
In the conference version of this
article~\citep{golovin10colt}, we considered an alternate feedback
model called the \emph{myopic feedback} model, in which after
activating $v$ we see the status of all edges exiting $v$ in the
social network, i.e.,  $\partial_{+}(u) := \set{(u,v) : v \in V} \cap
A$.  We claimed that the objective $f$ as defined previously is \term
submodular in the independent cascade model with myopic feedback, and
hence the greedy policy obtains a $(1-\frac1e)$ approximation for it.
We hereby retract this claim, and furthermore give a counterexample demonstrating
that $f$ is not \term submodular under myopic feedback.

Consider a graph $G = (V, E)$ with vertices
$V := \set{u, v, w}$, and edges
$E := \set{(u,v), (v, w)}$.
The edge parameters are $p_{uv} = 1$ and $p_{vw} = 1-\epsilon$.
Let $\hat{f}(U) = |U|$ and construct $f$ from $\hat{f}$ accordingly.
We let $\prlz = \set{(u, \rlz_u)}$, where $\rlz_u((u,v)) = 1$ and
$\rlz_u((v,w)) = \ ?$.
Let $\prlz' = \set{(u, \rlz_u), (v, \rlz_v)}$ where $\rlz_v((v,w)) = 0$.
Clearly, $\prlz \subset \prlz'$.
Note $\diff{\prlz}{w} = \epsilon$, since the marginal benefit of
$w$ over $\dom(\prlz)$ is one if $(v,w)$ is dead, and zero if it
is live, and the former occurs with probability $\epsilon$.
In contrast, $\diff{\prlz'}{w} = 1$, since $\prlz'$ contains the
observation that $(v, w)$ is dead.
Hence $\diff{\prlz}{w} < \diff{\prlz'}{w} $, which violates \term
 submodularity.
However, we conjecture that the greedy policy still obtains a constant
factor approximation even in the myopic feedback model.

\daniel{Greedy policy probably gets an $\Omega(1)$ approx under myopic feedback.}
} %

\subsection{The Minimum Cost Cover Objective} \label{ssec:viral-min-cost-cover}

We may also wish to adaptively run our campaign until a certain level of market penetration has been achieved, e.g., a certain number of people have adopted the product. We can formalize this goal using the minimum cost cover objective.
For this objective, we have an instance of Adaptive
Stochastic Minimum Cost Cover, in which we are given a quota $Q \le \hat{f}(V)$ (quantifying the desired level of market penetration) and
we must adaptively select nodes to activate until the set of all active nodes $S$ satisfies
$\hat{f}(S) \ge Q$.  We obtain the following result.

\begin{theorem} \label{thm:viral-marketing-min-cost-cover}
Fix a monotone submodular function $\hat{f}:2^V \to
\NonNegativeReals$ indicating
the reward for
influencing a set, and a quota $Q \le \hat{f}(V)$.
Suppose the objective is
$f(S, \rlz) := \min\set{Q, \hat{f}(\sigma(S, \rlz))}$, where
 $\sigma(S, \rlz)$ is the set of all activated nodes
when $S$ is the seed set of activated nodes and $\rlz$
is the realization.
Let $\eta$ be any value such that
$\hat{f}(S) > Q - \eta$ implies $\hat{f}(S) \ge Q$ for all $S$.
Then any $\alpha$-approximate greedy policy $\policy$ on average costs at most
$\alpha\paren{\ln \paren{\frac{Q}{\eta}} + 1}^2$ times the average cost of
the best \emph{policy} obtaining $Q$ reward
for the Adaptive Viral Marketing problem
in the independent cascade model with
full-adoption feedback as
described above.
That is,
$\cavg{\policy} \le \alpha\paren{\ln \paren{\frac{Q}{\eta}} +
  1}^2 \cavg{\policy^*}$
for any $\policy^*$ that covers every realization.
\end{theorem}

\newcommand{\activenodes}[1]{\ensuremath{V^{+}\!\paren{#1} }}
\begin{proof}
 We prove \thmref{thm:viral-marketing-min-cost-cover} by recourse to
\thmref{thm:min-set-cover-avg-generalized}.
We have already established that $f$ is \term submodular, in the proof
of \thmref{thm:viral-marketing}.  It remains to show that $f$ is
strongly \term monotone and strongly adaptive submodular,
that these instances are \certifying, and
that $Q$ and $\eta$ equal the corresponding terms
in the statement of \thmref{thm:min-set-cover-avg-generalized}.

We start with strong \term monotonicity.  Fix $\prlz$,  $\elem
\notin \dom(\prlz)$, and $\outcome \in \outcomes$.  We must show
\begin{equation}
  \label{eq:viral-min-cost1}
  \expctoverrlz{\rvrlz}{f(\dom(\prlz), \rvrlz) \ \mid \ \rvrlz \sim \prlz} \le
\expctoverrlz{\rvrlz}{f(\dom(\prlz) \cup \set{\elem}, \rvrlz) \ \mid \ \rvrlz \sim
  \prlz, \rvrlz(\elem) = \outcome}.
\end{equation}
Let  $\activenodes{\prlz}$ denote the active nodes after selecting
$\dom(\prlz)$ and observing $\prlz$.
By definition of the full adoption feedback model,
$\activenodes{\prlz}$ consists of precisely
those nodes $v$ for which there exists a path $P_{uv}$ from some $u \in
\dom(\prlz)$ to $v$ via exclusively live edges.  The edges whose
status we observe consist of all edges exiting nodes in
$\activenodes{\prlz}$.
It follows that every path
from any $u \in \activenodes{\prlz}$ to any $v \in V \setminus
\activenodes{\prlz}$ contains at least one edge which is observed by
$\prlz$ to
be dead.
Hence, in every $\rlz \sim \prlz$, the set of nodes activated by
selecting $\dom(\prlz)$ is the same.
Therefore $\expctoverrlz{\rvrlz}{f(\dom(\prlz), \rvrlz) \ \mid \ \rvrlz \sim
  \prlz} = \hat{f}(\activenodes{\prlz})$.
Similarly, if we define $\prlz' := \prlz \cup \set{(\elem,
  \outcome)}$, then
$\expctoverrlz{\rvrlz}{f(\dom(\prlz) \cup \set{\elem}, \rvrlz) \ \mid \ \rvrlz \sim
  \prlz, \rvrlz(\elem) = \outcome} = \hat{f}(\activenodes{\prlz'})$.
Note that once activated, nodes never become inactive.
Hence, $\prlz \subseteq \prlz'$ implies
$\activenodes{\prlz} \subseteq \activenodes{\prlz'}$.
Since $\hat{f}$ is monotone by assumption, this means
$\hat{f}(\activenodes{\prlz}) \le \hat{f}(\activenodes{\prlz'})$
which implies \eqnref{eq:viral-min-cost1} and strong \term monotonicity.

Next we establish strong adaptive submodularity. Given that we have already
established adaptive submodularity, it is sufficient to also prove pointwise
submodularity. For a fixed realization $\rlz$ we have a set of live edges
$\set{(u,v) : X_{uv} = 1}$ which induce a set system in which $u$ covers
all nodes reachable from $u$ via live edges.
Let $S_u$ denote this set.
It is straightforward to verify that a monotone submodular function $\hat{f}$ on
nodes induces a monotone submodular function on sets of those nodes.
That is,
\[
g(A) := \hat{f}\paren{ \cup_{u \in A} S_u }
\]
is submodular whenever $\hat{f}$ is.
In particular, $g$ is submodular if for every $A$ and $B$ we have
$g(A) + g(B) \ge g(A \cap B) + g(A \cup B)$ however one can easily verify that
this set of constraints is a subset of the corresponding submodularity
constaints on $\hat{f}$.

Next we establish that these instances are \certifying.
Note that for every $\rlz$ we have
$f(V, \rlz) = \min \set{Q, \hat{f}(V)} = Q$.
From our earlier remarks, we know that
$f(\dom(\prlz), \rlz) = \hat{f}(\activenodes{\prlz})$ for every $\rlz
\sim \prlz$.
Hence for all $\prlz$ and $\rlz, \rlz'$ consistent with $\prlz$, we
have $f(\dom(\prlz), \rlz) = f(\dom(\prlz), \rlz')$ and so
$f(\dom(\prlz), \rlz) = Q$ if and only if $f(\dom(\prlz), \rlz') = Q$,
which proves that the instance is \certifying.

Finally we show that $Q$ and $\eta$
equal the corresponding terms
in the statement of \thmref{thm:min-set-cover-avg-generalized}.
As noted earlier, $f(V, \rlz) = Q$ for all
$\rlz$.  We defined $\eta$ as some value such that
$\hat{f}(S) > Q - \eta$ implies $\hat{f}(S) \ge Q$ for all $S$.
Since $\range(f) = \set{\min \set{Q, \hat{f}(S)} : S \subseteq V}$, it
follows that we cannot have
$f(S, \rlz) \in (Q - \eta, Q)$ for any $S$ and $\rlz$, so that $\eta$
satisfies the requirements of the corresponding term in
\thmref{thm:min-set-cover-avg-generalized}.  Hence we may apply
\thmref{thm:min-set-cover-avg-generalized} on this \certifying
instance with $Q$ and $\eta$ to obtain the claimed result.
\end{proof}

\ignore{
The fact that  the eventual spread of the influence ${f}$ in monotone
submodular~\citep{kempe03} also implies,
in conjunction with the results of~\cite{wolsey82}, that
the greedy algorithm will obtain a quota $Q$ of value at a cost of
at most $\ln(Q)+1$ times the cost of the optimal set
$\argmin_{S} \set{c(S) : f(S) \ge Q}$ if $f$ takes on only integral values.
} %

\daniel{Include remark on the counterexample for the myopic feedback
  model?}
\daniel{Include comment on ``batched'' adaptive selection (i.e.,
  select $b$ nodes to target, get feedback, repeat $k$ times)?}

\section{Application: Automated Diagnosis and Active Learning} \label{sec:active-learning}
An important problem in AI is automated diagnosis. For example, suppose we have different hypotheses about the state of a patient, and can run medical tests to rule out inconsistent hypotheses. The goal is to adaptively choose tests to infer the state of the patient as quickly as possible.

A similar problem arises in active learning. Obtaining labeled data to train a classifier is typically expensive, as it often involves asking an expert. In active learning (\cf~\citet{cohn96active,mccallum98}), the key idea is that some labels are more informative than others: labeling a few unlabeled examples can imply the labels of many other unlabeled examples, and thus the cost of obtaining the labels from an expert can be avoided. As is standard, we assume that we are given a set of hypotheses $\hypotheses$,
and a set of unlabeled data points $\data$ where each $x \in \data$
is independently drawn from some
distribution $\distrib$.  Let $\labels$ be the set of possible labels.
Classical learning theory yields \emph{probably approximately correct} (PAC) bounds, bounding the number $n$ of examples drawn i.i.d. from $\distrib$ needed to output a hypothesis $h$ that will have expected error at most $\eps$ with probability at least $1 - \delta$,
for some fixed $\eps, \delta > 0$.  That is, if $\target$ is the
target hypothesis (with zero error), and
$\error(h) := \probover{x \sim \distrib}{h(x) \neq \target(x)}$
is the error of $h$,
 we require
$\prob{\error(h)  \le \eps } \ge 1 - \delta$.
The latter probability is taken with respect to $\distrib(\data)$;  the learned
hypothesis $h$ and thus $\error(h)$ depend on it.
A key challenge in active learning is to avoid bias: actively selected examples are no longer i.i.d., and thus sample complexity bounds for passive learning no longer apply.  If one is not careful, active learning may require more samples than passive learning to achieve the same generalization error.
One natural approach to active learning that is guaranteed to perform at least as well as passive learning is \emph{pool-based active learning} \citep{mccallum98}: The idea is to draw $n$ \emph{unlabeled} examples i.i.d. However, instead of obtaining all labels, labels are adaptively requested until the labels of all unlabeled examples are implied by the obtained labels. Now we have obtained $n$ labeled examples drawn i.i.d., and classical PAC bounds still apply. The key question is how to request the labels for the pool to infer the remaining labels as quickly as possible.

In the case of binary labels (or test outcomes) $L = \set{-1, 1}$, various authors have considered greedy
policies which generalize binary
search~\citep{garey74,loveland85,arkin93,kosaraju99,dasgupta04,guillory09,nowak09}.
The simplest of these, called \emph{generalized binary search} (\gbs) or the
\emph{splitting algorithm}, works as follows.
Define the \emph{version
  space} \vs to be the set of hypotheses consistent with the observed
labels (here we assume that there is no label noise).
In the worst-case setting, \gbs selects a query $x \in
\data$ that minimizes $\left| \sum_{h \in \vs} h(x) \right|$.
In the Bayesian setting we assume we are given a prior $\prior$ over
hypotheses; in this case \gbs selects a query $x \in
\data$ that minimizes $\left| \sum_{h \in \vs} \prior(h) \cdot h(x) \right|$.  Intuitively
these policies myopically attempt to shrink a measure of the version space (i.e.,
the cardinality or the probability mass) as quickly as
possible.
The former provides an $\cO(\log |\hypotheses|)$-approximation for the
worst-case number of queries~\citep{arkin93}, and the latter provides an
 $\cO(\log \frac{1}{\min_{h} \prior(h)})$-approximation for the
expected number of queries~\citep{kosaraju99,dasgupta04} and a natural
generalization of GBS obtains the same guarantees with a larger set of
labels~\citep{guillory09}.
Kosaraju~\etal also prove that running GBS on a modified prior
$\prior'(h) \propto \max\set{\prior(h), 1/|\hypotheses|^{2}\log
  |\hypotheses|}$ is sufficient to obtain an  $\cO(\log |\hypotheses|)$-approximation.

 \begin{figure}
 \centering
 \includegraphics[width=0.6\textwidth]{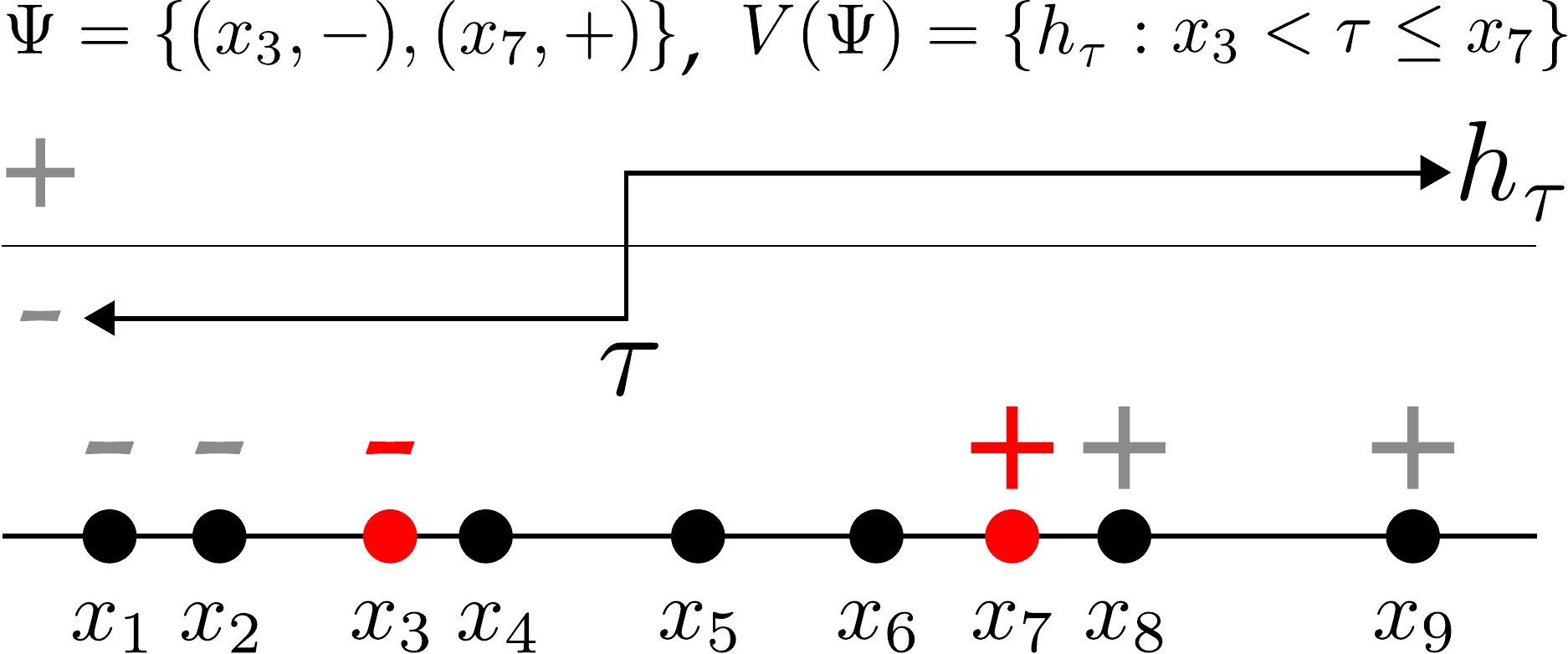}
 \caption{Illustration of the Active Learning problem, in the simple
   special case of one-dimensional data and
binary threshold hypotheses $\hypotheses = \set{h_\tau : \tau \in
  \reals}$, where $h_\tau(x) = 1$ if $x \ge \tau$ and $0$ otherwise.
 \label{fig:activelearning}}
 \end{figure}

Viewed from this perspective of the previous sections, shrinking the
version space amounts to ``covering'' all false hypotheses with stochastic
sets (i.e., queries), where query $x$ covers all hypotheses that
disagree with the target hypothesis $\target$ at $x$.  That is,
$x$ covers $\set{h : h(x) \neq \target(x)}$.  As
in~\secref{sec:viral-marketing}, these sets may be correlated in
complex ways determined by the set of possible hypotheses.
As we will show, the reduction in version space mass is \term
submodular, and this allows us to obtain
a new analysis of GBS using \term submodularity,
which has a weaker approximation guarantee than is optimal, but
is arguably more amenable to extensions
and generalizations than previous analyses.


\begin{theorem} \label{thm:optimal-decision-trees}
  In the Bayesian setting in which there is a prior $\prior$ on a
  finite set of hypotheses $\hypotheses$, the generalized binary search algorithm
  makes $\OPT \cdot \paren{ \ln \paren{\frac{1}{\min_{h} \prior(h)}} + 1}^2$ queries in
  expectation to identify a hypothesis drawn from $\prior$,
  where $\OPT$ is the minimum expected number of queries
  made by any policy.
  If $\min_{h} \prior(h)$ is sufficiently small,
  running the algorithm on a modified prior
$\prior'(h) \propto \max\set{\prior(h), 1/|\hypotheses|^{2}}$ improves the approximation factor to
$\cO\paren{\paren{\ln |\hypotheses|}^2}$.
\end{theorem}

\looseness -1

We devote the better part of the remainder of this section to the
proof of \thmref{thm:optimal-decision-trees}, which has several
components.
We first address the important special case of a uniform prior
over hypotheses, i.e., $\prior(h) = 1/|\hypotheses|$ for all $h \in
\hypotheses$, and then we reduce the case with a general prior to a
uniform prior.
We wish to appeal to Theorem~\ref{thm:min-set-cover-avg-generalized},
so we convert the problem into an Adaptive Stochastic Min Cost Cover problem.

\subsection{The Reduction to Adaptive Stochastic Min Cost Cover}
Define a realization $\rlz_h$ for each hypothesis $h \in \hypotheses$.
The ground set is $\groundset = \data$, and the
outcomes are binary; we define $\outcomes = \set{-1,1}$ instead of
using $\set{0,1}$ to be consistent with our earlier exposition.
For all $h\in \hypotheses$ we set $\rlz_h \equiv h$, meaning
$\rlz_h(x) = h(x)$ for all $x \in \data$.
To define the objective function, we first need some notation.
Given observed labels $\prlz \subset \data \times \outcomes$, let
$\vs(\prlz)$ denote the version space, i.e., the set of hypotheses for
which $h(x) = \prlz(x)$ for all $x \in \dom(\prlz) $.
See \figref{fig:activelearning} for an illustration of an active
learning problem in the case of
indicator hypotheses.
For a set of hypotheses $\vs$,  let $\prior(\vs) := \sum_{h \in \vs}
\prior(h)$ denote their total prior probability.
Finally, let $\prlz(S, h) = \set{(x, h(x)) \ : \ x \in S }$
be the function with domain $S$ that agrees with $h$ on $S$.
We define the objective function by
\[
f(S, \rlz_h) \ :=  \ 1 - \prior(\vs(\prlz(S, h)  ) ) \ = \
\prior\!\paren{ \set{h' \ : \ \exists x \in S, h'(x) \neq h(x)}}
\]
and use $\rlzmass{\rlz_h} = \prior(h) = 1/|\hypotheses|$ for all $h$.
Let $\policy^*$ be an optimal policy for this
Adaptive Stochastic Min Cost Cover instance.  Note that there is an exact correspondence between
policies for the original problem of finding the target hypothesis and
our problem of covering the true realization; $\target$ is identified as the target
hypothesis if and only if the version space is reduced to
$\set{\target}$ which occurs if and only if $\rlz_{\target}$ is covered.
Hence $\acst{\policy^*} = \OPT$.
Note that because we have assumed a uniform prior over hypotheses, we have
$f(\data, \rlz_h) = 1 -  1/|\hypotheses|$ for all $h$.
Furthermore, the instances are \certifying.

\begin{lemma} \label{lem:active-learning-certifying}
The instances described above are \certifying for arbitrary priors $\prior$.
\end{lemma}

\begin{proof}
Intuitively, theses instances are \certifying because to cover $\rlz_{\target}$ a
policy must identify $\rlz_{\target}$. More formally, these instances are \certifying because for any
$\rlz_{h}$ and $\prlz$ such that $\rlz_{h} \sim \prlz$, we have that
$f(\dom(\prlz), \rlz_{h}) = f(\data, \rlz_{h})$ implies
$\vs(\prlz) = \set{h}$.   This in turn means that $\rlz_{h}$ is the \emph{only}
realization consistent with $\prlz$, which trivially implies that any
realization $\rlz' \sim \prlz$ also has $f(\dom(\prlz), \rlz') =
f(\data, \rlz')$; hence the instance is \certifying.
\end{proof}

\subsection{The Uniform Prior}  We next prove that the instances
generated are adaptive submodular and strongly adaptive
monotone under a uniform prior.

\begin{lemma} \label{lem:active-learning-adapt-submod-uniform}
In the instances described above, $f$ is strongly adaptive
monotone and strongly \term submodular and with respect to a
uniform prior $\prior$.
\end{lemma}

\begin{proof}
Demonstrating strong adaptive monotonicity under a uniform prior
amounts to proving that adding labels
cannot grow the version space, which is clear in our model.
That is, each query $x$ eliminates some subset of
hypotheses, and as more queries are performed, the subset of
hypotheses eliminated by $x$ cannot grow.
Moving on to \term submodularity, consider the expected marginal contribution of $x$
under two partial realizations $\prlz, \prlz'$ where $\prlz$ is a
subrealization of $\prlz'$ (i.e., $\prlz \subset \prlz'$), and $x
\notin \dom(\prlz')$.
Let $\prlzxo$ be the partial
realization with domain $\dom(\prlz)  \cup \set{x}$ that agrees with
$\prlz$ on its domain, and maps $x$ to $\outcome$.
For each $\outcome \in \outcomes$,
let
$a_{\outcome} :=
\prior(\vs(\substitute{\prlz}{x}{\outcome}))$, $b_{\outcome} :=
\prior(\vs(\substitute{\prlz'}{x}{\outcome}))$.
Since a hypothesis eliminated from the version space cannot later
appear in the version space, we have
$a_{\outcome} \ge b_{\outcome}$ for all $\outcome$.
Next, note the expected reduction in version space mass (and hence the expected
marginal contribution) due to selecting $x$ given partial realization
$\prlz$ is
\begin{equation}
  \label{eq:active01}
  \diff{\prlz}{x} = \sum_{\outcome \in \outcomes} a_{\outcome} \cdot
  \prob{\rvrlz(x) \neq \outcome \ \mid \ \rvrlz \sim \prlz} =
  \sum_{\outcome} a_{\outcome} \paren{\frac{ \sum_{\outcome' \neq
        \outcome} a_{\outcome'} }{  \sum_{\outcome'} a_{\outcome'}  }
  } = \frac{ \sum_{\outcome \neq \outcome'} a_{\outcome} a_{\outcome'} }{  \sum_{\outcome'} a_{\outcome'}}.
\end{equation}
The corresponding quantity for $\prlz'$ has $b_{\outcome}$ substituted
for $a_{\outcome}$ in~\eqnref{eq:active01}, for each $\outcome \in \outcomes$.
To prove \term submodularity
we must show
$\diff{\prlz}{x} \ge \diff{\prlz'}{x}$ and to do so it suffices to show that $\partial
  \phi / \partial z_{\outcome} \ge 0$ for each $\outcome$ and
$\vec{z} \in \set{\vec{c} \in [0,1]^{\outcomes} : \sum_{\outcome}
  c_{\outcome} > 0 }$, where
$\phi(\vec{z}\,) :=  \paren{\sum_{\outcome \neq \outcome'} z_{\outcome}
  z_{\outcome'}} /  \paren{\sum_{\outcome'} z_{\outcome'}}$ has the
same functional form as the expression for $\diff{\prlz}{x}$ in~\eqnref{eq:active01}.
 This is because $\partial\phi / \partial
z_{\outcome} \ge 0$ for each $\outcome$ implies that growing the version space
in any manner cannot decrease the expected marginal benefit of query $x$, and
hence shrinking it in any manner cannot increase the expected marginal benefit
of $x$.
It is
indeed the case that $\partial\phi / \partial
z_{\outcome} \ge 0$ for each $\outcome$.  More specifically, it holds that
$$
\frac{\partial \phi}{\partial z_{a}} =
\frac{\sum_{b \neq a} z_{b}^2
  +\sum_{(b, c) : b \neq c, b \neq a, c \neq a} z_{b}
  z_{c} }{ \paren{\sum_{b} z_{b}}^2  } \ge 0,$$
which can be derived through elementary calculus.

To further show strong adaptive submodularity, we prove pointwise submodularity
of the objective. For any fixed realization, $h$, the objective
$S \mapsto f(S, \rlz)$ amounts to a weighted set cover problem on the
incorrect hypotheses (plus a constant), and is thus submodular.
\end{proof}

Hence we can apply Theorem~\ref{thm:min-set-cover-avg-generalized} to
this \certifying instance with maximum reward threshold $Q =  1 -  1/|\hypotheses|$, and
minimum gap $\eta = 1/|\hypotheses|$, to obtain an upper bound of
$\OPT \paren{\ln \paren{|\hypotheses|-1} + 1}^2$ on the number of
queries made by the
generalized binary search algorithm (which corresponds exactly to the greedy policy
for Adaptive Stochastic Min Cost Cover)
under the assumption of a uniform
prior over $\hypotheses$.

\subsection{Arbitrary Priors}
Now consider general priors over $\hypotheses$.
We construct the
Adaptive Stochastic Min Cost Cover
instance as before, only we change the objective function to
\begin{equation}
  \label{eq:active-learning-obj}
f(S, \rlz_h) \ :=  \ 1 - \prior(\vs(\prlz(S, h)  ) ) + \prior(h)\mbox{.}
\end{equation}
First note that the instances remain \certifying.  The proof of \lemref{lem:active-learning-certifying}
goes through completely unchanged by the modification of $f$.  We proceed to
show adaptive submodularity and strong adaptive monotonicity.

\begin{lemma} \label{lem:active-learning-adapt-submod-arbitrary-priors}
The objective function $f$ as described in \eqnref{eq:active-learning-obj}
is strongly adaptive monotone and strongly \term submodular with
respect to arbitrary priors $\prior$.
\end{lemma}

\begin{proof}
The modified objective is still \term submodular, because
$(S, \rlz_h) \mapsto \prior(h)$ is clearly so, and because \term submodularity is defined
via linear inequalities it is preserved under taking nonnegative linear combinations.
Note that $f(\data, \rlz_h) = 1$ for all $\rlz_h$.
It is still strongly adaptive submodular, because it is still pointwise
submodular.

Showing $f$ is strongly adaptive monotone requires slightly more
work than before.
Fix $\prlz, x \notin \dom(\prlz)$, and
$\outcome \in \outcomes$.  We must show
\begin{equation}
  \label{eq:active-learning2}
  \expctoverrlz{\rvrlz}{f(\dom(\prlz), \rvrlz) \ \mid \ \rvrlz \sim \prlz} \le
\expctoverrlz{\rvrlz}{f(\dom(\prlz) \cup \set{x}, \rvrlz) \ \mid \ \rvrlz \sim
  \prlz, \rvrlz(x) = \outcome}.
\end{equation}
Plugging in
the definition of $f$, the
inequality we wish to prove may be simplified to
\begin{equation}
  \label{eq:active1}
  \expctoverrlz{\rvrlz}{\prior(\rvrlz) \ \mid \ \rvrlz \sim \prlz} -
\expctoverrlz{\rvrlz}{\prior(\rvrlz) \ \mid \ \rvrlz \sim \prlzxo} \
\le \ \prior(\vs(\prlz)) - \prior(\vs(\prlzxo))\mbox{.}
\end{equation}
where $\rvrlz$ is the random realization of the hypothesis, and
$\prior(\rlz_h) = \prior(h)$ for all $h$.
Let $\vselim := \vs(\prlz) - \vs(\prlzxo)$ be the set of hypotheses
eliminated from the version space by the observation $h(x) = \outcome$.
Rewriting \eqnref{eq:active1}, we get
\begin{equation}
  \label{eq:active2}
\sum_{h \in \vs(\prlz)} \frac{\prior(h)^2}{\prior(\vs(\prlz))} \
- \sum_{h \in \vs(\prlzxo)}
  \frac{\prior(h)^2}{\prior(\vs(\prlzxo))} \ \le \ \prior(\vselim)
\mbox{.}
\end{equation}
Let $\operatorname{LHS}_{\ref{eq:active2}}$ denote the left hand side
of~\eqnref{eq:active2}.  We prove~\eqnref{eq:active2}  as follows.
\[
\begin{array}{lclr}
\operatorname{LHS}_{\ref{eq:active2}}
& \le &  \sum_{h \in
    \vselim} \prior(h)^2 / \prior(\vs(\prlz)) &
  [\text{since }\prior(\vs(\prlzxo)) \le \prior(\vs(\prlz)) ]\vspace{2mm}\\
 & \le & \sum_{h \in
    \vselim} \prior(h) \cdot
    \prior(\vs(\prlz)) / \prior(\vs(\prlz)) & \hspace{0mm} [\text{since } h \in
  \vs(\prlz) \implies \prior(h)
  \le \prior(\vs(\prlz))  ]\vspace{2mm}\\
 & = &  \prior(\vselim) & \\
\end{array}
\]

\noindent We conclude that $f$ is \term submodular and strongly
adaptive monotone.
\end{proof}

Hence we can apply Theorem~\ref{thm:min-set-cover-avg-generalized} to
this \certifying instance
with maximum reward threshold $Q =  1$, and
minimum gap $\eta = 1/\min_{h} \prior(h)$.   As a result we obtain an upper bound of
$\OPT \paren{\ln \paren{1/\min_{h} \prior(h)} + 1}^2$ on the number of
queries made by
generalized binary search for arbitrary priors, completing the proof
of \thmref{thm:optimal-decision-trees}.

\subsection{Improving the Approximation Factor for Highly Nonuniform Priors}
To improve this to an $\cO\paren{\paren{\log |\hypotheses|}^2}$-approximation
in the
event that $\min_{h} \prior(h)$ is extremely small
using the
observation of~\citet{kosaraju99}, call a policy $\policy$
\emph{progressive} if it eliminates at least one hypothesis from its
version space in each query.  Let
$\prior'(h) =  \max\set{\prior(h),
  1/|\hypotheses|^{2}}/Z$
be the modified prior, where $Z := \sum_{h'}  \max\set{\prior(h'),
  1/|\hypotheses|^{2}}$ is the normalizing constant.
Let $\c(\policy, h)$ be the cost (i.e., \# of queries) of $\policy$ under target $h$.  Then
$\acst{\policy, p} := \sum_{h} \c(\policy,h) p(h)$ is the expected cost of $\policy$
under prior $p$.
We will show that $\acst{\policy, \prior'}$ is a good approximation to
$\acst{\policy, \prior}$.
Call $h$ \emph{rare} if $\prior(h) < 1/|\hypotheses|^{2}$, and
\emph{common} otherwise.
First, note that $\sum_{h'}  \max\set{\prior(h'), 1/|\hypotheses|^{2}}
\le 1 + 1/|\hypotheses|$, and so
$\prior'(h) \ge \frac{|\hypotheses|}{|\hypotheses| + 1} \prior(h)$,
for all $h$.   Hence for all $\policy$, we have
$\acst{\policy, \prior'} \ge \frac{|\hypotheses|}{|\hypotheses| + 1} \acst{\policy, \prior}$.
Next, we show $\acst{\policy, \prior'} \le \acst{\policy, \prior} + 1$.
Consider the quantity $\acst{\policy, \prior'} - \acst{\policy, \prior} =  \sum_{h}
\c(\policy,h) \paren{\prior'(h) - \prior(h)}$.
The positive contributions must come from rare hypotheses.
However, the total probability mass of these under $\prior'$ is at
most $1/|\hypotheses|$, and since $\policy$ is progressive
$\c(\policy,h) \le |\hypotheses|$ for all $h$, hence the difference in costs is at most one.
Let $\alpha := \paren{\ln \paren{\frac{1}{\min_{h} \prior'(h)}} + 1}^2 \le
\paren{\ln \paren{|\hypotheses|^2 + |\hypotheses|} + 1}^2$ be the approximation
factor for generalized binary search when run on $\prior'$.
Let $\policy$ be the policy of generalized binary search, and
let $\policy^*$ be an optimal policy under prior $\prior$.
Then
\[
\acst{\policy, \prior} \le \frac{|\hypotheses|+1}{|\hypotheses|} \,\acst{\policy,
  \prior'}
\le \frac{|\hypotheses|+1}{|\hypotheses|} \, \alpha \, \acst{\policy^*,
  \prior'} \le  \frac{|\hypotheses|+1}{|\hypotheses|} \, \alpha \, \paren{\acst{\policy^*, \prior} + 1}.
\]
Thus for a general prior a simple modification of GBS
yields an $\cO\paren{\paren{\log |\hypotheses|}^2}$-approximation.

\subsection{Extensions to Arbitrary
  Costs, Multiple Classes, and Approximate Greedy Policies}
This result easily generalizes to handle
the setting of multiple classes / test outcomes  (i.e., $|\outcomes| \ge 2$), and
$\alpha$-approximate greedy policies, where we lose a factor of
$\alpha$ in the approximation factor.
As we describe
in \appendixA,
we can
generalize \term submodularity to incorporate costs on items,
which allows us to extend this result to handle query costs as well.
%
%
%
Recently, \citet{gupta10approximation} showed how to simultaneously remove the
dependence on both costs and probabilities from the approximation ratio.
Specifically, within the context of studying an adaptive travelling
salesman problem
they investigated the \emph{Optimal Decision Tree} problem, which is
equivalent to the active learning problem we consider here.  Using a clever, more
complex algorithm than adaptive greedy,
they achieve an $\cO\paren{\log |\hypotheses|}$-approximation
in the case of non-uniform  costs and general priors.

\subsection{Extensions to Active Learning with Noisy Observations}
\thmref{thm:optimal-decision-trees} and the extensions mentioned so far
are in the noise free case, i.e., the result of query $x$ and observes
$\target(x)$, where $\target$ is the target hypothesis.
Many practical problems may have \emph{noisy} observations.
\citet{nowak09} considered the case in which the outcomes are binary,
i.e., $\outcomes = \set{-1,1}$, the same query may be asked multiple
times, and for each instance of each query the noise is independent.
In this case he gives performance guarantees for generalized binary
search.  While this setting may be appropriate if the noise is due to
measurement error, in some applications the noise is
\emph{persistent}, i.e., if query $x$ is asked several times, the observation
is always the same.  Recently, \citet{golovin10nips}
and~\citet{bellala10modified} have used the adaptive submodularity framework to obtain
the first algorithms with provable (squared--logarithmic) approximation guarantees for
active learning with persistent noise.

\section{Experiments} \label{sec:experiments}

Greedy algorithms are often straightforward to develop and implement,
which explains their popular use in practical applications, such as
Bayesian experimental design and Active Learning, as discussed in
\secref{sec:active-learning} (also see the excellent introduction of \citet{nowak09})
and Adaptive Stochastic Set Cover, e.g., for filter design in streaming databases as discussed in \secref{sec:stochastic-set-cover}.
Besides allowing us to prove approximation guarantees for such algorithms, \term submodularity provides the following immediate practical benefits:
\begin{OneLiners}
\item[1.]  The ability to use lazy evaluations to speed up its execution.
\item [2.] The ability to generate data-dependent bounds on the optimal value. 
\end{OneLiners}
In this section, we empirically evaluate their benefits within a sensor selection application, in a setting similar to the one described by~\citet{vldb04}.  
In this application,  we have deployed a network $\sensors$ of wireless sensors, e.g., to monitor temperature in a building or traffic in a road network. Since sensors are battery constrained, we must adaptively select $k$ sensors, and then, given those sensor readings, predict, e.g., the temperature at all remaining locations.  This prediction is possible since temperature measurements will typically be correlated across space.  Here, we will consider the case where sensors can fail to report measurements due to hardware failures, environmental conditions or interference.

\subsection{The Sensor Selection Problem with Unreliable Sensors}

More formally, we imagine every location $\sensor\in\sensors$ is associated with a random variable $\cX_{\sensor}$ 
describing the temperature at that location, and there is a joint probability
distribution $\mass{\bx_{\sensors}} := \Pr{\cX_{\sensors} = \bx_{\sensors}}$ that models the correlation between temperature
values.  Here, $\cX_{\sensors}=[\cX_{1},\dots,\cX_{\numsensors}]$ is
the random vector over all temperature values. We follow \citet{vldb04} and assume that the joint distribution of the sensors is multivariate Gaussian.
A sensor $\sensor$ can make a noisy observation $\cY_{\sensor}=\cX_{\sensor}+\varepsilon_{\sensor}$, where $\varepsilon_{\sensor}$ is zero mean Gaussian noise with known variance $\sigma^{2}$.  
If some measurements $\cY_{A}=\by_{A}$ are obtained at a subset of locations, then the
conditional distribution
$\mass{\bx_{\sensors} \mid \by_{A}} := \Pr{\cX_{\sensors} =
    \bx_{\sensors} \mid\cY_{A}=\by_{A}}$ allows
predictions at the unobserved locations, e.g., by predicting
$\mathbb{E}[\cX_{\sensors}\mid\cY_{A}=\by_{A}]$ (which minimizes the mean squared error).
Furthermore, this conditional distribution quantifies the
\emph{uncertainty} in the prediction: Intuitively, we would like to
select sensors that minimize the predictive uncertainty.  One way to
quantify the predictive uncertainty is to use the remaining
Shannon entropy 
$$\entropy{ \cX_{\sensors}\mid\cY_{A} = \by_{A} }
:= \expct{-\log_2 \paren{\mass{ \cX_{\sensors}\mid \by_{A} }}}.$$

\noindent \looseness -1 We would like to adaptively select  $k$
sensors, to maximize the expected reduction in Shannon entropy (\cf~\citet{Sebastiani2000,KrauseG09}).  However, in practice, sensors are often unreliable, and might fail to report their measurements. We assume that after selecting a sensor, we find out whether it has failed or not 
before deciding which  sensor to select next.  We suppose that each sensor has an associated probability $\pfail{\sensor}$ of failure, in which case no reading is reported, and that sensor failures are independent of each other and of the ambient temperature at $\sensor$.
Thus we have an instance of the Stochastic Maximization problem with $\groundset := \sensors$, 
$\outcomes := \set{\text{working}, \text{failed}}$, and 
\begin{equation}
  \label{eq:exp-obj}
f(A, \rlz) := \entropy{ \cX_{\sensors}} - \entropy{ \cX_{\sensors}
 \  \mid \ \by_{ \set{  \sensor \ : \ \rlz(\sensor) = \text{working}  }  }   }  .
\end{equation}

\noindent For multivariate normal distributions, the entropy is given as
$$\entropy{ \cX_{\sensors}\mid\cY_{A} = \by_{A} }=\frac{1}{2}\ln (2\pi e)^{\numsensors}\left| \Sigma_{\sensors A}\left(\Sigma_{AA}+\sigma^{2} I\right)^{-1}\Sigma_{A \sensors}\right|,$$
where for sets $A$ and $B$, $\Sigma_{AB}$ denotes the covariance (matrix) between random vectors $\cX_{A}$ and $\cX_{B}$. Note that the predictive covariance does not depend on the actual observations $\by_{\cA}$, only on the set $A$ of chosen locations. Thus,
$$\entropy{ \cX_{\sensors}\mid\cY_{A} = \by_{A} } = \entropy{ \cX_{\sensors}\mid\cY_{A}},$$
where as usual, 
$\entropy{ \cX_{\sensors} \  \mid \ \cY_{A}} = \expct{ \entropy{
    \cX_{\sensors} \  \mid \ \cY_{A}= \by_A } }$.
As \citet{Krause05a} show, the function
\begin{equation}
  \label{eq:experiments-infogain}
  g(A) := \infogain{\cX_{\sensors}}{\cY_{A}}=\entropy{\cX_{\sensors}}-\entropy{\cX_{\sensors}\mid\cY_{A}}
\end{equation}
is monotone submodular, whenever the observations $\cY_{\sensors}$ are conditionally independent given $\cX_{\sensors}$.

This insight allows us to apply the result of
\secref{sec:stochastic-maximization} to show that the objective $f$ defined in~\eqnref{eq:exp-obj}
is adaptive monotone submodular, using 
$\hat{f}(S) := g(\set{\sensor : (\sensor, \text{working}) \in S})$ for
any $S \subseteq \groundset \times \outcomes$.

\subsubsection{Data and Experimental Setup}

Our first data set consists of temperature measurements from the network of $46$ sensors deployed at Intel Research Berkeley, which were sampled  at $30$ second intervals for $5$ consecutive days (starting Feb. $28^{\text{th}}$, $2004$). We define our objective function with respect to the empirical covariance estimated from the data.

We also use data from traffic sensors deployed along the highway I-$880$ South in California. We use traffic speed data for all working days from $6$ AM to $11$ AM for one month, from $357$ sensors. The goal is to predict the speed on all $357$ road segments. We again estimate the empirical covariance matrix.

\ignore{
All experiments were conducted using a Matlab R$2009$b
implementation of the adaptive greedy algorithms on a laptop with a
$2.26$ GHz Intel Core 2 Duo CPU and $4$ GB of DDR$3$ RAM.
} %

\subsubsection{The Benefits of Lazy Evaluation}
For both data sets, we run the adaptive greedy algorithm, using both the
naive implementation (\algref{alg:greedy}) and the accelerated version
using lazy evaluations (\algref{alg:acc-greedy}). We vary the
probability of sensor failure, 
and evaluate the execution time and the number of evaluations of the function $g$ (defined in
\eqnref{eq:experiments-infogain}) each algorithm makes.  
Figures~\ref{fig:berkeley-time} and~\ref{fig:traffic-time} plot
execution time given a $50\%$ sensor failure rate, on a computer with a
$2.26$ GHz dual core processor and $4$ GB RAM.  
In these applications, function evaluations are the bottleneck in the computation, so the
number of them serves as a machine-independent proxy for the running time. 
Figures~\ref{fig:berkeley-evals} and~\ref{fig:traffic-evals} show the
performance ratio in terms of this proxy. 
On the temperature data set, lazy evaluations speed up the computation
by a factor of between roughly $3.5$ and $7$, depending on the failure
probability.  On the larger traffic data set, we obtain speedup factors
between $30$ and $38$.
We find that the benefit of the lazy evaluations
increases with the problem size and with the
failure probability.  
The dependence on problem size must ultimately be explained in terms
of structural properties of the instances,
which also benefit the nonadaptive accelerated greedy algorithm.
The dependence on failure probability has a simpler explanation.
Note that in these applications, if the accelerated greedy algorithm selects $\sensor$, which then 
fails, then it does not need to make any additional function evaluations to select the
next sensor.  Contrast this with the naive greedy algorithm, which 
makes a function evaluation for each sensor that has not been selected
so far.  

\subsubsection{The Benefits of the Data Dependent Bound}
While \term submodularity allows us to prove worst-case performance
guarantees for the adaptive greedy algorithm, in many practical
applications it can be expected that these bounds are quite loose.
For our sensor selection application, we use the data dependent
bounds of \lemref{lem:rate-equation} to compute an upper bound
$\avgbound$ on $\max_{\policy} \avgf(\prune{\policy}{k})$ as described below,
and compare it with the performance guarantee
of~\thmref{thm:max-cover}. 
For the accelerated greedy algorithm, we use the upper bounds on the
marginal benefits stored in the priority queue instead of recomputing
the marginal benefits, and thus expect somewhat looser bounds. We find
that for our application, the bounds are tighter than the worst
case bounds. We also find that the ``lazy'' data dependent bounds are
almost as tight as the ``eager'' bounds using the 
eagerly recomputed marginal benefits $\diff{\prlz}{\elem}$  for the latest and
greatest $\prlz$, though the former have slightly higher variance.  
Figures~\ref{fig:berkeley-rewards} and~\ref{fig:traffic-rewards} show the
performance of the greedy algorithm as well as the three bounds on the
optimal value.

Two subtleties arise when using the data-dependent bounds 
to bound $\max_{\policy} \avgf(\prune{\policy}{k})$.
The first is that \lemref{lem:rate-equation} tells us that  
$\diff{\prlz}{\prune{\policy^*}{k}} \ \le \ \max_{A \subseteq \groundset, |A| \le
  k} \ \sum_{e \in A} \diff{\prlz}{e}$, whereas we would like to bound
the difference between
the optimal reward and 
the algorithm's current expected reward, conditioned on seeing
$\prlz$, i.e., $\expct{f(\played{\prune{\policy^*}{k}}{\rvrlz}, \rvrlz)  -
  f(\dom(\prlz), \rvrlz) \ \mid \ \rvrlz \sim \prlz  }$.
However, in our applications $f$ is strongly \term monotone, and strong
\term monotonicity implies that for any $\policy^*$ we have 
\begin{equation}
  \label{eq:experiments1}
  \expct{f(\played{\prune{\policy^*}{k}}{\rvrlz}, \rvrlz)  -
  f(\dom(\prlz), \rvrlz) \ \mid \ \rvrlz \sim \prlz  } \le \diff{\prlz}{\prune{\policy^*}{k}}. 
\end{equation}
Hence, if we let $\OPT(\prlz) := \max_{\policy}
\expct{f(\played{\prune{\policy}{k}}{\rvrlz}, \rvrlz) \ \mid \ \rvrlz \sim
  \prlz}$, 
 \lemref{lem:rate-equation} implies that 
\begin{equation}
  \label{eq:experiments2}
 \OPT(\prlz)  \le  \expct{f(\dom(\prlz), \rvrlz) \ \mid \ \rvrlz \sim
   \prlz} + \max_{A \subseteq \groundset, |A| \le
  k} \ \sum_{e \in A} \diff{\prlz}{e}. 
\end{equation}

The second subtlety is that we obtain a sequence of bounds from
\eqnref{eq:experiments2}.
If we consider the (random) sequence of partial realizations observed by the adaptive
greedy algorithm, $\emptyset = \prlz_0 \subset \prlz_1 \subset \cdots
\subset \prlz_k$, we obtain $k+1$ bounds $\bound_0, \ldots,
\bound_{k}$, where 
$\bound_{i} :=  \expct{f(\dom(\prlz_i), \rvrlz) \ \mid \ \rvrlz \sim
   \prlz_i} + \max_{A \subseteq \groundset, |A| \le
  k} \ \sum_{e \in A} \diff{\prlz_i}{e}$.
Taking the expectation over $\rvrlz$, note that for any $\policy$, and
any $i$, 
$$\avgf(\prune{\policy}{k}) \le \expct{\OPT(\prlz_i)} \le \expct{\bound_i}.$$
Therefore for any $0 \le i \le k$ , $\bound_i$ is a random variable
whose \emph{expectation} is an upper bound on the optimal expected reward of
any policy.  At this point we may be tempted to use the minimum of
these, i.e., $\bound_{\min} := \min_i \set{\bound_i}$ as our ultimate bound.
However, a collection of random variables $X_0, \ldots, X_k$ with
$\expct{X_i} \ge \tau$ for all $i$ does not, in general, satisfy 
$\min_i \set{X_i} \ge \tau$.  While it is possible in our case, with its
independent sensor failures, to use concentration inequalities to bound 
$\min_i \set{\bound_i} - \min_i \set{\expct{\bound_i}}$ with high
probability, and thus add an appropriate term to obtain a true upper
bound from $\bound_{\min}$, we take a different approach;
we simply use the average bound 
$\avgbound := \frac{1}{k+1}\sum_{i=0}^{k} \bound_i$.
Of course, depending on the application, a particular bound $\bound_i$
(chosen independently of the sequence $ \prlz_0, \prlz_1, \ldots,
\prlz_k$) may be superior.
For example, if $g$ is modular, then $\bound_0$ is best, whereas if
$g$ exhibits strong diminishing returns, then bounds $\bound_i$ with
larger values of $i$ may be significantly tighter.

\newcommand{\figwidth}{0.40\textwidth}
\newcommand{\figheightA}{0.22\textwidth}
\newcommand{\figheightB}{0.24\textwidth}

\begin{figure*}%
\centering 
\subfigure[\emph{Temperature Data: Execution time (sec) for 
 the naive vs accelerated implementations of adaptive greedy
  vs.\ the budget $k$ on number of sensors selected, 
  when $\pfail{\sensor} = 0.5$ for all $\sensor$, plotted with standard errors.}]{
\includegraphics[width=\figwidth]{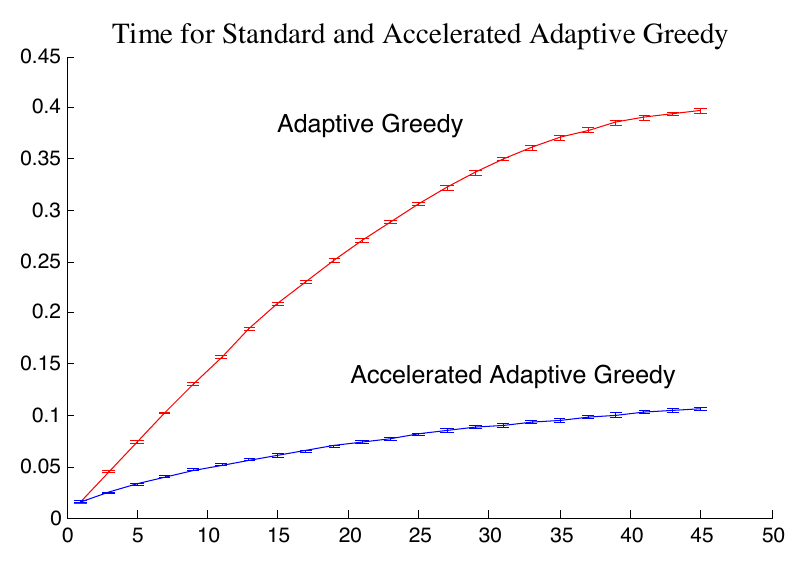} %
 \label{fig:berkeley-time}
}
\hspace{5mm}
\subfigure[\emph{Traffic Data: Execution time (sec) for 
 the naive vs accelerated implementations of adaptive greedy
  vs.\ the budget $k$ on number of sensors selected, 
  when $\pfail{\sensor} = 0.5$ for all $\sensor$, plotted with standard errors.}]{
\includegraphics[width=0.38\textwidth]{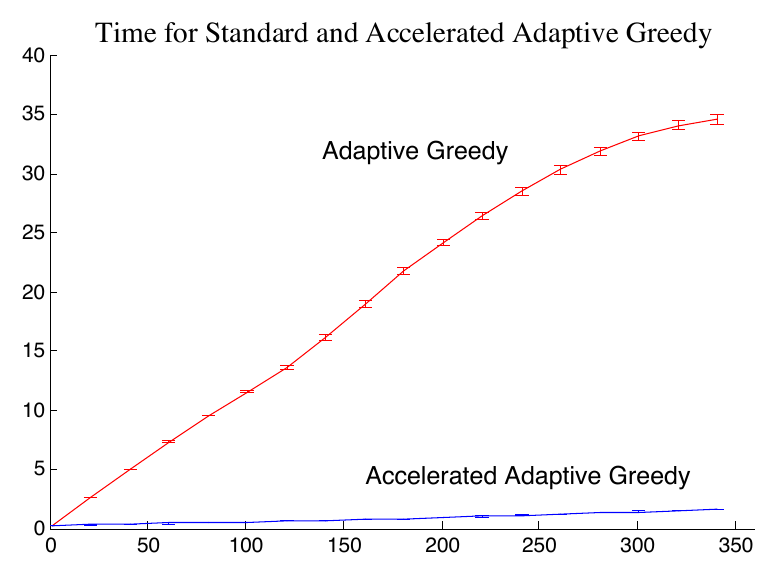}
 \label{fig:traffic-time}
}
\vspace{1mm}
\subfigure[\emph{Temperature Data: The ratio of function evaluations
  made by the naive vs accelerated implementations of adaptive greedy
  vs.\ the budget $k$ on number of sensors selected, for various
  failure rates.  Averaged over $100$ runs.}]{
\includegraphics[width=\figwidth]{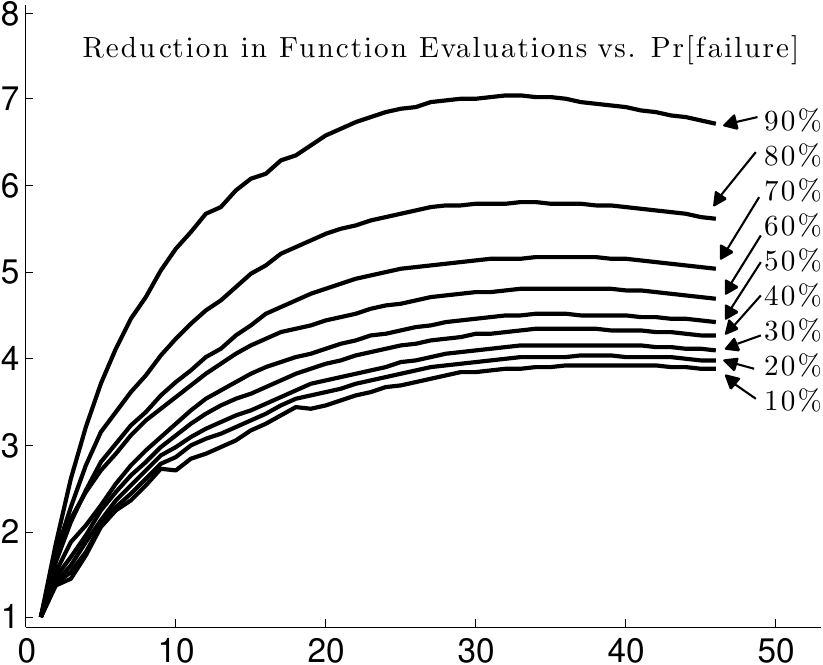}
 \label{fig:berkeley-evals}
 }
\hspace{5mm}
\subfigure[\emph{Traffic Data: The ratio of function evaluations
  made by the naive vs accelerated implementations of adaptive greedy
  vs.\ the budget $k$ on number of sensors selected, for various
  failure rates. Averaged over $10$ runs.}]{
\includegraphics[width=\figwidth]{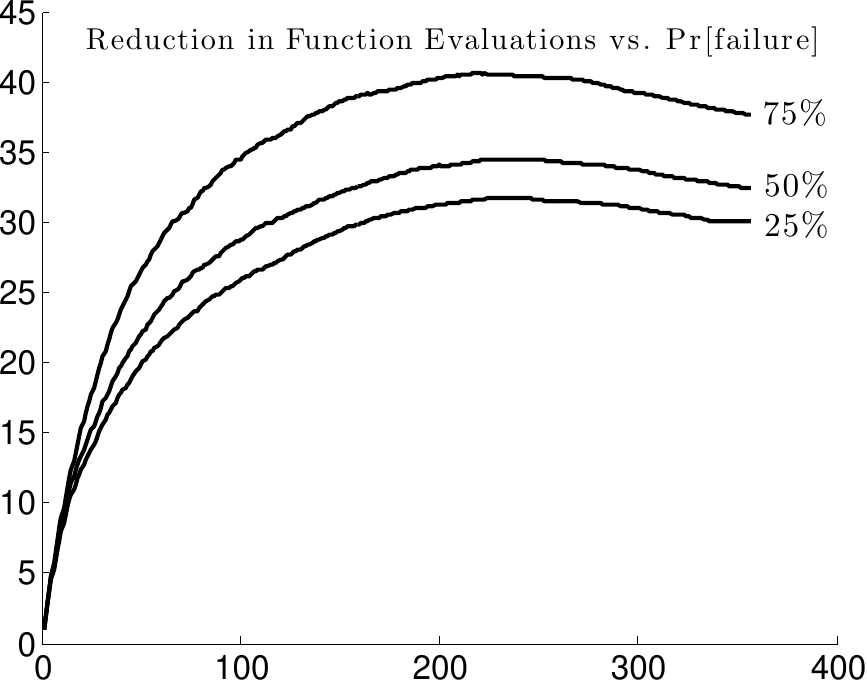}
 \label{fig:traffic-evals}
}
\vspace{1mm}
\subfigure[\emph{Temperature Data: Rewards \& bounds on the \mbox{optimal} value when $\pfail{\sensor} = 0.5$ for
   all $\sensor$ vs.\ the budget $k$ on number of sensors
   selected, plotted with standard errors. %
}]{
\includegraphics[width=\figwidth, height=2in]{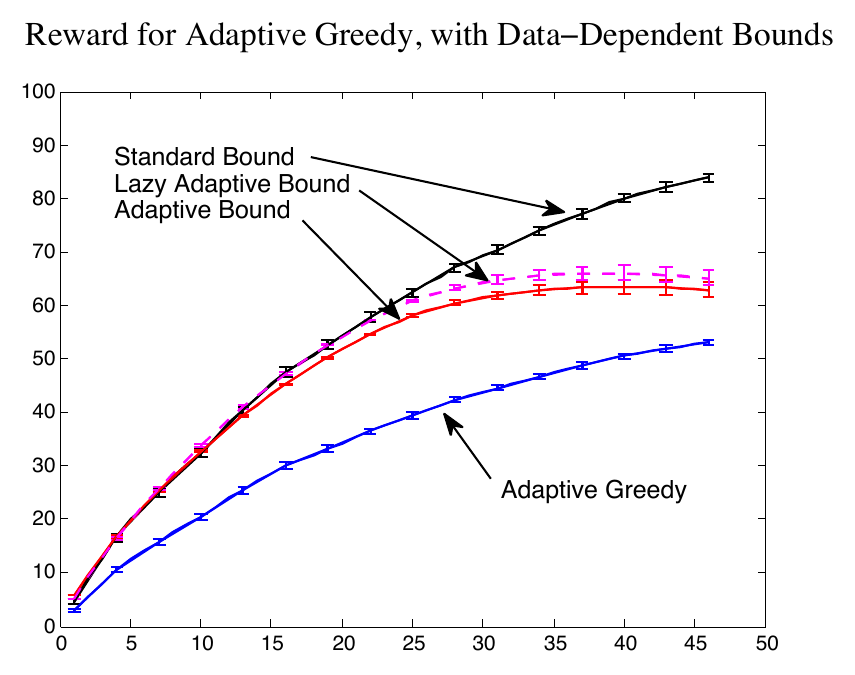}
 \label{fig:berkeley-rewards}
 }
\hspace{5mm}
\subfigure[\emph{Traffic Data: Rewards \& bounds on the optimal value when $\pfail{\sensor} = 0.5$ for
   all $\sensor$ vs.\ the budget $k$ on number of sensors
   selected, plotted with standard errors. %
}]{
\includegraphics[width=\figwidth, height=2in]{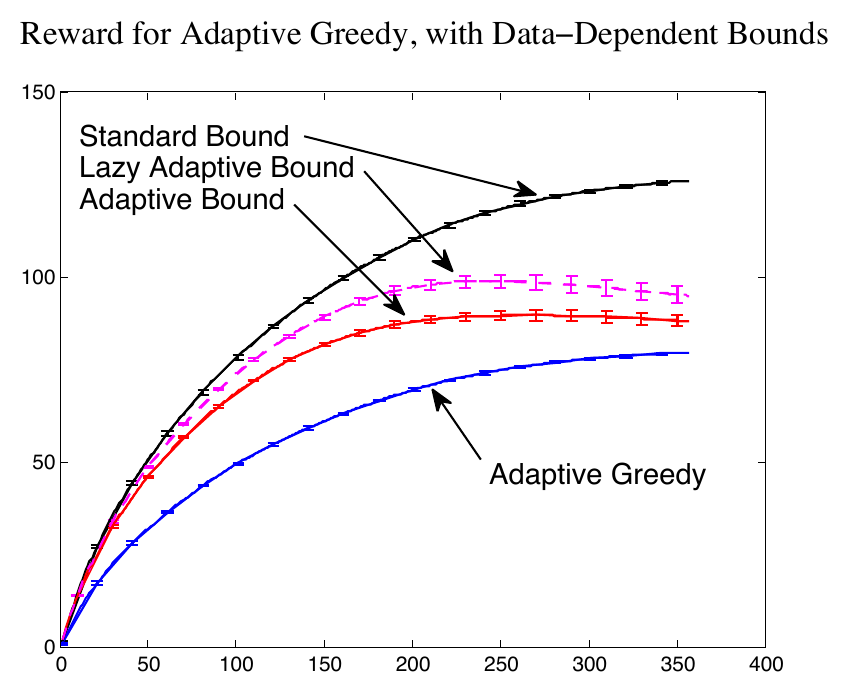}
 \label{fig:traffic-rewards}
 }
\caption{%
Experimental results.  
\label{fig:experiments}
 }%
\end{figure*}

\TechReportOnly{
\section{Adaptivity Gap}\label{sec:adaptgap}
An important question in adaptive optimization is \emph{how much
  better} adaptive policies can perform when compared to non-adaptive
policies.  This is quantified by the \emph{adaptivity gap}, which is the
worst-case ratio, over problem instances, of the performance of the optimal adaptive policy to the optimal non-adaptive solution.
\citet{AsadpourNS08} show that in the Stochastic Submodular
Maximization problem with independent failures (as considered in
\secref{sec:stochastic-maximization}), the expected value of the
optimal non-adaptive policy is at most a constant factor $1-1/e$ worse
than the expected value of the optimal adaptive policy. 
While we currently do not have lower bounds for the adaptivity gap of the
general Adaptive Stochastic Maximization problem~\eqref{eq:stochmax},
we can show that even in the case of \term submodular functions, the
min-cost cover and min-sum cover versions 
have large adaptivity gaps, and thus there
is a large benefit of using adaptive algorithms.  
In these cases, the adaptivity gap is defined as the worst-case ratio
of the expected cost of the optimal non-adaptive policy divided by 
the expected cost of the optimal adaptive policy.
For the Adaptive Stochastic Minimum Cost Coverage problem~\eqref{eq:stochcover}, \citet{goemans06stochastic} show the special case
of Stochastic Set Coverage without
multiplicities has an adaptivity gap of $\Omega(|\groundset|)$.
Below we exhibit an adaptive stochastic optimization instance with
adaptivity gap of $\Omega(|\groundset|/\log|\groundset|)$ for the
Adaptive Stochastic Min-Sum Cover problem~\eqref{eq:minsumcover}, which also happens to
have the same adaptivity gap for
Adaptive Stochastic Minimum Cost Coverage.

\begin{theorem}
Even for \term submodular functions, the adaptivity gap of
Adaptive Stochastic Min-Sum Cover 
is $\Omega(n/\log n)$, where $n=|\groundset|$. 
\end{theorem}
\begin{proof}
Suppose $\groundset = \{1,\dots,n\}$. Consider the active learning
problem where our hypotheses $h:\groundset\rightarrow\set{-1,1}$ are
threshold functions, i.e., $h(\elem)=1$ if $\elem\geq \ell$ and
$h(\elem)=-1$ if $\elem<\ell$ for some threshold $\ell$. There is a
uniform distribution over thresholds $\ell\in\{1,\dots,n+1\}$. In
order to identify the correct hypothesis with threshold $\ell$, our
policy must observe at least one of  $\ell-1$ or $\ell$ (and both
of them if $1<\ell\leq n$).  
Let $\hat{\policy}$ be an optimal non-adaptive
policy for this problem.
Note that $\hat{\policy}$ can be represented as a permutation of
$\groundset$, because
observing an element multiple times can only increase the cost
while providing no benefit over observing it once,
and each element must eventually be selected to guarantee coverage. 
For the min-sum cover objective, consider playing $\hat{\policy}$ for
$n/4$ time steps. Then $\prob{\ell \text{ observed in } n/4 \text{
    steps}} =n/4(n+1)$.  Likewise 
$\prob{\ell-1 \text{ observed in } n/4 \text{
    steps}} =n/4(n+1)$.
Since at least one of these events must occur to identify the correct
hypothesis, by a union bound 
$$ \prob{\hat{\policy} \text{ identifies the correct hypothesis in } n/4 \text{
    steps}} \quad  \le \quad \frac{n}{2(n+1)} \quad \le \quad  1/2.$$
Thus a lower bound on the expected cost of $\hat{\policy}$ is $n/8$, since for $n/4$ time
steps a cost of at least $1/2$ is incurred. Thus, for both the min-cost and min-sum cover objectives the cost of the optimal non-adaptive policy is $\Omega(n)$.

As an example adaptive policy, we can implement a natural binary
search strategy, which is guaranteed to identify the correct
hypothesis after $O(\log n)$  steps, thus incurring cost $O(\log n)$, 
proving an adaptivity gap of $\Omega(n/\log n)$.
\end{proof}

} %

\section{Hardness of Approximation} \label{sec:hardness}
In this paper, we have developed the notion of \term submodularity,
which characterizes when certain adaptive stochastic optimization
problems are well-behaved in the sense that a simple greedy policy
obtains a constant factor or polylogarithmic factor approximation to the
best policy.

In contrast, we can also show that without \term submodularity, the adaptive stochastic
optimization problems~\eqref{eq:stochmax},~\eqref{eq:stochcover}, and~\eqref{eq:minsumcover} are extremely inapproximable, even with (pointwise) \emph{modular} objective functions (i.e., those where for each $\rlz$, $f:2^{\groundset}\times \outcomes^{\groundset}\rightarrow\mathbb{R}$ is modular/linear in the first argument):
We cannot hope to achieve an $\cO(|\groundset|^{1-\varepsilon})$
approximation ratio for these problems, unless the polynomial hierarchy
collapses down to $\Sigma^P_2$.

\begin{theorem} \label{thm:hardness}
For all (possibly non-constant) $\beta \ge 1$,
no polynomial time algorithm for Adaptive Stochastic
Maximization with a budget of $\beta k$ items
can approximate the reward of an optimal policy with a budget of only $k$ items
to within a
multiplicative factor of $\cO(|\groundset|^{1-\varepsilon}/\beta)$ for any
$\varepsilon > 0$, unless $\class{PH} = \Sigma^P_2$. This holds even
for pointwise modular $f$.
\end{theorem}

We provide the proof of \thmref{thm:hardness} in Appendix~\ref{sec:proof-approx-hardness}.
Note that by setting $\beta = 1$, we obtain $\cO(|\groundset|^{1-\varepsilon})$
hardness for Adaptive Stochastic Maximization.
It turns out that
in the instance distribution we construct in the
proof of Theorem~\ref{thm:hardness} the
optimal policy covers every realization (i.e., always finds the
treasure) using a budget of
$k = \cO(|\groundset|^{\varepsilon/2})$
items.
Hence if $\class{PH} \neq \Sigma^P_2$ then
any randomized polynomial time algorithm wishing to cover this
instance must have a budget
$\beta = \Omega(|\groundset|^{1-\varepsilon})$ times larger than the
optimal policy, in order to ensure the ratio of rewards, which is
$\Omega(|\groundset|^{1-\varepsilon}/\beta)$, equals one.
This yields the following corollary.

\begin{corollary} \label{thm:coverage-hardness}
No polynomial time algorithm for Adaptive Stochastic Min
Cost Coverage
can approximate the cost of an optimal policy
to within a multiplicative factor of $\cO(|\groundset|^{1-\varepsilon})$ for any
$\varepsilon > 0$, unless $\class{PH} = \Sigma^P_2$. This holds even for pointwise modular $f$.
\end{corollary}

Furthermore, since in the instance distribution we construct the
optimal policy $\policy^*$ covers every realization using a budget of $k$, it has
$\costminsum{\policy^*} \le k$.  Moreover,
since we have shown that under our
complexity theoretic assumptions, any polynomial time randomized
policy $\policy$ with budget $\beta k$ achieves at most
$o(\beta / |\groundset|^{1-\varepsilon})$ of the (unit) value obtained by the
optimal policy with budget $k$, it follows that
$\costminsum{\policy} = \Omega(\beta k)$.  Since we require $\beta =
\Omega(|\groundset|^{1-\varepsilon})$ to cover any set of
realizations constituting, e.g., half of the probability mass, we obtain the following corollary.

\begin{corollary} \label{thm:min-sum-hardness}
No polynomial time algorithm for Adaptive Stochastic Min-Sum Cover
can approximate the cost of an optimal policy
to within a multiplicative factor of $\cO(|\groundset|^{1-\varepsilon})$ for any
$\varepsilon > 0$, unless $\class{PH} = \Sigma^P_2$. This holds even for pointwise modular $f$.
\end{corollary}

\section{Related Work} \label{sec:related-work}
There is a large literature on adaptive optimization under partial
observability which relates to \term submodularity, which can be broadly organized into several different categories. Here, we only review relevant related work that is not already discussed elsewhere in the manuscript.

\subsection{Adaptive Versions of Classic Non-adaptive Optimization Problems} Many approaches consider stochastic generalizations of specific classic non-adaptive optimization problems, such as Set Cover~\citep{goemans06stochastic,liu08near}, Knapsack \citep{dean08approximating,dean05adaptivity} and Traveling Salesman \citep{gupta10approximation}. In contrast, in this paper our goal is to introduce a general problem structure -- \term submodularity -- that unifies a number of adaptive optimization problems. This is similar to how the classic notion of submodularity unifies various optimization problems such as Set Cover, Facility Location, nonadaptive Bayesian Experimental Design, etc.

\subsection{Competitive Online Optimization}  Another active area of
research in sequential optimization is the study of \emph{competitive
  online algorithms}.
A particularly relevant example is Online Set Cover~\cite{alon09},
where there is a known set system, an arbitrary sequence of elements
is presented to the algorithm, and the algorithm must irrevocably
select sets to purchase such that at all times the purchased sets
cover all elements which have appeared so far.  \citet{alon09}~obtain a
polylogarithmic approximation to this problem, via an online
primal--dual framework which has been profitably applied to many other
problems.  \citet{buchbinder2009} provide a detailed treatment of this
framework.  Note that competitive analysis focuses on worst--case
scenarios.  In contrast, we assume probabilistic information about the
world and optimize for the average case.

\subsection{(Noisy) Interactive Submodular Set Cover}  Recent work
by~\citet{guillory10interactive,guillory2011-noisy-interactive-submod-cover}
considers a class of adaptive
optimization problems over a family of monotone submodular objectives $\set{f_h
  : h \in \hypotheses}$.
In their problem, one must cover a monotone submodular objective
$f_{\target}$ which
depends on the (initially unknown) target hypothesis $\target \in \hypotheses$, by adaptively
issuing queries and getting responses.  Unlike traditional pool-based
active learning, each query may generate a response from a \emph{set} of
valid responses depending on the target hypothesis.
The reward is calculated by evaluating $f_{\target}$ on the set of (query, response)
pairs observed, and the goal is to obtain some threshold $\quota$ of
objective value at minimum total query cost, where queries may have
nonuniform costs.
In the noisy variant of the problem~~\citep{guillory2011-noisy-interactive-submod-cover}, the set of (query, response)
pairs observed need not be consistent with any hypothesis in $\hypotheses$, and the goal
is to obtain $\quota$ of value for all hypotheses that are ``close''
to being consistent with the observations.
For both variants, \citeauthor{guillory2011-noisy-interactive-submod-cover}
consider the worst-case policy cost, and
provide greedy algorithms optimizing clever hybrid objective
functions.
They prove an approximation guarantee of $\ln(\quota |\hypotheses|)+1$ for
integer valued objective functions $\set{f_h}_{h\in \hypotheses}$ in
the noise--free case, and similar logarithmic approximation guarantees
for the noisy case.

\ignore{ %
In their problem, there is a set of hypotheses $\hypotheses$, queries
$\groundset$, responses $\outcomes$.
For each query $e$ and hypothesis $h$, there is a set of valid
responses $e(h) \subseteq \outcomes$ that may be generated by $e$ if $h$
is the true hypothesis.
Each hypothesis $h$ also has an associated monotone submodular objective function
$f_h:2^{\groundset \times \outcomes} \to \NonNegativeIntegers$,
and the goal is to obtain some quota $\quota$ of reward,
measured by evaluating $f_{\target}$ on the set of (query, response)
pairs observed,
where $\target$ is the (initially unknown) true hypothesis.
Hence Interactive Submodular Set Cover combines the problem of learning about the true
hypothesis and covering an objective function depending on
it, in much the same way Adaptive Stochastic Minimum Cost Cover does.
Guillory and Bilmes consider the worst-case policy cost, and
provide a greedy algorithm optimizing a clever hybrid objective function
with an approximation guarantee of
$\ln(\quota |\hypotheses|)+1$ for integer valued $\set{f_h}_{h\in \hypotheses}$.
} %

While similar in spirit to this work, there are several significant
differences between the two.
Guillory and Bilmes focus on worst-case
policy cost, while we focus mainly on average-case policy cost.
The structure of \term submodularity depends on the prior $\rlzprior$, whereas
there is no such dependence
in Interactive Submodular Set Cover.
This dependence in turn allows us to obtain results, such as
\thmref{thm:min-set-cover-avg-generalized} for \certifying instances,
whose approximation guarantee does not depend on the number of
realizations in the way that the guarantees for  Interactive
Submodular Set Cover depend on $|\hypotheses|$.  As
Guillory and Bilmes
prove, the latter dependence
is fundamental under reasonable complexity-theoretic assumptions\footnote{They
  reduce to Set Cover and use the result of \citet{feige98threshold},
  which requires the assumption
$\NP \nsubseteq \text{DTIME}(n^{\cO(\log\log n)})$, but it suffices to
assume only $\P \neq \NP$ using the Set Cover approximation hardness
result of~\citet{RazS97} instead.}.
An interesting open problem within the \term submodularity framework
that is highlighted by the work on  Interactive Submodular Set Cover
is to identify useful instance-specific properties that
are sufficient to improve upon the worst-case approximation guarantee of
\thmref{thm:min-set-cover-wc-generalized}.

\subsection{Greedy Frameworks for Adaptive Optimization} The paper
that is perhaps
closest in spirit to this work is the one on Stochastic Depletion problems by \citet{chan09}, who also identify a general class of adaptive optimization problems than can be near-optimally solved using greedy algorithms (which in their setting give a factor 2 approximation).
However, the similarity is mainly on a conceptual level: The problems and approaches, as well as example applications considered, are quite different.

\subsection{Stochastic Optimization with Recourse} A class of adaptive
optimization problems  studied extensively in operations research
since \citet{dantzig55linear} is the area of \emph{stochastic optimization with recourse}. Here, an optimization problem, such as Set Cover, Steiner Tree or Facility Location, is presented in multiple stages. At each stage, more information is revealed, but costs of actions increase.  A key difference to the problems studied in this paper is that in these problems, information gets revealed independently of the actions taken by the algorithm. There are general efficient, sampling based (approximate) reductions of multi-stage optimization to the deterministic setting; see, e.g.,\citet{gupta05wednesday}.

\subsection{Bayesian Global Optimization}
Adaptive Stochastic Optimization is also related to the problem of
Bayesian Global Optimization (for a recent
survey of the area, \cf \cite{nandotut}). In Bayesian Global Optimization, the goal is to
adaptively select inputs in order to maximize an unknown function that
is expensive to evaluate (and can possibly only be evaluated using
noisy observations). A common approach that has been successful in
many applications (for a recent application in machine learning, \cf \cite{lizotte07gait}), is to assume a prior distribution, such as a Gaussian process, over the unknown objective function. Several criteria for selecting inputs have been developed, such as the Expected Improvement \citep{ego98} criterion. However, while recently performance guarantees where obtained in the no-regret setting \citep{gruenewaelder10regret,srinivas10gaussian}, we are not aware of any approximation guarantees for Bayesian Global Optimization.

\subsection{Probabilistic Planning} The problem of decision making
under partial observability has also been extensively studied in
stochastic optimal control. In particular, Partially Observable Markov
Decision Processes~\citep{smallwood73optimal}, abbreviated as POMDPs, are a general framework that captures many adaptive optimization problems under partial observability. Unfortunately, solving POMDPs is PSPACE hard \citep{papadimitriou87complexity}, thus typically heuristic algorithms with no approximation guarantees are applied \citep{pineau06anytime,ross08}.
For some special instances of POMDPs related to Multi-armed Bandit
problems, (near-)optimal policies can be found. These include the
(optimal) Gittins-index policy for the classic Multi-armed Bandit problem~\citep{gittins79dynamic} and approximate policies for the Multi-armed Bandit problem with metric switching costs~\citep{guha09multi} and special cases of the Restless Bandit problem \citep{guha09approximation}. The problems considered in this paper can be formalized as POMDPs, albeit with exponentially large state space (where the world state represents the selected items and state/outcome of each item). Thus our results can be interpreted as widening the class of partially observable planning problems that can be efficiently approximately solved.

\subsection{Previous Work by the Authors \& Subsequent Developments}
This manuscript is an extended version of a paper that appeared in the
Conference on Learning Theory \cite{golovin10colt}.
More recently, \citet{golovin11matroid_arxiv} proved performance
guarantees for the greedy policy for the problem of
maximizing the expected value of a policy under constraints more
complex than simply selecting at most $k$ items.
These include \emph{matroid} constraints, where a policy can only
select independent sets of items and the greedy policy obtains a
$1/2$--approximation for adaptive monotone submodular objectives, and
more generally $p$-\emph{independence system} constraints, where the
greedy policy obtains a $1/(p+1)$--approximation.
\citet{golovin10nips} and, shortly thereafter,~\citet{bellala10modified}, used the adaptive submodularity framework to obtain the first algorithms
with provable (squared logarithmic) approximation guarantees for the difficult and fundamental problem of
active learning with persistent noise.
Finally, \citet{golovin11aaai} used adaptive submodularity in the
context of a dynamic conservation planning, and obtain competitiveness guarantees
for an ecological reserve design problem.

\section{Conclusions}
\label{sec:conclusions}
Planning under partial observability is a central but notoriously difficult problem in artificial intelligence.
In this paper, we identified a novel, general class of adaptive optimization problems under uncertainty that are amenable to efficient, greedy (approximate) solution.
In particular, we introduced the concept of \emph{\term
  submodularity}, generalizing submodular set functions to adaptive
policies.  Our generalization is based on a natural adaptive analog of the
diminishing returns property well understood for set functions.
In the special case of deterministic
distributions, \term submodularity reduces to the classical notion of
submodular set functions. We proved that several guarantees carried by the
non-adaptive greedy algorithm for submodular set functions generalize
to a natural adaptive greedy algorithm in the case of \term submodular
functions\TechReportOnly{, for constrained maximization and certain natural
coverage problems with both minimum cost and minimum sum objectives}.  We also showed how the adaptive greedy algorithm can be accelerated using lazy evaluations, and how one can compute data-dependent bounds on the optimal solution. 
We illustrated the usefulness of the concept by
giving several examples of \term submodular objectives arising in
diverse AI applications including sensor placement, viral marketing, automated diagnosis and
pool-based active learning. Proving \term submodularity for these
problems allowed us to recover existing results in these applications
as special cases and lead to natural generalizations. 
Our experiments on real data indicate that adaptive submodularity can
provide practical benefits, such as significant speed ups  
and tighter data-dependent bounds.
We believe that our results provide an interesting step in the
direction of exploiting structure to solve complex stochastic optimization and planning problems under partial observability.

\subsection*{Acknowledgments}
This research was partially supported by ONR grant N00014-09-1-1044,
NSF grant CNS-0932392, NSF grant IIS-0953413, DARPA MSEE grant FA8650-11-1-7156, a 
gift by Microsoft Corporation, an Okawa Foundation Research Grant, and by the Caltech Center for the Mathematics of Information.
We wish to thank Vitaly Feldman and Jan Vondr\'{a}k for providing an elegant proof of~\lemref{lem:price-integral}.
\bibliography{jair}

\section{Additional Proofs and Incorporating Item Costs}
 \label{sec:proofs}
In this appendix we provide all of the proofs omitted from the main text.
For the results of \secref{sec:greedy},
we do so by first explaining how our results generalize to the case where items have costs, and then proving generalizations which incorporate item costs.

\subsection{Incorporating Costs: Preliminaries} \label{sec:incorporating-costs}

In this section we provide the preliminaries required to define and analyze the versions of our problems with non-uniform item costs.
We suppose each item $\elem\in\groundset$ has a
cost $c(\elem)$, and the cost of a set
$S\subseteq\groundset$ is given by the modular function $c(S)=\sum_{\elem\in S}c(\elem)$.
We define the generalizations of problems~(\ref{eq:stochmax}), (\ref{eq:stochcover}), and (\ref{eq:minsumcover}) in \secref{sec:proofs-max-cover},~\secref{sec:proofs-min-cost-cover}, and \secref{sec:proofs-min-sum-cover},  respectively.

Our results are with respect to the greedy policy $\greedypolicy$ and $\alpha$-approximate greedy policies.  With costs, the greedy policy
selects an item maximizing $\diff{\prlz}{e}/c(e)$, where $\prlz$ is the current partial realization.

\begin{definition}[Approximate Greedy Policy with Costs]
A policy $\policy$ is an $\alpha$-\emph{approximate greedy policy} if
for all $\prlz$ such that there exists $e \in \groundset$ with $\diff{\prlz}{e} > 0$,
$$\policy(\prlz) \in \set{e \ : \ \frac{\diff{\prlz}{e}}{c(e)} \ \ge
  \ \frac{1}{\alpha} \max_{e'} \paren{\frac{\diff{\prlz}{e'}}{c(e')} }
},$$
and $\policy$ terminates upon observing any $\prlz$ such that $\diff{\prlz}{e} \le 0$ for all $e \in
\groundset$.
That is, an $\alpha$-approximate greedy policy always obtains at least
$(1/\alpha)$ of the maximum possible ratio of conditional expected
marginal benefit to cost, and terminates when no more benefit can be
obtained in expectation.
A \emph{greedy policy} is any $1$-approximate greedy policy.
\end{definition}

It will be convenient to imagine the policy executing over time, such that
when a policy $\policy$ selects an item $e$, it starts to \emph{run} $e$, and \emph{finishes running} $e$ after
$c(e)$ units of time.
We next generalize the definition of policy truncation.  Actually we require three such generalizations, which are all equivalent in the unit cost case.

\begin{definition}[Strict Policy Truncation] \label{def:strict-prune}
The \emph{strict level $t$ truncation} of a policy $\policy$, denoted by
$\strictprune{\policy}{t}$, is obtained by running $\policy$ for $t$ time units, and
unselecting items whose runs have not finished by time $t$.
Formally, $\strictprune{\policy}{t}$ has domain
$\set{\prlz \in \dom(\policy) \ : \ c(\policy(\prlz)) + \sum_{e \in \dom(\prlz)} c(e) \le t}$,
and agrees with $\policy$ everywhere in its domain.
\end{definition}

\begin{definition}[Lax Policy Truncation] \label{def:lax-prune}
The \emph{lax level $t$ truncation} of a policy $\policy$, denoted by
$\laxprune{\policy}{t}$, is obtained by running $\policy$ for $t$ time units, and
selecting the items running at time $t$.
Formally, $\laxprune{\policy}{t}$ has domain
$\set{\prlz \in \dom(\policy) \ : \ \sum_{e \in \dom(\prlz)} c(e) < t}$,
and agrees with $\policy$ everywhere in its domain.
\end{definition}

\begin{definition}[Policy Truncation with Costs] \label{def:policy-truncation-with-costs}
  The \emph{level-$t$-truncation} of a policy $\policy$, denoted by
  $\prune{\policy}{t}$, is a randomized policy obtained by
  running $\policy$ for $t$ time units, and
  if some item $e$ has been running for $0 \le \tau < c(e)$ time at time $t$,
  selecting $e$ independently with probability $\tau/c(e)$.
  Formally, $\prune{\policy}{t}$ is a randomized policy that agrees with $\policy$
  everywhere in its domain, has
  $\dom(\strictprune{\policy}{t}) \subseteq \dom(\prune{\policy}{t}) \subseteq \dom(\laxprune{\policy}{t})$ with certainty,  and  includes each $\prlz \in \dom(\laxprune{\policy}{t}) \setminus \dom(\strictprune{\policy}{t})$
 in its domain
independently with probability $\paren{t - \sum_{e \in \dom(\prlz)} c(e)}/c(\policy(\prlz))$.
\end{definition}

In the proofs that follow, we will need a notion of the conditional
expected cost of a policy, as well as an alternate characterization of
\term monotonicity, based on a notion of policy concatenation.
We prove the equivalence of our two \term monotonicity conditions in~\lemref{lem:monotonicity-equivalence}.

\begin{definition}[Conditional Policy Cost]
The \emph{conditional policy cost} of $\policy$ conditioned on
$\prlz$, denoted $\condcost{\prlz}{\policy}$, is the expected cost of
the items $\policy$ selects under $\rlzmass{\rlz \mid \prlz}$.  That is,
$\condcost{\prlz}{\policy} := \expctoverrlz{\rvrlz}{c(\played{\policy}{\rvrlz})  \ \mid \ \rvrlz \sim
  \prlz}$.
\end{definition}

\begin{definition}[Policy Concatenation] \label{def:policy-concatenation}
Given two policies $\policy_1$ and $\policy_2$
define $\append{\policy_1}{\policy_2}$ as the policy obtained by running $\policy_1$ to completion, and then running policy  $\policy_2$ as if from a fresh start, ignoring the information
 gathered\footnote{Technically, if under any realization $\rlz$ policy
   $\policy_2$ selects an item that
   $\policy_1$ previously selected, then $\append{\policy_1}{\policy_2}$
   cannot be written as a function from a set of partial realizations
   to $\groundset$, i.e., it is not a policy.  This can be amended by
   allowing partial realizations to be multisets over elements of
   $\groundset \times \outcomes$, so that, e.g., if $e$ is played
   twice
   then $(e, \prlz(e))$ appears twice in
   $\prlz$.  However, in the interest of readability we will avoid this more cumbersome multiset
   formalism, and abuse notation slightly by calling
   $\append{\policy_1}{\policy_2}$ a policy.  This issue arises
   whenever we run some policy and then run another from a fresh start.}
 during the running of $\policy_1$.
\end{definition}

\begin{definition}[\Term Monotonicity (Alternate Version)] \label{def:monotonicity-append}
A function $f:2^{\groundset} \times O^{\groundset} \to \NonNegativeReals$
is \emph{\term monotone} with respect to distribution $\rlzprior$
if for all policies $\policy$ and $\policy'$,
 it holds that
$\avgf(\policy)\leq \avgf( \append{\policy'}{\policy})$, where
$\avgf(\policy) := \expctoverrlz{\rvrlz}{f(\played{\policy}{\rvrlz}, \rvrlz)}$
is defined w.r.t.~$\rlzprior$.
\end{definition}

\begin{lemma}[\Term Monotonicity Equivalence] \label{lem:monotonicity-equivalence}
Fix a function $f:2^{\groundset} \times O^{\groundset} \to \NonNegativeReals$.
Then $\diff{\prlz}{\elem} \ge 0$ for all $\prlz$ with $\Pr{\rvrlz \sim \prlz} > 0$ and all $e \in \groundset$
if and only if
for all policies $\policy$ and $\policy'$,
$\avgf(\policy)\leq \avgf( \append{\policy'}{\policy})$.
\end{lemma}

\begin{proof}
Fix policies $\policy$ and $\policy'$.
We begin by proving $\avgf( \append{\policy'}{\policy}) = \avgf( \append{\policy}{\policy'})$.  Fix any $\rlz$ and
note that
$\played{\append{\policy'}{\policy}}{\rlz} = \played{\policy'}{\rlz} \cup  \played{\policy}{\rlz} = \played{\append{\policy}{\policy'}}{\rlz}$.
Hence $$\avgf(\append{\policy'}{\policy}) = \expctoverrlz{\rvrlz}{f(\played{\append{\policy'}{\policy}}{\rvrlz}, \rvrlz)} =
\expctoverrlz{\rvrlz}{f(\played{\append{\policy}{\policy'}}{\rvrlz}, \rvrlz)} = \avgf(\append{\policy}{\policy'}).$$
Therefore
$\avgf(\policy)\leq \avgf(\append{\policy'}{\policy})$ holds if and
only if $\avgf(\policy)\leq \avgf( \append{\policy}{\policy'})$.

We first prove the forward direction.
Suppose $\diff{\prlz}{\elem} \ge 0$ for all $\prlz$ and all $e \in \groundset$.
Note the expression $\avgf( \append{\policy}{\policy'}) - \avgf(\policy)$ can be written as a conical combination of (nonnegative) $\diff{\prlz}{\elem}$ terms, i.e.,
for some $\alpha \ge \mathbf{0}$,  $\avgf( \append{\policy}{\policy'}) - \avgf(\policy) = \sum_{\prlz,e} \alpha_{(\prlz,e)}\, \diff{\prlz}{\elem}$.
Hence $\avgf( \append{\policy}{\policy'}) - \avgf(\policy) \ge 0$ and so $\avgf(\policy)\leq \avgf(\append{\policy}{\policy'}) = \avgf(\append{\policy'}{\policy})$.

We next prove the backward direction, in contrapositive form.
Suppose $\diff{\prlz}{\elem} < 0$ for some $\prlz$ with $\Pr{\rvrlz \sim \prlz} > 0$ and $e \in \groundset$.
Let $e_1, \ldots, e_r$  be the items in $\dom(\prlz)$ and define policies
$\policy$ and $\policy'$ as follows.  For $i = 1, 2, \ldots, r$, both
$\policy$ and $\policy'$ select $e_i$ and observe $\rvrlz(e_i)$.
If either policy observes $\rvrlz(e_i) \neq \prlz(e_i)$ it immediately terminates, otherwise it continues.
If $\policy$ succeeds in selecting all of $\dom(\prlz)$ then it terminates.
If $\policy'$ succeeds in selecting all of $\dom(\prlz)$ then it selects $e$ and then terminates.
We claim $\avgf( \append{\policy}{\policy'}) - \avgf(\policy) < 0$.
Note that
$\played{ \append{\policy}{\policy'} }{\rlz} = \played{\policy }{\rlz}$ unless $\rlz \sim \prlz$, and if
$\rlz \sim \prlz$ then
$\played{ \append{\policy}{\policy'} }{\rlz} = \played{\policy }{\rlz} \cup \set{e}$ and also
$\played{\policy }{\rlz}= \dom(\prlz)$.
Hence
\begin{eqnarray*}
  \label{eq:1app}
  \avgf( \append{\policy}{\policy'}) - \avgf(\policy) & = &
 \expctoverrlz{\rvrlz}{f( \played{ \append{\policy}{\policy'} }{\rvrlz}  , \rvrlz) -
   f(  \played{\policy }{\rvrlz}  , \rvrlz)} \\
 & = & \expctoverrlz{\rvrlz}{f( \played{ \append{\policy}{\policy'} }{\rvrlz}  , \rvrlz) -
   f( \played{\policy }{\rvrlz}   , \rvrlz) \ | \ \rvrlz\sim\prlz }\cdot  \proboverrlz{\rvrlz}{\rvrlz \sim \prlz}  \\
 & = & \expctoverrlz{\rvrlz}{f(  \dom(\prlz) \cup \set{e}, \rvrlz) -
   f( \dom(\prlz) , \rvrlz) \ | \ \rvrlz\sim\prlz } \cdot \proboverrlz{\rvrlz}{\rvrlz \sim \prlz} \\
 & = & \diff{\prlz}{\elem} \cdot \proboverrlz{\rvrlz}{\rvrlz \sim \prlz}
\end{eqnarray*}
The last term is negative, as $\proboverrlz{\rvrlz}{\rvrlz \sim \prlz} > 0$ and $ \diff{\prlz}{\elem} < 0$ by assumption.
Therefore
$\avgf(\policy) > \avgf( \append{\policy}{\policy'}) = \avgf( \append{\policy'}{\policy})$, which completes the proof.
\ignore{
We next prove the backward direction, in contrapositive form.
Suppose $\avgf(\policy) > \avgf( \append{\policy'}{\policy})$ for some $\policy$ and $\policy'$.
Then $\avgf(\policy) > \avgf( \append{\policy}{\policy'})$ and $\avgf( \append{\policy}{\policy'}) - \avgf(\policy) < 0$.
However, as we mentioned previously $\avgf( \append{\policy}{\policy'}) - \avgf(\policy) $ can be written as a conical combination of $\diff{\prlz}{\elem}$ terms.  Since a conical combination of nonnegative terms is nonnegative, this implies there exist some $\prlz$ and $e$
such that $\diff{\prlz}{\elem} < 0$.
}
\end{proof}
\subsection{Adaptive Data Dependent Bounds with Costs} \label{sec:costs-data-dependent-bounds}

The adaptive data dependent bound has the following generalization with costs.

\begin{lemma}[The Adaptive Data Dependent Bound with Costs] \label{lem:rate-equation-with-costs}
Suppose we have made observations $\prlz$ \mbox{after}
selecting $\dom(\prlz)$.  Let $\policy^*$ be any policy.
Then for \term monotone submodular $f:2^\groundset \times
\outcomes^{\groundset} \to \NonNegativeReals$
\begin{equation}
  \label{eq:data-dependent-bound-adaptive-with-costs}
\diff{\prlz}{\policy^*} \ \le \ Z \ \le \
\condcost{\prlz}{\policy^*} \max_{e} \paren{\frac{\diff{\prlz}{e}}{c(e)}}
\end{equation}
where $Z = \max_w \set{ \sum_{e \in \groundset} w_e \,\diff{\prlz}{e} \ :
  \ \sum_e
  c(e) w_e \le \condcost{\prlz}{\policy^*} \text{ and } \forall e \in \groundset, 0 \le w_e \le 1
}$.
\end{lemma}

\begin{proof}
Order the items in $\dom(\prlz)$ arbitrarily, and
consider the policy $\policy$ that for each $e \in \dom(\prlz)$ in
order selects $e$, terminating if $\rvrlz(e) \neq \prlz(e)$ and
proceeding otherwise, and, should it succeed in selecting all of
$\dom(\prlz)$ without terminating (which occurs iff $\rvrlz \sim
\prlz$), then proceeds to run $\policy^*$ as if from a fresh start,
forgetting the observations in $\prlz$.
By construction
the expected marginal benefit of running the $\policy^*$ portion of
$\policy$ conditioned on $\rvrlz \sim \prlz$ equals $\diff{\prlz}{\policy^*}$.
For all $e \in \groundset$, let $w(e) = \Pr{e \in
  \played{\policy}{\rvrlz} \ \mid \ \rvrlz \sim \prlz}$
be the probability that $e$ is selected when running $\policy$,
conditioned on $\rvrlz \sim \prlz$.
Whenever some $e \in \groundset \setminus \dom(\prlz)$ is selected by $\policy$, the current partial realization
$\prlz'$ contains $\prlz$ as a subrealization; hence \term
submodularity implies $\diff{\prlz'}{e} \le \diff{\prlz}{e}$.
It follows that the total contribution of $e$ to $\diff{\prlz}{\policy^*}$
is upper
bounded by $w(e) \cdot \diff{\prlz}{e}$.
Summing over $e  \in \groundset \setminus \dom(\prlz)$,
we get a bound of
$\diff{\prlz}{\policy^*} \le \sum_{e  \in \groundset \setminus
     \dom(\prlz)} w(e) \diff{\prlz}{e}$.
Next, note that each $e \in \groundset \setminus
\dom(\prlz)$ contributes $w(e)c(e)$ cost to $\condcost{\prlz}{\policy^*}$.
Hence it must be the case that
$\sum_{e  \in \groundset \setminus \dom(\prlz)} w(e)c(e) \le \condcost{\prlz}{\policy^*}$.
Obviously, $w(e) \in [0,1]$ for all $e$, since $w(e)$ is a probability.
Hence $\diff{\prlz}{\policy^*} \le \sum_{e  \in \groundset \setminus \dom(\prlz)} w(e)
\diff{\prlz}{e} \le Z$ because setting $w_e = w(e)$ is feasible for the
the linear program for which $Z$ is the optimal value.

To show $Z \le \condcost{\prlz}{\policy^*} \max_{e} \paren{\diff{\prlz}{e} /
  c(e)}$, consider any feasible solution $w$ to the linear program
defining $Z$.  It attains objective value
$$\sum_{e \in \groundset} w_e \,\diff{\prlz}{e} \le \sum_{e \in
  \groundset} w_e c(e) \frac{\diff{\prlz}{e}}{c(e)} \le \sum_{e \in
  \groundset} w_e c(e) \max_{e \in \groundset} \paren{\frac{\diff{\prlz}{e}}{c(e)}} \le
\condcost{\prlz}{\policy^*}  \max_{e \in \groundset} \paren{\frac{\diff{\prlz}{e}}{c(e)}} $$
since $ \sum_{e \in \groundset} w_e c(e) \le \condcost{\prlz}{\policy^*} $ by
the feasibility of $w$.
\end{proof}

A simple greedy algorithm can be used to compute $Z$; we provide
pseudocode for it in~\algref{alg:data-dependent-bound}.
The correctness of this algorithm is more readily discerned upon rewriting
the linear program using variables $x_e = c(e) w_e$ to obtain
$$Z = \max_x \set{ \sum_{e \in \groundset} x_e \paren{\diff{\prlz}{e}/c(e)} \ :
  \ \sum_e
  x_e \le \condcost{\prlz}{\policy^*}\text{ and } \forall e \in \groundset, 0
  \le x_e \le c(e)}.$$
Intuitively, it is clear that to optimize $x$ we should shift mass
towards variables with the highest $\diff{\prlz}{e}/c(e)$ ratio.
Clearly, any optimal solution has $\sum_e x_e = \condcost{\prlz}{\policy^*}$.
Moreover, in any optimal solution,
$\diff{\prlz}{e}/c(e) > \diff{\prlz}{e'}/c(e')$ implies
$x_e = c(e)$ or $x_{e'} = 0$,
since otherwise it would be possible to shift mass from $x_{e'}$ to
$x_{e}$ and obtain an increase in objective value.
If the $\diff{\prlz}{e}/c(e)$ values are distinct for distinct items,
there will be a unique solution satisfying these constraints, which
\algref{alg:data-dependent-bound} will compute.
Otherwise, we imagine perturbing each $\diff{\prlz}{e}$ by
independent random quantities $\epsilon_e$ drawn uniformly from
$[0,\epsilon]$ to make them distinct.  This changes the optimum value
by at most $|\groundset| \epsilon$, which vanishes as we let $\epsilon $ tend
towards zero.  Hence any solution satisfying $\sum_e x_e = \condcost{\prlz}{\policy^*} $
and $\diff{\prlz}{e}/c(e) > \diff{\prlz}{e'}/c(e')$ implies
$x_e = c(e)$ or $x_{e'} = 0$ is optimal.
Since \algref{alg:data-dependent-bound} outputs the value of such a
solution, it is correct.

\begin{algorithm}
 \KwIn{Groundset $\groundset$; Partial realization $\prlz$; Costs
   $c:\groundset \to \nats$;  Budget $C = \condcost{\prlz}{\policy^*}$;
   Conditional expected marginal benefits $\diff{\prlz}{e}$ for all $e \in \groundset$.}
\KwOut{\ignore{Data dependent bound} $Z = \max_w \set{ \sum_{e \in \groundset} w_e \,\diff{\prlz}{e} \ :
  \ \sum_e
  c(e) w_e \le \condcost{\prlz}{\policy^*} \text{ and } \forall e \in \groundset, 0 \le w_e \le 1
}$}
\Begin{
  \mbox{Sort $\groundset$ \ignore{in decreasing order} by
    $\diff{\prlz}{e}/c(e)$, so that $\frac{\diff{\ \prlz}{e_1\ }}{c(e_1)}
  \ge \frac{\diff{\ \prlz}{e_2\ }}{c(e_2)} \ge \ldots \ge
  \frac{\diff{\ \prlz}{e_n\ }}{c(e_n)}$}\;

   Set $w \gets \mathbf{0}$;  $i \gets 0$; $a \gets 0$; $z \gets 0$; $e \gets \NULL$\;
   \While{$a < C$}{
       $i \gets i+1$; \ \  $e \gets e_i$\;
       $w_e \gets \min \set{1, C - a}$\;
       $a \gets a + c(e) w_e$; \ \
       $z \gets z + w_e \diff{\prlz}{e}$;
   }
   Output $z$;
 } \label{alg:data-dependent-bound} \caption{Algorithm to compute the
   data dependent bound $Z$ of Lemma~\ref{lem:rate-equation-with-costs}.}
\end{algorithm}

\subsection{The Max-Cover Objective} \label{sec:proofs-max-cover}

With item costs, the Adaptive Stochastic Maximization problem becomes one of finding some
\begin{equation}
\policy^{*}\in\argmax_{\policy} \avgf(\prune{\policy}{\budget})
\label{eq:stochmax-costs}
\end{equation} where $\budget$ is a budget on the cost of selected
items, and we define $\avgf(\policy)$ for a randomized policy
$\policy$ to be
$\avgf(\policy) := \expct{f(\played{\policy}{\rvrlz}, \rvrlz)}$ as before,
where the expectation is now over both $\rvrlz$ and the internal
randomness of $\policy$ which determines
$\played{\policy}{\rlz}$ for each $\rlz$.
We prove the following generalization of~\thmref{thm:max-cover}.

\begin{theorem} \label{thm:max-cover-with-costs}
Fix any $\alpha \ge 1$ and item costs $c:\groundset \to \nats$.
If $f$ is \term monotone and \term submodular with respect to the
distribution
$\rlzprior$, and $\policy$ is an
$\alpha$-approximate greedy policy, then for all policies $\policy^*$ and positive
integers $\ell$ and $k$
\[
\avgf(\prune{\policy}{\ell}) > \paren{1 - e^{-\ell/\alpha k}}
\avgf(\prune{\policy^*}{k})
\mbox{.}
\]
\end{theorem}

\begin{proof}
The proof goes along the lines of the performance analysis of the
greedy algorithm for maximizing a submodular function subject to
a cardinality constraint of~\citet{nemhauser78}.  An
extension of that analysis to $\alpha$-approximate greedy algorithms,
which is analogous to ours but for the nonadaptive case, is shown by~\citet{goundan07}.
For brevity, we will assume without loss of generality that
 $\policy = \prune{\policy}{\ell}$ and $\policy^* = \prune{\policy^*}{k}$.
Then for all $i$, $0 \le i < \ell$
\begin{equation}
  \label{eqn:ee1}
  \avgf(\policy^*) \ \ \le \ \ \avgf(
  \append{\prune{\policy}{i}}{\policy^*}) \ \ \le \ \ \avgf(\prune{\policy}{i}) + \alpha k \paren{\avgf(\prune{\policy}{i+1}) - \avgf(\prune{\policy}{i})}.
\end{equation}
The first inequality is due to the \term monotonicity of $f$ and~\lemref{lem:monotonicity-equivalence}, from which we
  may infer $\avgf(\policy_2) \le \avgf(\append{\policy_1}{\policy_2})$ for any $\policy_1$ and $\policy_2$.
The second inequality may be obtained as
a corollary of~\lemref{lem:rate-equation-with-costs} as follows.
Fix any partial realization $\prlz$ of the form
$\set{(e, \rlz(e)) : e \in \played{\prune{\policy}{i}}{\rlz}}$
for some $\rlz$.
Consider $\diff{\prlz}{\policy^*}$, which equals
the expected marginal benefit of the $\policy^*$ portion of
$\append{\prune{\policy}{i}}{\policy^*}$ conditioned on $\rvrlz \sim
\prlz$.
\lemref{lem:rate-equation-with-costs} allows us to bound it as
$$\expct{\diff{\prlz}{\policy^*}} \ \le \ \expct{\condcost{\prlz}{\policy^*}} \cdot
\max_{e} \paren{\diff{\prlz}{e}/c(e)},$$
where the expectations are taken
over the internal randomness of $\policy^*$, if there is any.  Note
that since $\policy^*$ has the form $\prune{\policy'}{k}$ for some
$\policy'$ we know that for all $\rlz$,
$\expct{c(\played{\policy^*}{\rlz})} \le k$, where the
expectation is again taken over the internal randomness of $\policy^*$.
Hence $\expct{\condcost{\prlz}{\policy^*}} \le k$ for all $\prlz$.  It
follows that $\expct{\diff{\prlz}{\policy^*}} \le k \cdot
\max_{e} \paren{\diff{\prlz}{e}/c(e)}$.
By definition of an $\alpha$-approximate greedy policy, $\policy$
obtains at least $(1/\alpha)\max_{e}
\paren{\diff{\prlz}{e}/c(e)} \ge \expct{\diff{\prlz}{\policy^*}} / \alpha k$
expected marginal benefit per unit cost
 in a step immediately following its observation of $\prlz$.
Next we take an appropriate convex combination of the previous inequality
with different values of $\prlz$.
Let $\rvprlz$ be a random partial realization distributed as
$\rlzmassover{\rvprlz}{\prlz} := \prob{\rvprlz = \prlz \mid \exists \rlz.\ \prlz = \set{(e, \rlz(e)) : e \in \played{\prune{\policy}{i}}{\rlz}}}$
Then
\begin{eqnarray*}
  \avgf(\prune{\policy}{i+1}) -  \avgf(\prune{\policy}{i}) & \ge & \expct{ \frac{1}{\alpha}\,\max_{e}
\paren{\frac{\diff{\rvprlz}{e}}{c(e)}} } \\
 &  \ge & \expct{
\frac{\expct{\diff{\rvprlz}{\policy^*}}}{\alpha k} }\\
  & = & \frac{ \avgf(
    \append{\prune{\policy}{i}}{\policy^*})  -
    \avgf(\prune{\policy}{i})} {\alpha k}
\end{eqnarray*}
\noindent
A simple rearrangement of terms then yields the second inequality in~(\ref{eqn:ee1}).

Now define $\Delta_i := \avgf(\policy^*) - \avgf(\prune{\policy}{i})$,
so that~(\ref{eqn:ee1}) implies $\Delta_i \le \alpha k(\Delta_i
  - \Delta_{i+1})$, from which we infer
  $\Delta_{i+1} \le \paren{1-\frac{1}{\alpha k}}\Delta_{i}$ and hence
$\Delta_{\ell} \le \paren{1 - \frac{1}{\alpha k}}^{\ell} \Delta_0 <
e^{-\ell/\alpha k} \Delta_0$, where for this last inequality we have used
the fact that $1 - x < e^{-x}$ for all
$x > 0$.
Thus
$\avgf(\policy^*) - \avgf(\prune{\policy}{\ell}) < e^{-\ell/\alpha k} \paren{\avgf(\policy^*) -
  \avgf(\prune{\policy}{0})} \le e^{-\ell/\alpha k} \avgf(\policy^*)$ so
$\avgf(\policy) > (1-e^{-\ell/\alpha k})\avgf(\policy^*)$.

\end{proof}

\subsection{The Min-Cost-Cover Objective} \label{sec:proofs-min-cost-cover}

In this section, we provide arbitrary item cost generalizations
of~\thmref{thm:min-set-cover-avg-generalized} and
\thmref{thm:min-set-cover-wc-generalized}.
With item costs the Adaptive Stochastic Minimum Cost Cover problem
becomes one of finding, for some quota on utility $\quota$,
\begin{equation}
\policy^{*}\in\argmin_{\policy} \acst{\policy}\text{ such that }
f(\played{\policy}{\rlz},\rlz)\geq \quota\text{ for all
}\rlz,\label{eq:stochcover-with-costs}
\end{equation}
where  $\acst{\policy}:=\expctoverrlz{\rlz}{c(\played{\policy}{\rvrlz})}$.
Without loss of generality, we may take a truncated version of $f$,
namely $(A, \rlz) \mapsto \min \set{\quota, f(A, \rlz)}$, and
rephrase Problem~(\ref{eq:stochcover-with-costs}) as finding
\begin{equation}
  \label{eq:stochcover-with-costs-coverage}
  \policy^{*}\in\argmin_{\policy} \acst{\policy}\text{ such that }
\policy \text{ covers }\rlz \text{ for all
}\rlz.
\end{equation}
Hereby, recall that $\policy$ covers $\rlz$ if
$\expct{f(\played{\policy}{\rlz},\rlz)} = f(\groundset,\rlz)$,
where the expectation is over any internal randomness of $\policy$.
We will consider only
Problem~(\ref{eq:stochcover-with-costs-coverage}) for the
remainder.
We also consider the worst-case variant of this problem, where we
replace the expected cost $\acst{\policy}$ objective with the worst-case cost
$\wcst{\policy} := \max_{\rlz}{c(\played{\policy}{\rlz})}$.

The definition of coverage (Definition~\ref{def:coverage}
in~\secref{sec:min-cost-cover} on page~\pageref{def:coverage}) requires no
modification to handle item costs.
Note, however, that coverage is all-or-nothing in the sense that
covering a realization $\rlz$ with probability less than one does not count
as covering it.  A corollary of this is that
only items whose runs have finished help with coverage,
whereas currently running items do not.  For a simple example,
consider the case where $\groundset = \set{e}$, $c(e) = 2$,
$f(A, \rlz) = |A|$, and
policy $\policy$ that selects $e$ and then
terminates.  Then $\prune{\policy}{1}$ is a randomized policy which is
$\policy$ with probability $\frac12$, and is the empty policy with
probability $\frac12$, so
$\expct{f(\played{\policy}{\rlz},\rlz)} = \frac12 < 1 =
f(\groundset,\rlz)$ for each $\rlz$.  Hence, even though half the time
$\prune{\policy}{1}$ covers all realizations, it is counted as not
covering any. \\

\noindent
We begin with a claim relating pointwise submodularity to strong adaptive
submodularity.

\begin{lemma}\label{lem:pointwise-submod}
  If $f$ is adaptive submodular with respect to $\rlzprior$ and $f$ is pointwise
  submodular meaning $S \mapsto f(S, \rlz)$ is submodular for all $\rlz$,
  then $f$ is strongly adaptive submodular with respect to $\rlzprior$.
\end{lemma}

\begin{proof}
  By assumption $f$ is adaptive submodular with respect to $\rlzprior$, so it is
  sufficient to prove that \eqnref{eqn:strong-adapt-submod} holds, i.e.,
$\diff{\prlz ; \prlz'}{\elem} \ge \diff{\prlz'}{\elem}$.
Fix any $\prlz \subseteq \prlz'$ and $\elem \in \groundset$.
Let $\delta_{\rlz}(e, S) := f(S \cup \set{e}, \rlz) - f(S, \rlz)$.
From the definition of $\diff{\prlz ; \prlz'}{\elem}$ and $\diff{\prlz'}{\elem}$,
we have
\begin{eqnarray*}
\diff{\prlz ; \prlz'}{\elem} & = & \sum_{\rlz} \prob{\rlz \mid \prlz'} \delta_{\rlz}(e, \dom(\prlz))  \\
 & \ge & \sum_{\rlz} \prob{\rlz \mid \prlz'} \delta_{\rlz}(e, \dom(\prlz')) \\
 & = & \diff{\prlz'}{\elem}
\end{eqnarray*}
where we have used
$\delta_{\rlz}(e, \dom(\prlz)) \ge \delta_{\rlz}(e, \dom(\prlz'))$ by
the pointwise submodularity of $f$.
\end{proof}

\noindent
Now we provide an approximation guarantee for the average-case
policy cost with arbitrary item costs.

\begin{theorem} \label{thm:min-set-cover-avg-generalized-with-costs}
Suppose $f:2^{\groundset} \times \outcomes^{\groundset} \to
\NonNegativeReals$ is strongly adaptive submodular and strongly adaptive
monotone with respect to $\rlzprior$ and there exists $Q$ such that
$f(\groundset, \rlz) = Q$ for all $\rlz$.
Let $\eta$ be any value such that
$f(S, \rlz) > Q - \eta$ implies $f(S, \rlz) = Q$ for all $S$ and
$\rlz$.
Let $\delta = \min_{\rlz} \rlzprior$ be the minimum probability of
any realization.
Let $\policycover$ be an optimal policy
minimizing the expected cost of items selected
to guarantee every realization is covered.
Let $\policy$ be an $\alpha$-approximate
greedy policy with respect to the item costs.
Then in general
\[
\acst{\policy} \le  \alpha\,
\acst{\policycover}\paren{\ln \paren{\frac{Q}{\delta \eta}} + 1}^2
\]
and for \certifying instances
\[
\acst{\policy} \le  \alpha\,
\acst{\policycover}\paren{\ln \paren{\frac{Q}{\eta}} + 1}^2
\mbox{.}
\]
Note that if $\range(f) \subset \integers$, then $\eta = 1$ is a valid
choice, so for general and
\certifying instances we have $\acst{\policy} \le  \alpha\,\acst{\policycover}\paren{\ln (Q/\delta) + 1}^2$
and $\acst{\policy} \le
\alpha\,\acst{\policycover}\paren{\ln (Q) + 1}^2$, respectively.
%
\end{theorem}

%
%


\newcommand{\arcs}{\ensuremath{A}}
\newcommand{\xarcs}{\ensuremath{\arcs_x}}
\newcommand{\xsources}{\ensuremath{U_x}}
\newcommand{\xtargets}{\ensuremath{V_x}}
\newcommand{\xsucc}{\ensuremath{S_x}}  
\newcommand{\acondcost}[2]{\ensuremath{\hat{c}\paren{ {#2} \! \mid \! {#1} }  }}
\newcommand{\altcost}[3]{\ensuremath{\hat{c}_{#1}\paren{#2, #3}}}
\newcommand{\altcondcost}[2]{\ensuremath{\hat{c}\paren{ {#2} \! \mid \! {#1} }  }}
\newcommand{\acavg}[1]{{\hat{c}_{\text{avg}}{(#1)}}}

\subsection*{Proof of Theorem~\ref{thm:min-set-cover-avg-generalized-with-costs}}

We first require some additional definitions.
We extend $\obj$ by defining
$\obj(\groundsubset, \prlz)$ via
\begin{equation}
  \label{eq:objective-ext}
  \obj(\groundsubset, \prlz) := \expct{f(\groundsubset, \rvrlz) \mid \rvrlz \sim \prlz}
\end{equation}
Further, let
\begin{equation} \label{eqn:short-expect-value}
 f(\prlz) := f(\dom(\prlz), \prlz) = \expctoverrlz{\rvrlz}{f(\dom(\prlz), \rvrlz) \ \mid \ \rvrlz \sim \prlz}
\end{equation}

\begin{definition}[Execution Trace]
Given a policy $\policy$ and a realization $\rlz$, let the
\emph{trace} $\trace = \trace(\rlz)$ of
$\policy$ under $\rlz$ be the sequence of partial realizations specifying
its set of observations through time, i.e.,
$\prlz_0 \subset \prlz_1 \subset
\cdots \subset \prlz_{\ell}$, such that $\dom(\prlz_{i}) \setminus
\dom(\prlz_{i-1})$ consists precisely of the $\ith$ item selected by
$\policy$.
\end{definition}

\begin{definition}[Execution Arborescence]
For a policy $\policy$,  define its \emph{execution arborescence} as the
digraph
$(\dom(\policy), \arcs)$ with vertices $\dom(\policy)$ and edges
\[
\arcs := \set{(\prlz, \prlz') : \exists \rlz, \prlz' = \prlz \cup \set{(\policy(\prlz), \rlz(\policy(\prlz)))}}.
\]
In other words, its vertices are the partial realizations that may be encountered by $\policy$, and it has a directed edge from $\prlz$ to $\prlz'$ if it may immediately encounter $\prlz'$ after $\prlz$ in some execution trace.
\end{definition}

\begin{definition}[Spanning Edges, Sources, Targets, and Successors]
\label{def:graph-defs}
Given a policy $\policy$ with execution arborescence with arcs $\arcs$, and some positive $x \in \Real$, let $\xarcs$ be
the edges that span $x$ expected benefit, i.e.,
\[
\xarcs := \set{(\prlz, \prlz') : (\prlz, \prlz') \in \arcs,\ f(\prlz) < x \le f(\prlz') }
\]
Further let $\xsources$ be the \emph{sources} of edges in $\xarcs$ and $\xtargets$ be the \emph{targets}.
That is, $\xsources := \set{\prlz : \exists \prlz', (\prlz, \prlz') \in \xarcs}$ and
$\xtargets := \set{\prlz' : \exists \prlz, (\prlz, \prlz') \in \xarcs}$.
Finally, the \emph{successors} $\xsucc(\prlz)$ of some $\prlz \in \xsources$ are
the sources $\prlz'$ that are minimal supersets of $\prlz$ in $\xsources$ when
interpreting partial realizations as set of (item, observation) pairs.  Formally,
\[
\xsucc(\prlz) := \set{\prlz' : \prlz' \in \xsources, \nexists \prlz'',  \prlz \subset \prlz'' \subset \prlz'}
\]
\end{definition}

\newcommand{\concat}[2]{\append{{#1}[x]}{#2}}

\begin{definition}[Concatenative Pseudopolicies] \label{def:psuedopolicies}
  Given two policies $\policy$ and $\policy^*$ and some nonnegative $x \in
  \Real$, we define the \emph{concatenative pseudopolicy}
  $\concat{\policy}{\policy^*}$ as the stochastic process that, on each
  realization $\rlz$, runs policy $\policy$ until it is just about to achieve $x$ expected reward
  (formally, until reaching $\ptrace{\rlz}{x}$),
  and then runs $\policy^*$ as if from scratch.
  Hence on realization $\rlz$ it will play
  $\dom(\ptrace{\rlz}{x}) \cup \played{\policy^*}{\rlz}$ where $\ptrace{\rlz}{x}$ is defined relative to
  $\policy$.
\end{definition}

Note that $\concat{\policy}{\policy^*}$ isn't a proper policy, because it decides when to switch to executing
$\policy^*$ based on information about $\rlz$ that it cannot infer from its current partial realization.
Specifically, upon reaching $\prlz \in \xsources$ it will select $\policy^*(\emptyset)$ if there exists
$\prlz'$ such that $(\prlz, \prlz') \in \xarcs$ and $\rlz \sim \prlz'$, and will select
$\policy(\prlz)$ otherwise.

%
%

\subsubsection*{The Strategy}
Our overall strategy will be to bound the expected cost $\acst{\policy}$ of
$\policy$ by bounding the price $\price$ it pays per unit of
\emph{expected reward}
gained as it runs (measured as $\progress{\prlz}$ where $\prlz$ is the current
partial realization) and then integrating over the run.  To bound the price
$\policy$ pays, we define an alternative cost scheme for policies, bound the
price $\policy$ pays in terms of the alternative cost of an optimal policy
$\piavg$, and finally bound the alternative cost of $\piavg$ in terms of its
actual cost.


\subsubsection*{Progress bounds}

First, observe that
\begin{equation} \label{eqn:reward-from-opt}
\diff{\prlz; \prlz'}{\piavg} \ge Q - x
\text{ for all } (\prlz, \prlz') \in \xarcs
\end{equation}
This holds because
$f(\prlz) < x$ by definition of $\xarcs$, and
$f(\played{\piavg}{\rlz}, \rlz) = Q$ for all $\rlz$ since $\piavg$ covers
every realization.  Since $Q$ is the maximum possible reward,
if $\diff{\prlz; \prlz'}{\piavg} < Q - x$ then we can generate a
violation of strong \term monotonicity by
fixing some $\rlz' \sim \prlz'$, selecting
$\played{\piavg}{\rlz'}$, and then selecting
$\dom(\prlz)$ to reduce the expected reward.

The bound in~\eqnref{eqn:reward-from-opt} is useful when $x$ is far from
$Q$, however we will require a stronger bound for $x$ very close to $Q$.
We will use slightly different analyses for general instances and for
\certifying instances.

We begin with general instances.
For these, we will prove a stronger bound for obtaining the final
$\delta\eta$ reward, i.e., for $x \in [Q - \delta\eta, Q]$.
\begin{equation} \label{eqn:reward-from-opt2}
\diff{\prlz; \prlz'}{\piavg} \ge \delta \eta
\text{ for all } (\prlz, \prlz') \in \xarcs
\end{equation}
Fix $\prlz \in \dom(\policy)$ and any $\rlz' \sim \prlz$.
We say $\prlz$ \emph{covers} $\rlz'$ if
$\policy$ covers $\rlz'$ by the time it observes $\prlz$.
By definition of $\delta$ and $\eta$, if some $\rlz' \sim \prlz'$ is
not covered by $\prlz$ then
$Q - f(\dom(\prlz), \prlz') \ge \delta \eta$.
Hence the last item that $\policy$ selects, say upon observing
$\prlz$, must increase its conditional expected
value from $f(\dom(\prlz), \prlz') \le Q - \delta \eta$ to $Q$
for all $\rlz \sim \prlz'$.

For \certifying instances we show that
\begin{equation} \label{eqn:reward-from-opt3}
\diff{\prlz; \prlz'}{\piavg} \ge \eta
\text{ for all } (\prlz, \prlz') \in \xarcs
\end{equation}
using a similar argument for $x \in [Q-\eta, Q]$.
We first argue that
the last item that $\policy$ selects must increase its conditional expected
value from at most $Q - \eta$ to $Q$.
For suppose $\policy$ currently observes $\prlz$, and has not achieved
conditional value $Q$, i.e., $f(\prlz) < Q$.
Then some $\rlz \sim \prlz$ is uncovered.
Since the instance is \certifying, \emph{every} $\rlz$ with
$\rlz \sim \prlz$ is uncovered, and has
$f(\dom(\prlz), \rlz) < f(\groundset, \rlz)  =
Q$.  By definition of $\eta$, for each $\rlz$ with
$\rlz \sim \prlz$ we then have $f(\dom(\prlz), \rlz) \le Q - \eta$, which
implies $f(\dom(\prlz), \prlz') \le Q - \eta$.

\subsubsection*{Upper bounding the price $\policy$ pays with Alternate Costs}

Next we upper-bound the price $\policy$ pays in terms of an alternative pricing
scheme, $\hat{c}$, for $\piavg$. Note the cost of $\piavg$ under $\hat{c}$
may be higher than $\acst{\piavg}$.
We will eventually bound the alternative cost in terms of the true cost.

Fix $x$ and any $e$ that may be played by $\piavg$
under some realization $\rlz$.
Consider any $(\prlz, \prlz') \in \xarcs$ with $\rlz \sim \prlz'$.
Fix some $\prlz_e \in \dom(\piavg)$ such that $e = \piavg(\prlz_e)$.
We charge $e$ a price of $c(e) / \diff{\prlz \cup \prlz_e}{e}$
per-unit reward in this case.
Hence, under realization $\rlz$,
the alternative cost is
\[
c(e) \paren{
\frac{f\paren{\dom(\prlz) \cup \dom(\prlz_e) \cup \set{e}, \rlz} -
f\paren{\dom(\prlz) \cup \dom(\prlz_e), \rlz}}{\diff{\prlz \cup \prlz_e}{e}}
}
\]
Taking the expectation over $\rlz$ such that $\rlz \sim \prlz'$ and $\rlz \sim \prlz_e$ (which is equivalent to $\rlz \sim  \prlz' \cup \prlz_e$) yields an
 expected alternative cost of
\begin{equation} \label{eqn:expected-alt-cost}
\altcost{e}{\prlz}{\prlz'} := c(e) \paren{
\frac{
\diff{\prlz \cup \prlz_e; \prlz' \cup \prlz_e}{e}
}{\diff{\prlz \cup \prlz_e}{e}}
}
\end{equation}
For brevity, we will define
\[
\Delta_{a, e} := \diff{\prlz \cup \prlz_e; \prlz' \cup \prlz_e}{e}
\]
for arc
$a = (\prlz, \prlz') \in \xarcs$ and $\prlz_e$.
Note we could have multiple $\prlz_e$ such that $\piavg(\prlz_e) = e$,
however for notational convenience we can imagine creating separate copies of
$e$ for each such partial realization and assume without loss of generality that
for each item there is a unique partial realization that $\piavg$ plays it
under; That is, for each $e$ there is at most one $\prlz_e$ such that $\piavg(\prlz_e) = e$.

Next, for arc $a = (\prlz, \prlz')$ define $w_{a, e}$ to be the probability
that $\piavg$ plays $e$ conditioned on $\prlz'$:
\[
w_{a, e} := \prob{e \in \played{\piavg}{\rvrlz} \mid \rvrlz \sim \prlz'} =
\prob{\prlz_e \mid \prlz'}
\]

Next, consider the cost $\acondcost{\prlz; \prlz'}{\piavg}$ which is the expected
alternative cost paid by $\piavg$ starting from $\dom(\prlz)$
and conditioned on $\prlz'$ for some fixed
$x$ and $a = (\prlz, \prlz') \in \xarcs$.
Note that since the alternative cost depends on $x$ and $a$, this alternative
cost does as well.
Formally,
\[
\acondcost{\prlz; \prlz'}{\piavg} := \sum_{e} w_{a,e} \altcost{e}{\prlz}{\prlz'}
\]
The expected benefit of running $\piavg$ starting from $\prlz$ and conditioning on $\prlz'$ is
$\diff{\prlz; \prlz'}{\piavg} := \sum_{e} w_{a,e} \Delta_{a,e}$.
Hence
\begin{eqnarray}
\acondcost{\prlz; \prlz'}{\piavg} & = &
       \sum_{e} w_{a,e} \altcost{e}{\prlz}{\prlz'} \\
 & = & \sum_{e} w_{a,e} c(e)
               \paren{\frac{\Delta_{a,e}}{\diff{\prlz \cup \prlz_e}{e}}}\\
 & \ge & \sum_{e} w_{a,e} \Delta_{a,e}
   \min_{e'} \paren{\frac{c(e')}{\diff{\prlz \cup \prlz_e}{e'}}} \\
 & \ge & \sum_{e} w_{a,e} \Delta_{a,e}
   \min_{e'} \paren{\frac{c(e')}{\diff{\prlz}{e'}}} \\
 & = & \diff{\prlz; \prlz'}{\piavg} \min_{e'} \paren{\frac{c(e')}{\diff{\prlz}{e'}}}
\end{eqnarray}
Hence
\begin{equation}
   \label{eqn:bound-fix1}
\min_{e'} \paren{\frac{c(e')}{\diff{\prlz}{e'}}} \le
\frac{ \acondcost{\prlz; \prlz'}{\piavg} }{ \diff{\prlz; \prlz'}{\piavg} }
\end{equation}

Next, note that
$\xtargets := \set{\prlz' : \exists \prlz, (\prlz, \prlz') \in \xarcs}$
partitions the set of realizations.
We define the alternative cost of $\piavg$ as
\begin{equation}
   \label{eqn:alt-avg-cost}
\acavg{\piavg} :=
\sum_{(\prlz, \prlz') \in \xarcs} \acondcost{\prlz; \prlz'}{\piavg} \prob{\prlz'}
\end{equation}
Note it too is implicitly dependent on $x$.

\subsubsection*{Bounding the Alternate Costs}

Next we bound $\acavg{\piavg}$ in terms of $\cavg{\piavg}$ and problem
parameters $Q, \eta$ and $\delta$.
Ultimately we will show that for general instances
\begin{equation}
   \label{eqn:bound-fix2}
\acavg{\piavg} \le \cavg{\piavg} \paren{\ln \paren{\frac{Q}{\delta \eta}} + 1}
\end{equation}
and for \certifying instances
\begin{equation}
   \label{eqn:bound-fix2-cert}
\acavg{\piavg} \le \cavg{\piavg} \paren{\ln \paren{\frac{Q}{\eta}} + 1}.
\end{equation}

\noindent
It is sufficient to prove this bound element by element.
That is, we define
\[
\hat{c}(e) := \sum_{(\prlz, \prlz') \in \xarcs} \altcost{e}{\prlz}{\prlz'} \prob{\prlz'}
\]
and prove $\hat{c}(e) / c(e) \le \paren{\ln \paren{\frac{Q}{\delta \eta}} + 1}$
for general instances, and $\hat{c}(e) / c(e)
\le \paren{\ln \paren{\frac{Q}{\eta}} + 1}$ for \certifying instances.




\newcommand{\Deltap}[1]{\ensuremath{\Delta[{#1}]}}
\newcommand{\deltap}[1]{\ensuremath{\delta[{#1}]}}
\newcommand{\qp}[1]{\ensuremath{q[{#1}]}}
\newcommand{\qpp}[1]{\ensuremath{q'[{#1}]}}
\newcommand{\prlzj}{\ensuremath{{\prlz_v}}}
\newcommand{\epsp}[1]{\ensuremath{\eps({#1})}}

To simplify the exposition 
we define some additional notation.
Fix $\policy$, $\piavg$, $x$, and $e$.
Let $\qp{\prlz} := \prob{\prlz  \mid \prlz_e}$ and let
\[
\qpp{\prlz} := \sum_{\prlz' : (\prlz, \prlz') \in \xarcs} \qp{\prlz'}
\]

Let $\Deltap{\prlz} := \diff{\prlz \cup \prlz_e}{e}$.
Let
\[
  \deltap{\prlz} := \sum_{\prlz' : (\prlz, \prlz') \in \xarcs}
  \paren{\frac{\qp{\prlz'}}{\qpp{\prlz}}} \diff{\prlz \cup \prlz_e; \prlz' \cup \prlz_e}{e}
\]
be the conditional expected benefit of selecting $e$
in the concatenative pseudopolicy $\concat{\policy}{\policy^*}$
immediately upon seeing $\prlz$ (see definition~\ref{def:psuedopolicies}).

By construction, $\set{\prlz' : \prlz' \in \xtargets}$
partitions the set of all
realizations (since for each $\rlz$ there is a unique edge in the execution
arborescence where the policy passes $x$ expected reward in the trace
$\trace\paren{\rlz}$ as the policy runs). Furthermore, conditioned on
$\exists (\prlz, \prlz') \in \xarcs$ such that $\rvrlz \sim \prlz'$,
we pay
$c(e) \deltap{\prlz} / \Deltap{\prlz}$. Taking the expectation over
$\prlz \in \xsources$, we obtain
\begin{equation}
   \label{eqn:ratio-bound1}
\frac{\hat{c}(e)}{c(e)} = \sum_{\prlz \in \xsources} \frac{\qpp{\prlz} \deltap{\prlz}}{\Deltap{\prlz}}
\end{equation}

Next we need a subclaim about how $\Deltap{\prlz}$ relates to to
$\Deltap{\prlz'}$ for $\prlz \subset \prlz'$.  Recall the definition of
$\xsucc(\prlz)$ (from definition~\ref{def:graph-defs}).

\begin{lemma} \label{lem:deltas-potential1}
  Fix $\policy$, $\piavg$, $x$, and $e$ and let $p_i$, $p'_i$, $\Deltap{\prlz}$, and
  $\deltap{\prlz}$ be as above. For any $\prlz \in \xsources$,
  if for some $\epsilon \in \reals$
\[
  \qpp{\prlz} \deltap{\prlz} \ge
  \paren{\qpp{\prlz} + \epsilon\paren{\qp{\prlz} - \qpp{\prlz}} } \Deltap{\prlz}
\]
then
\[
  \expct{\Deltap{\prlz'} \mid \prlz' \in \xsucc(\prlz)} := \sum_{\prlz' \in \xsucc(\prlz)} \prob{\prlz' \cup \prlz_e \mid \prlz  \cup \prlz_e} \Deltap{\prlz'} \ \le \
  \Deltap{\prlz} \paren{1 - \epsilon}
\]
\end{lemma}

\begin{proof}
Fix $\prlz$.
We claim that by strong adaptive submodularity,
\begin{equation} \label{eqn:basic-bound1}
 \qp{\prlz} \Deltap{\prlz} \ \ge \ \qpp{\prlz} \deltap{\prlz} + \sum_{\prlzj \in \xsucc(\prlz)} \qp{\prlzj} \Deltap{\prlzj}
\end{equation}

To see this, note that $\set{\prlz' : (\prlz, \prlz') \in \xarcs} \cup \set{\prlzj \in \xsucc(\prlz)}$
partitions the set of realizations, and thus
\begin{eqnarray} \label{eqn:basic-bound2}
 \qp{\prlz} \Deltap{\prlz} & = & \qpp{\prlz} \deltap{\prlz} + \sum_{\prlzj \in \xsucc(\prlz)} \prob{\prlzj \mid \prlz_e}
       \diff{\prlz \cup \prlz_e; \prlzj \cup \prlz_e}{e}  \\
 & \ge & \qpp{\prlz} \deltap{\prlz} +
         \sum_{\prlzj \in \xsucc(\prlz)} \prob{\prlzj \mid \prlz_e} \diff{\prlzj \cup \prlz_e}{e}  \\
 & \equiv & \qpp{\prlz} \deltap{\prlz} + \sum_{\prlzj \in \xsucc(\prlz)} \qp{\prlzj} \Deltap{\prlzj}
\end{eqnarray}

Given~\eqnref{eqn:basic-bound1} and the assumption that
$\qpp{\prlz} \deltap{\prlz} \ge \paren{\qpp{\prlz} + (\qp{\prlz} - \qpp{\prlz})\epsilon } \Deltap{\prlz}$,
we have
\[
\qp{\prlz} \Deltap{\prlz} \ \ge \ \paren{\qpp{\prlz} + (\qp{\prlz} - \qpp{\prlz})\epsilon } \Deltap{\prlz} + \sum_{\prlzj \in \xsucc(\prlz)} \qp{\prlzj} \Deltap{\prlzj}
\]
Hence
\[
\sum_{\prlzj \in \xsucc(\prlz)} \qp{\prlzj} \Deltap{\prlzj} \le (\qp{\prlz} - \qpp{\prlz})(1-\epsilon)\Deltap{\prlz}
\]
Dividing by $(\qp{\prlz} - \qpp{\prlz})$ yields
\[
\sum_{\prlzj \in \xsucc(\prlz)} \prob{\prlzj \mid \prlz \cup \prlz_e} \Deltap{\prlzj}
= \sum_{\prlzj \in \xsucc(\prlz)} \paren{\frac{\qp{\prlzj}}{\qp{\prlz} - \qpp{\prlz}}} \Deltap{\prlzj}
\le (1-\epsilon)\Deltap{\prlz}
\]
which completes the proof.
\end{proof}

Recall we wish to upper-bound $\frac{\hat{c}(e)}{c(e)}$ using
\eqnref{eqn:ratio-bound1}.
Let $\epsp{\prlz} \in \reals$ be set such that
\[
\qpp{\prlz} \deltap{\prlz} =
  \paren{\qpp{\prlz} +
  \epsp{\prlz} \paren{\qp{\prlz} - \qpp{\prlz}} } \Deltap{\prlz}
\]
for all $\prlz$.
Then
\begin{eqnarray}
\frac{\hat{c}(e)}{c(e)} & = & \sum_{\prlz \in \xsources} \frac{\qpp{\prlz} \deltap{\prlz}}{\Deltap{\prlz}}\\
 & = & \sum_{\prlz \in \xsources} \paren{\qpp{\prlz} +
  \epsp{\prlz} \paren{\qp{\prlz} - \qpp{\prlz}} } \\
 & = & \sum_{\prlz \in \xsources} \qpp{\prlz} + \sum_{\prlz \in \xsources}
  \epsp{\prlz} \paren{\qp{\prlz} - \qpp{\prlz}} \\
 & = & 1 + \sum_{\prlz \in \xsources}
  \epsp{\prlz} \paren{\qp{\prlz} - \qpp{\prlz}} \\
 & \le & 1 + \sum_{\prlz \in \xsources}
  \epsp{\prlz}^+ \paren{\qp{\prlz} - \qpp{\prlz}} \\
\end{eqnarray}
where $\epsp{\prlz}^+ := \max \paren{0, \epsp{\prlz}}$.

\begin{lemma} \label{lem:deltas-potential2}
For general instances
\[
\sum_{\prlz \in \xsources}
  \epsp{\prlz}^+ \paren{\qp{\prlz} - \qpp{\prlz}} \ \le \ \ln \paren{\frac{Q}{\delta \eta}}
\]
and for \certifying instances
\[
\sum_{\prlz \in \xsources}
  \epsp{\prlz}^+ \paren{\qp{\prlz} - \qpp{\prlz}} \ \le \ \ln \paren{\frac{Q}{\eta}}
\]
\end{lemma}

\begin{proof}
From Lemma~\ref{lem:deltas-potential1}, we have
\begin{equation}\label{eqn:local-progress}
\expct{\Deltap{\prlz'} \mid \prlz' \in \xsucc(\prlz) \text{ and } \prlz_e} \ \le \
  \Deltap{\prlz} \paren{1 - \epsp{\prlz}}
\end{equation}
From adaptive submodularity, we also have $\Deltap{\prlz'} \le \Deltap{\prlz}$
for all $\prlz' \in \xsucc(\prlz)$, and hence
$\expct{\Deltap{\prlz'} \mid \prlz' \in \xsucc(\prlz)  \text{ and } \prlz_e} \ \le \ \Deltap{\prlz}$.
Hence
\[
\expct{\Deltap{\prlz'} \mid \prlz' \in \xsucc(\prlz) \text{ and } \prlz_e} \ \le \
  \Deltap{\prlz} \paren{1 - \epsp{\prlz}^+}
\]
Note
\[
\sum_{\prlz \in \xsources}
  \epsp{\prlz}^+ \paren{\qp{\prlz} - \qpp{\prlz}} \le \sum_{\prlz \in \xsources}
  \epsp{\prlz}^+ \qp{\prlz}
\]
We will bound the latter sum by constructing an unbiased estimate of it, and
taking the expectation of that estimate.
Our unbiased estimator is based on a random trace $\trace(\rvrlz)$ of $\policy$
\begin{equation}
R(\rvrlz) := \sum_{\prlz \in \xsources, \prlz \in \trace(\rvrlz)} \epsp{\prlz}^+
\end{equation}
where $\rvrlz$ is sampled conditioned on $\rvrlz \sim \prlz_e$, i.e.,
each $\rlz$ is sampled with probability $\prob{\rlz \mid \prlz_e}$.

For any $\prlz \in \xsources$, the probability of seeing $\prlz$ conditioned on
$e$ being selected (i.e. $\rvrlz \sim \prlz_e$) is exactly $\qp{\prlz}$.
Hence
\begin{equation}
\expct{R(\rvrlz) \mid \rvrlz \sim \prlz_e} := \sum_{\prlz \in \xsources \cap \trace(\rvrlz)} \qp{\prlz} \epsp{\prlz}^+
\end{equation}
Next consider the random sequence of partial realizations in
$\xsources \cap \trace(\rvrlz)$, which we will denote as
$\prlz_1, \prlz_2, \ldots, \prlz_k$.
We assume without loss of generality that no policy selects an item that
will give it zero reward in expectation.
Let $Y_i(\rvrlz)$ be a random variable such that
\[
\Deltap{\prlz_{i+1}} = (1 - Y_i)\Deltap{\prlz_{i}}
\]
By adaptive submodularity, $Y_i \ge 0$ for all $i$, and also
$\expct{Y_i \mid \rvrlz \sim \prlz_e} = \epsp{\prlz_i}$ for all $i$
from \eqnref{eqn:local-progress}.
Next, note that because the maximum possible reward is $\quota$ and the
minimum increment is $\eta$, and the minimum probability of any realization is
$\delta$, we have for all $\prlz$
$\Deltap{\prlz} \le \quota$, and
$\Deltap{\prlz} > 0 \implies \Deltap{\prlz} \ge \delta \eta$.
Hence
\[
\Deltap{\prlz_{k}} = \Deltap{\prlz_{1}} \prod_{i=1}^{k-1} (1 - Y_i)
\]
implies
\[
\prod_{i=1}^{k-1} (1 - Y_i) = \Deltap{\prlz_{k}} /  \Deltap{\prlz_{1}} \ge \delta \eta / \quota
\]
Using $1-x \le e^x$, we conclude
\[
\delta \eta / \quota \le \prod_{i=1}^{k-1} (1 - Y_i) \le \exp \paren{-\sum_{i=1}^{k-1}Y_i}
\]
and hence $\sum_{i=1}^{k-1} Y_i \le \ln \paren{\frac{\quota}{\delta \eta}}$.
Hence in general,
\begin{eqnarray}
\expct{R(\rvrlz) \mid \rvrlz \sim \prlz_e} & = &
\expct{\sum_{\prlz \in \xsources, \prlz \in \trace(\rvrlz)} \epsp{\prlz}^+ \mid \rvrlz \sim \prlz_e}\\
 & = & \sum_{i \ge 1} \expct{Y_i(\rvrlz) \mid \rvrlz \sim \prlz_e}\\
 & = & \expct{\sum_{i \ge 1} Y_i(\rvrlz) \mid \rvrlz \sim \prlz_e}\\
 & \le & \ln \paren{\frac{\quota}{\delta \eta}}
\end{eqnarray}

For \certifying instances, the argument is similar, except that we have
$\Deltap{\prlz} > 0 \implies \Deltap{\prlz} \ge \eta$ because the the minimum
increment is $\eta$, and the reward is determined by the current partial
realization. Hence $\Deltap{\prlz_{k}} \ge \eta$, and so
$\sum_{i=1}^{k-1} Y_i \le \ln \paren{\frac{\quota}{\eta}}$ in this case, so
$\expct{R(\rvrlz) \mid \rvrlz \sim \prlz_e} \le \ln \paren{\frac{\quota}{\eta}}$.
\end{proof}

\subsubsection*{Putting it all together}

We show how to obtain the result for general instances. The result for
\certifying instances
is strictly analogous.

From \eqnref{eqn:reward-from-opt} and \eqnref{eqn:reward-from-opt2},
we have that for all $(\prlz, \prlz') \in \xarcs$
\begin{equation} \label{eqn:joint-progress-bound}
\diff{\prlz; \prlz'}{\piavg} \ge \max \paren{Q - x, \delta \eta}.
\end{equation}
From \eqnref{eqn:bound-fix1} and the fact that $\policy$ is an
$\alpha$-approximate greedy policy by assumption we have that
for all $(\prlz, \prlz') \in \xarcs$
\begin{equation}
\frac{c(\policy(\prlz))}{\diff{\prlz}{\policy(\prlz)}} \le \frac{\alpha \acondcost{\prlz; \prlz'}{\piavg} }{ \diff{\prlz; \prlz'}{\piavg} }.
\end{equation}
Combining this with \eqnref{eqn:joint-progress-bound} yields
\begin{equation}
\frac{c(\policy(\prlz))}{\diff{\prlz}{\policy(\prlz)}} \le \frac{\alpha \acondcost{\prlz; \prlz'}{\piavg} }{ \max \paren{Q - x, \delta \eta}}
\end{equation}
for all $(\prlz, \prlz') \in \xarcs$.

The expected price $\policy$ pays per unit of expected reward at the instant it
reaches $x$ expected reward is
\begin{eqnarray}
\price(x) & := & \sum_{(\prlz, \prlz') \in \xarcs} \prob{\prlz'} \frac{c(\policy(\prlz))}{\diff{\prlz}{\policy(\prlz)}} \\
 & \le & \sum_{(\prlz, \prlz') \in \xarcs} \prob{\prlz'} \frac{\alpha \acondcost{\prlz; \prlz'}{\piavg} }{ \max \paren{Q - x, \delta \eta}} \\
 & = & \frac{\alpha \acavg{\piavg}}{ \max \paren{Q - x, \delta \eta}}
\end{eqnarray}
since $\set{\prlz' : \exists \prlz, (\prlz, \prlz') \in \xarcs}$ partitions
the set of realizations.

Next, we can apply Lemma~\ref{lem:price-integral}
to bound $\acst{\policy}$ as follows.

\begin{eqnarray}
\acst{\policy} & = & \int_{x=0}^{Q} \price(x) dx\\
 & \le & \int_{x=0}^{Q} \frac{\alpha \acavg{\piavg}}{ \max \paren{Q - x, \delta \eta}} dx \\
 & = & \alpha \acavg{\piavg} \int_{x=0}^{Q} \frac{1}{ \max \paren{Q - x, \delta \eta}} dx \\
 & \le & \alpha \cavg{\piavg} \paren{\ln \paren{\frac{Q}{\delta \eta}} + 1} \int_{x=0}^{Q} \frac{1}{ \max \paren{Q - x, \delta \eta}} dx\\
 & \le & \alpha \cavg{\piavg} \paren{\ln \paren{\frac{Q}{\delta \eta}} + 1}^2
\end{eqnarray}


\noindent
This completes the proof for general instances (as mentioned, the result for
\certifying instances
is strictly analogous, merely with different bounds used to relate $\acavg{\piavg}$ to $\cavg{\piavg}$),
aside from
proving~\lemref{lem:price-integral}.
The elegant proof that we present here is
due to \citet{feldman12lemma}.

\begin{lemma} \label{lem:price-integral}
Suppose $f:2^{\groundset} \times \outcomes^{\groundset} \to
\NonNegativeReals$ is \term submodular and strongly adaptive
monotone with respect to $\rlzprior$ and there exists $Q$ such that
$f(\groundset, \rlz) = Q$ for all $\rlz$.
Fix any policy $\policy$ that covers every realization, and let
$\price(x)$ be defined as in the proof
of~\thmref{thm:min-set-cover-avg-generalized-with-costs}.
Then
\[
\acst{\policy} \ = \ \int_{x=0}^{Q} \price(x) dx\mbox{.}
\]
\end{lemma}

\begin{proof}
Recall
\[
\price(x) := \sum_{(\prlz, \prlz') \in \xarcs} \prob{\prlz'} \frac{c(\policy(\prlz))}{\diff{\prlz}{\policy(\prlz)}}
\]
We use an alternative charging scheme to compute the expected cost of
$\policy$: When the policy observes $\prlz$ and selects $e =
\policy(\prlz)$, we charge it based on the outcome $\outcome
= \rlz(e)$.  Specifically, we charge it proportionally to its gain in
expected reward, as measured by $\prlz \mapsto \progress{\prlz}$.  Formally, upon observing $\prlz' := \prlz \cup
\set{(e, \outcome)}$ after selecting $e =
\policy(\prlz)$, the algorithm is charged
\begin{equation}
  \label{eq:charge-def}
  \charge{e}(\prlz, \prlz') := c(e) \paren{\frac{\progress{\prlz'} - \progress{\prlz}}{\diff{\prlz}{e}}}.
\end{equation}
It is straightforward to show that
$\expct{\charge{e}(\prlz, \prlz \cup \set{(e, \rvrlz(e))}) \ \mid \ \rvrlz \sim \prlz} = c(e)$,
using the basic fact that if event $A$ is the union of
finitely many disjoint events $B_1, \ldots, B_m$ and $X$ is any random
variable, then $\sum_i \prob{B_i \mid A} \expct{X \mid B_i} = \expct{X
\mid A}$.  Hence, in expectation this charging scheme charges the
policy exactly $\acst{\policy}$.  Moreover, the rate at which it charges
$\policy$ upon observing $\prlz$ and selecting $e = \policy(\prlz)$ is
exactly $c(e) / \diff{\prlz}{e}$ per unit expected gain.
Let $\price(\prlz) := c(\policy(\prlz)) / \diff{\prlz}{\policy(\prlz)}$.
Under this
charging scheme the policy pays
$\price(\prlz)dx$
to obtain an additional $dx$ expected reward, given it already has $x$, the true realization is $\rlz$, and $(\prlz, \prlz') \in \xarcs$ for some $\prlz'$ with
$\rlz \sim \prlz'$.
Let $\ptrace{\rlz}{x}$ be the partial realization in the execution trace of $\policy$ under $\rlz$ immediately before it obtains $x$ reward in expectation.
Formally $\ptrace{\rlz}{x}$ is the maximal element of $\dom(\policy)$
such that $\rlz \sim \prlz$ and $f(\prlz) < x$, where $f(\prlz)$ is defined as in
\eqnref{eqn:short-expect-value}.

The total alternative cost of $\policy$ under $\rlz$ is
then $\int_{x=0}^{Q} \price(\ptrace{\rlz}{x}) dx$, and we can compute
the expected cost of $\policy$ as
$$ \acst{\policy} = \expct{\int_{x=0}^{Q} \price(\ptrace{\rvrlz}{x})
  dx}. $$
where the expectation is taken with respect to $\rvrlz$.
We can then obtain the claimed equality by exchanging the integral and the expectation operators:
$$\acst{\policy} \ = \  \int_{x=0}^{Q}
\expct{\price(\ptrace{\rvrlz}{x})} dx \ = \ \int_{x=0}^{Q} \price(x)
dx.$$
We can justify this operator exchange in several ways, for example we can apply Tonelli's theorem since the prices $\price$ are always non-negative.
\end{proof}

\noindent
Next we consider the worst-case cost.
We generalize~\thmref{thm:min-set-cover-wc-generalized} by
incorporating arbitrary item costs.

\begin{theorem} \label{thm:min-set-cover-wc-generalized-with-costs}
Suppose $f:2^{\groundset} \times \outcomes^{\groundset} \to
\NonNegativeReals$
is \term monotone and \term submodular
with respect to $\rlzprior$, and
let $\eta$ be any value such that
$f(S, \rlz) > f(\groundset, \rlz) - \eta$ implies $f(S, \rlz) =
f(\groundset, \rlz)$ for all $S$ and $\rlz$.
Let $\delta = \min_{\rlz} \rlzprior$ be the minimum probability of
any realization.
Let $\policy_{wc}^{*}$ be the optimal policy
minimizing the worst-case cost $\wcst{\cdot}$
while guaranteeing that every realization is covered.  Let $\policy$ be an $\alpha$-approximate
greedy policy with respect to the item costs.
Finally, let $Q := \expctoverrlz{\rvrlz}{f(\groundset, \rvrlz)}$ be the
maximum possible expected reward.
Then
\[
 \wcst{\policy} \le \alpha \, \wcst{\policy_{wc}^*}\,
 \paren{\ln \paren{\frac{Q}{\delta \eta}} +1}
\mbox{.}
\]
\end{theorem}

\begin{proof}
Let $\policy$ be an $\alpha$-approximate greedy policy.
Let $k = \wcst{\policy_{wc}^*}$, let $\ell = \alpha k \ln \paren{Q/
  \delta \eta}$, and apply \thmref{thm:max-cover-with-costs} with
 these parameters to yield
 \begin{equation}
   \label{eq:wc1costs}
\avgf(\prune{\policy}{\ell}) \ > \ \paren{1 - e^{-\ell/\alpha k}}
\avgf(\policy_{wc}^*) \ = \  \paren{1 - \frac{\delta \eta}{Q}}\avgf(\policy_{wc}^*)
\mbox{.}
 \end{equation}
 Since $\policy_{wc}^*$ covers every realization by assumption, $\avgf(\policy_{wc}^*) =
 \expctoverrlz{\rvrlz}{f(\groundset, \rvrlz)} = Q$, so rearranging terms
 of~\eqnref{eq:wc1costs} yields $Q - \avgf(\prune{\policy}{\ell}) < \delta \eta$.
 Since $\avgf(\prune{\policy}{\ell}) \le
 \avgf(\laxprune{\policy}{\ell})$ by the \term monotonicity of $f$,
it
 follows that $Q - \avgf(\laxprune{\policy}{\ell}) < \delta \eta$.
By definition of $\delta$ and $\eta$, if some $\rlz$ is
not covered by $\laxprune{\policy}{\ell}$ then
$Q - \avgf(\laxprune{\policy}{\ell}) \ge \delta \eta$.
Thus $Q - \avgf(\laxprune{\policy}{\ell}) < \delta \eta$ implies $Q -
 \avgf(\laxprune{\policy}{\ell}) = 0$, meaning $\laxprune{\policy}{\ell}$ covers
 every realization. %

 We next claim that $\laxprune{\policy}{\ell}$ has worst-case cost at most $\ell +
 \alpha k$.  It is sufficient to show that the final item executed by
 $\laxprune{\policy}{\ell}$ has cost at most $\alpha k$ for any
 realization.
  As we will prove, this follows from the facts that $\policy$ is an
 $\alpha$-approximate greedy policy and $\policy_{wc}^*$ covers
 every realization at cost at most $k$.
The data dependent bound, \lemref{lem:rate-equation-with-costs} on page~\pageref{lem:rate-equation-with-costs},
 guarantees that
 \begin{equation}
   \label{eq:1app-wc}
   \max_e \paren{\frac{\diff{\prlz}{e}}{c(e)}} \ge
 \frac{ \diff{\prlz}{ \policy_{wc}^* } } {\condcost{\prlz}{\policy_{wc}^*}}
\ge \frac{ \diff{\prlz}{ \policy_{wc}^* }}{k}.
 \end{equation}
Suppose $\prlz \in \dom(\policy)$.
We would like to say that $\max_e  \diff{\prlz}{e} \le \diff{\prlz}{ \policy_{wc}^* }$.
Supposing this is true, any item $e$ with cost $c(e) > \alpha k$
must have $\diff{\prlz}{e} / c(e) < \diff{\prlz}{ \policy_{wc}^* } / \alpha k $, and
hence cannot be selected by any $\alpha$-approximate greedy policy
upon observing $\prlz$ by~\eqnref{eq:1app-wc}, and thus the
 final item executed by $\laxprune{\policy}{\ell}$ has cost at most $\alpha k$ for any
 realization.  So we next show that $\max_e  \diff{\prlz}{e} \le \diff{\prlz}{ \policy_{wc}^* }$.
Towards this end, note that \lemref{lem:pick-all-bound} implies
\begin{equation}
  \label{eq:2b-app-wc}
\max_e  \diff{\prlz}{e} \le \expct{f(\groundset,\rvrlz)   \mid  \rvrlz
  \sim \prlz} - \expct{f(\dom(\prlz),\rvrlz)   \mid  \rvrlz
  \sim \prlz}.
\end{equation}
and to prove $\max_e  \diff{\prlz}{e} \le \diff{\prlz}{ \policy_{wc}^* }$ it suffices to
show
\begin{equation}
  \label{eq:2c-app-wc}
\expct{f(\groundset,\rvrlz)   \mid  \rvrlz
  \sim \prlz} \le  \expct{f(\played{\policy_{wc}^*}{\rvrlz} \cup \dom(\prlz),\rvrlz)   \mid  \rvrlz
  \sim \prlz}.
\end{equation}

Proving \eqnref{eq:2c-app-wc} is quite straightforward if $f$ is \emph{strongly} \term
monotone.  Given that $f$ is only \term
monotone, it requires some additional effort.
So fix $A \subset \groundset$ and let $\policy_{A}$ be a non-adaptive policy that selects all
items in $A$ in some arbitrary order.
Let $\prlzset := \set{\prlz : \dom(\prlz) = A}$.
Apply~\lemref{lem:pick-all-bound} with $\policy' =
\append{\policy_{A}}{\policy_{wc}^*}$ and any $\prlz \in \prlzset$
to obtain
\begin{equation}
  \label{eq:3app-wc}
  \expct{f(\played{\policy_{wc}^*}{\rvrlz} \cup
  A, \rvrlz)  \mid  \rvrlz \sim \prlz} \le \expct{f(\groundset,
  \rvrlz)   \mid  \rvrlz \sim \prlz}.
\end{equation}
Note that $\sum_{\prlz \in \prlzset} \Pr{\rvrlz \sim \prlz} \cdot \expct{f(\played{\policy_{wc}^*}{\rvrlz} \cup
  A, \rvrlz)  \mid  \rvrlz \sim \prlz} =
\avgf(\append{\policy_{A}}{\policy_{wc}^*}) \ge \avgf(\policy_{wc}^*)
= \expct{f(\groundset,\rvrlz)  }$.
Since we know
$\expct{f(\groundset,\rvrlz)} = \sum_{\prlz \in \prlzset} \Pr{\rvrlz \sim \prlz} \cdot \expct{f(\groundset,\rvrlz)   \mid  \rvrlz \sim \prlz}$,
an averaging argument together with \eqnref{eq:3app-wc}
then implies that for all $\prlz \in \prlzset$
\begin{equation}
  \label{eq:4app-wc}
  \expct{f(\played{\policy_{wc}^*}{\rvrlz} \cup
 A, \rvrlz)  \mid  \rvrlz \sim \prlz} = \expct{f(\groundset,
  \rvrlz)  \mid \rvrlz \sim \prlz}.
\end{equation}

Since $\prlz$ was an arbitrary partial realization with $\dom(\prlz) =
A$, and $A \subseteq \groundset$ was arbitrary,
fix $\prlz \in \dom(\policy)$ and let $A = \dom(\prlz)$.
With these settings, \eqnref{eq:4app-wc} implies
\eqnref{eq:2c-app-wc}, and thus $\max_e \diff{\prlz}{e} \le
\diff{\prlz}{ \policy_{wc}^* }$, and thus an $\alpha$-approximate greedy policy
can never select an item with cost exceeding $\alpha k$, where
$k = \wcst{\policy_{wc}^*}$.  Hence
$\wcst{\laxprune{\policy}{\ell}} - \wcst{\prune{\policy}{\ell}} \le
\alpha k$, and so $\wcst{\laxprune{\policy}{\ell}} \le \ell + \alpha k$.
This completes the proof.
\end{proof}

\begin{lemma} \label{lem:pick-all-bound}
Fix \term monotone submodular objective $f$.
For any policy $\policy$ and any $\prlz \in \dom(\policy)$ we have
$$\expct{f(\played{\policy}{\rvrlz}, \rvrlz)\  \mid \ \rvrlz \sim \prlz} \le \expct{f(\groundset,
  \rvrlz) \  \mid \ \rvrlz \sim \prlz}.$$
\end{lemma}

\begin{proof}
Augment $\policy$ to a new policy $\policy'$ as follows.
Run $\policy$ to completion, and let $\prlz'$ be the partial
realization consisting of all of the states it has observed.  If $\prlz \subseteq \prlz'$,
then proceed to select all the remaining items in $\groundset$ in any
order.  Otherwise, if $\prlz \nsubseteq \prlz'$ then terminate.
Then %
\begin{equation}
  \label{eq:2app-wc}
  \expct{f(\played{\policy}{\rvrlz}, \rvrlz)\  \mid \ \rvrlz \sim \prlz} \le
\expct{f(\played{\policy'}{\rvrlz}, \rvrlz)\  \mid \ \rvrlz \sim \prlz}  =
\expct{f(\groundset,
  \rvrlz) \  \mid \ \rvrlz \sim \prlz}
\end{equation}
where the inequality is by repeated application of the \term monotonicity of $f$, and the equality
is by construction.
\end{proof}

In \secref{sec:min-cost-cover} we described how the
result of
\citet{feige98threshold} implies that there is
no polynomial time
\mbox{$(1-\epsilon)\ln \paren{Q/\eta}$} approximation algorithm for
\certifying instances of
Adaptive Stochastic Min Cost Cover, unless $\NP\subseteq
\text{DTIME}(n^{\cO(\log\log n)})$.  Here we show a related result
for general instances.

\begin{lemma} \label{lem:approx-hardness-for-c_avg}
For every constant $\epsilon > 0$, there is no $(1-\epsilon)\ln \paren{Q/\delta\eta}$ polynomial time approximation
algorithm for general instances of
Adaptive Stochastic Min Cost Cover, for either the average case
objective $\acst{\cdot}$ or the worst-case objective $\wcst{\cdot}$,
 unless $\NP\subseteq
\text{DTIME}(n^{\cO(\log\log n)})$.
\end{lemma}

\begin{proof}
We offer a reduction from the Set Cover problem.  Fix a Set Cover
instance with ground set $U$ and sets $\set{S_1, S_2, \ldots, S_m} \subseteq 2^{U}$ with
unit-cost sets.
Fix $Q, \eta$ and $\delta$ such that $1/\delta$ and $Q/\eta$ are
positive integers, and $\frac{Q}{\delta \eta} = |U|$.
Let $\groundset := \set{S_1, S_2, \ldots, S_m}$, and set the cost of
each item to one.
Partition $U$ into $1/\delta$ disjoint, equally sized subsets
$U_1, U_2, \ldots, U_{1/\delta}$.  Construct a realization
$\rlz_i$ for each $U_i$.  Let the set of states be $\outcomes = \set{\NULL}$.
Hence $\rlz_i(e) = \NULL$ for all $i$ and $e$, so that
no knowledge of the true realization is revealed by selecting items.
We use a uniform distribution over
realizations, i.e., $\rlzmass{\rlz_i} = \delta$ for all $i$.
Finally, our objective is
$f(\collection, \rlz_i) := | \cup_{S \in \collection}  (S \cap U_i )
|$, i.e., the number of elements in $U_i$ that we cover with sets in
$\collection$.
Since $|\outcomes| = 1$, every realization is consistent with every
possible partial realization $\prlz$.  Hence for any $\prlz$, we have
$\progress{\prlz} = \delta \hat{f}(\dom(\prlz))$, where
$\hat{f}( \collection ) = | \cup_{S \in \collection} S |$ is the
objective function of the original set cover instance.
Since $\hat{f}$ is submodular, $f$ is \term submodular.
Likewise, since $\hat{f}$ is monotone, and $|\outcomes| = 1$,
$f$ is strongly \term monotone.
Now, to cover any realization, we must obtain the maximum possible
value for all realizations, which means selecting a collection of sets
$\collection$ such that $\cup_{S \in \collection} S = U$.
Conversely, any $\collection$ such that $\cup_{S \in \collection} S =
U$ clearly covers $f$.  Hence this instance of Adaptive Stochastic Min
Cost Cover,
with either the average case
objective $\acst{\cdot}$ or the worst-case objective $\wcst{\cdot}$,
is equivalent to the original Set Cover instance.
Therefore, the result
from \citet{feige98threshold} implies that there is no polynomial
time algorithm for obtaining a $(1-\epsilon)\ln |U| =
(1-\epsilon)\ln \paren{Q/\delta\eta}$ approximation for Adaptive Stochastic Min
Cost Cover unless
$\NP\subseteq
\text{DTIME}(n^{\cO(\log\log n)})$.
\end{proof}

\subsection{The Min-Sum Objective} \label{sec:proofs-min-sum-cover}
\noindent In this section we prove
Theorem~\ref{thm:min-sum-set-cover}, which appears on page~\pageref{thm:min-sum-set-cover},
in the case where the items have arbitrary costs.
Our proof
resembles the analogous proof of~\citet{streeter07tr} for the non-adaptive min-sum
submodular cover problem, and, like that
proof, ultimately derives from an extremely elegant performance analysis
of the greedy algorithm for min-sum set cover due to~\citet{feige04}.\\

\noindent
The objective function $\costminsum{\cdot}$ generalized to
arbitrary cost items uses the strict truncation\footnote{See
  \defref{def:strict-prune} on page~\pageref{def:strict-prune}.}
$\strictprune{\policy}{t}$ in place of $\prune{\policy}{t}$ in the unit-cost definition:
\begin{equation}
  \label{eqn:min-sum-obj-with-costs}
  \costminsum{\policy} := \sum_{t =  0}^{\infty} \paren{\expctoverrlz{\rvrlz}{f(\groundset, \rvrlz)} -
      \avgf(\strictprune{\policy}{t})} = \sum_{\rlz} \rlzprior \sum_{t =
    0}^{\infty} \paren{f(\groundset, \rlz) - f(\groundset(\strictprune{\policy}{t},
  \rlz), \rlz)}.
\end{equation}

\noindent
We will prove that any $\alpha$-approximate greedy policy $\policy$
achieves a $4 \alpha$-approximation for the min-sum objective, i.e.,
$\costminsum{\policy} \le 4 \alpha \, \costminsum{\policy^*}$ for all policies
$\policy^*$.
To do so, we require the following lemma.

\begin {lemma} \label{lem:greedy_rate}
Fix an $\alpha$-approximate greedy policy $\policy$ for some \term monotone submodular function $f$
and let $s_i :=\alpha \paren{\avgf(\prune{\policy}{i+1}) -  \avgf(\prune{\policy} {i})}$.
For any policy $\policy^*$ and nonnegative integers $i$ and $k$, we have $\avgf(\prune {\policy^*} {k}) \le  \avgf(\strictprune {\policy} {i}) + k \cdot s_i $.
\end {lemma}

\begin{proof}
Fix $\policy, \policy^*, i$, and $k$.  By \term monotonicity
$\avgf(\prune {\policy^*} {k})  \le  \avgf(\append{\strictprune{\policy}{i}}{\prune {\policy^*} {k}})$.
We next aim to prove
\begin{equation}
  \label{eq:1mssc}
  \avgf(\append{\strictprune{\policy}{i}}{\prune {\policy^*} {k}}) \le  \avgf(\strictprune {\policy} {i}) + k \cdot s_i
\end{equation}
which is sufficient to complete the proof.
Towards this end, fix a partial realization $\prlz$ of the form $\set{(e, \rlz(e)) : e \in
  \played{\strictprune{\policy}{i}  }{\rlz}}$ for some $\rlz$.
Consider $\diff{\prlz}{\prune{\policy^*}{k}}$, which equals
the expected marginal benefit of the $\prune{\policy^*}{k}$ portion of
$\append{ \strictprune{\policy}{i} }{\prune{\policy^*}{k}}$
conditioned on $\rvrlz \sim
\prlz$.
\lemref{lem:rate-equation-with-costs} allows us to bound it as
$$\expct{\diff{\prlz}{\prune{\policy^*}{k}}} \ \le \ \expct{\condcost{\prlz}{\prune{\policy^*}{k}}} \cdot
\max_{e} \paren{\frac{\diff{\prlz}{e}}{c(e)}},$$
where the expectations are taken
over the internal randomness of $\policy^*$, if there is any.  Note
that
for all $\rlz$, we have
$\expct{c(\played{\prune{\policy^*}{k}}{\rlz})} \le k$, where the
expectation is again taken over the internal randomness of $\prune{\policy^*}{k}$.
Hence $\expct{\condcost{\prlz}{\prune{\policy^*}{k}}} \le k$ for all $\prlz$.  It
follows that $\expct{\diff{\prlz}{\prune{\policy^*}{k}}} \le k \cdot
\max_{e} \paren{\diff{\prlz}{e}/c(e)}$.
By definition of an $\alpha$-approximate greedy policy, $\policy$
obtains at least
\begin{equation}
  \label{eq:mssc2a}
 (1/\alpha)\max_{e} \paren{\diff{\prlz}{e} / c(e) }
\ge \expct{\diff{\prlz}{\prune{\policy^*}{k}}} / \alpha k
\end{equation}
expected marginal benefit per
unit cost in a step immediately following its observation of $\prlz$.
Next we take an appropriate convex combination of the previous inequality
with different values of $\prlz$.
Let $\rvprlz$ be a random partial realization distributed as
$\set{(e, \rvrlz(e)) : e \in  \played{\strictprune{\policy}{i}  }{\rvrlz}}$.
Then taking the expectation of~\eqnref{eq:mssc2a} over $\rvprlz$ yields
\begin{equation}
  \label{eq:decondition2}
  \avgf(\prune{\policy}{i+1}) -
  \avgf(\prune{\policy}{i}) \ge
\expctover{\prlz}{ \frac{\expct{\diff{\rvprlz}{\prune{\policy^*}{k}}}}{\alpha k} } = \frac{ \avgf(
    \append{\strictprune{\policy}{i}}{\prune{\policy^*}{k}})  -
    \avgf(\strictprune{\policy}{i})}{\alpha k} .
\end{equation}
Multiplying \eqnref{eq:decondition2}
by $\alpha k$, and
substituting in $s_i = \alpha \paren{\avgf(\prune{\policy}{i+1}) -
  \avgf(\prune{\policy} {i})}$, we conclude
$ks_i \ge \avgf(\append{\strictprune{\policy}{i}}{\prune{\policy^*}{k}})  -
    \avgf(\strictprune{\policy}{i})$
which immediately yields~\eqnref{eq:1mssc} and concludes the
proof.
\end{proof}

\noindent
Using Lemma \ref{lem:greedy_rate}, together with a geometric argument
developed by \citet{feige04}, we now prove
Theorem~\ref{thm:min-sum-set-cover}. \\

\begin {figure} [t!]
	\begin {center}
		\includegraphics [width=3in]{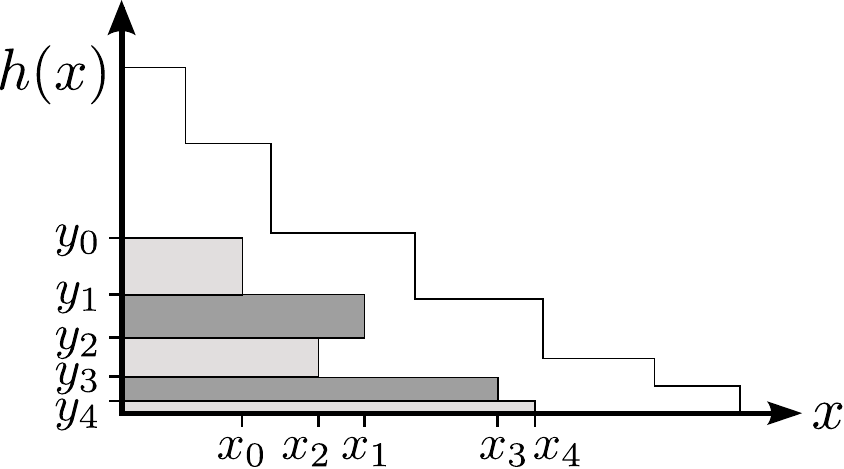} %
	\end {center}
	\caption {An illustration of the inequality
          $\int_{x=0}^{\infty} h(x) dx \ge \sum_{i \ge 0} x_i \paren { y_i - y_{i+1} }$.}
	\label {fig:proof_by_picture}
\end {figure}

\begin {proofof}{Theorem~\ref{thm:min-sum-set-cover}}
Let $\quota := \expctoverrlz{\rvrlz}{f(\groundset, \rvrlz)}$ be the maximum
possible expected reward, where the expectation is taken w.r.t. $\rlzprior$.
Let $\policy$ be an $\alpha$-approximate greedy policy.
Define $R_i := \quota - \avgf \paren  {\prune{\policy}{i}}$ and define $P_i := \quota - \avgf \paren  {\strictprune{\policy}{i}}$.
Let $x_i := \frac {P_i} {2 s_i}$, let $y_i := \frac {R_i} {2}$, and
let $h(x) := \quota - \avgf(\prune{\policy^*} {x})$.
We claim $\avgf \paren
{\strictprune{\policy}{i}} \le \avgf \paren  {\prune{\policy}{i}}$ and so
$P_i \ge R_i$.
This clearly holds if $\strictprune{\policy}{i}$ is the empty policy, and
otherwise $\policy$ can always select an item
that contributes zero marginal benefit, namely an item
it has already played previously.  Hence an $\alpha$-approximate greedy policy
$\policy$ can never select items with negative expected marginal
benefit, and so
$\avgf \paren {\strictprune{\policy}{i}} \le \avgf \paren  {\prune{\policy}{i}}$.
By Lemma~\ref {lem:greedy_rate}, $ \avgf \paren{\prune{\policy^*}
  {x_i} } \le
\avgf \paren{\strictprune{\policy} {i} } + x_i s_i$.
Therefore
\begin{equation}
  \label{eq:min-sum1}
h(x_i) \ge \quota - \avgf(\strictprune{\policy} {i}) - x_i \cdot s_i = P_i  -
\frac {P_i} {2} \ge \frac {R_i} {2} = y_i
\end{equation}

For similar reasons that $\avgf \paren {\strictprune{\policy}{i}} \le
\avgf \paren  {\prune{\policy}{i}}$, we have
$\avgf \paren {\prune{\policy}{i-1}} \le \avgf \paren  {\prune{\policy}{i}}$, and
so the sequence $\tuple {y_1, y_2, \ldots}$  is non-increasing.
The \term monotonicity and \term submodularity of $f$ imply that $h(x)$ is
non-increasing.  Informally, this is because otherwise, if
$\avgf(\prune{\policy^*} {x}) > \avgf(\prune{\policy^*} {x+1})$ for some
$x$, then the optimal policy must be sacrificing immediate rewards at
time $x$ in exchange for greater returns later, and it can be shown
that if such a strategy is optimal, then \term submodularity cannot hold.
\eqnref{eq:min-sum1} and the monotonicity of $h$ and $i \mapsto y_i$
imply that $\int_{x=0}^{\infty} h(x)dx
\ge \sum_{i \ge 0} x_i \paren { y_i - y_{i+1} }$ (see Figure \ref
{fig:proof_by_picture}).  The left hand side is a lower bound for
$\costminsum{\policy^*}$, and because $s_i = \alpha \paren {R_i -
  R_{i+1}}$ the right hand side simplifies to
$\frac{1}{4\alpha} \sum_{i \ge 0} P_i = \frac{1}{4\alpha}
\costminsum{\policy}$, proving $\costminsum{\policy} \le 4 \alpha \cdot \costminsum{\policy^*}$.
\end {proofof}

\newcommand{\gimmespace}{\phantom{$\paren{\frac{\sum }{2}}$}}

\begin{table}[p]
\subsection{A Symbol Table} \label{sec:symbols} \vspace{1em}
  \begin{tabular}{|l|p{5in}|}
\hline
$\groundset$, $\elem \in \groundset$ & Ground set of items, and an individual item. \\
\hline
$\outcomes$, $\outcome \in \outcomes$ & States an item may be in, or outcomes of selecting an item, and an individual state/outcome. \\
\hline
$\rlz$ & A realization, i.e., a function from items to states. \\
\hline
$\prlz$ & A partial realization, typically encoding the current set of observations;\\
 &  each $\prlz \subset \groundset \times \outcomes$ is a partial
 mapping from items to states. \\
\hline
$\rvrlz, \rvprlz$ & A random realization and a random partial
realization, respectively.\\
\hline
$\sim$ & The consistency relation: $\rlz \sim \prlz$ means $\prlz(e) = \rlz(e)$ for all $e \in \dom(\prlz)$.\\
\hline
$\rlzmasssym$ & The probability distribution on realizations.\\
\hline
$\rlzmasssym(\rlz \mid \prlz)$ & The conditional distribution on realizations: $\rlzmasssym(\rlz \mid \prlz) :=
\prob{\rvrlz = \rlz \mid \rvrlz \sim \prlz}$.\\
\hline
$\policy$ & A policy, which maps partial realizations to items. \\
\hline
$\played{\policy}{\rlz}$ & The set of all items selected by $\policy$ when run
under realization $\rlz$. \\
\hline
$\diff{\prlz}{\elem}$ & The conditional expected marginal benefit of $e$ conditioned on $\prlz$: \\
                              & $\diff{\prlz}{\elem} := \expct{f(\dom(\prlz) \cup \set{\elem}, \rvrlz) -
   f(\dom(\prlz), \rvrlz)\ | \ \rvrlz\sim\prlz}$. \\
\hline
$\diff{\prlz}{\policy}$ & The conditional expected marginal benefit of
policy $\policy$ conditioned on $\prlz$: \\
                              & $\diff{\prlz}{\policy} :=
                              \expct{f(\dom(\prlz) \cup \played{\policy}{\rvrlz}, \rvrlz) -
   f(\dom(\prlz), \rvrlz)\ | \ \rvrlz\sim\prlz}$. \\
\hline
$\substitute{\prlz}{\elem}{\outcome}$ & Shorthand for $\prlz \cup \set{(\elem, \outcome)}$.\\
\hline
$\budget$ & Budget on the cost of selected item sets.\\
\hline
$\prune{\policy}{k}$ & A truncated policy.  See Definition~\vref{def:policy-truncation} (unit costs) and Definition~\vref{def:policy-truncation-with-costs}. \\
\hline
$\strictprune{\policy}{k}$ & A strictly truncated policy.  See Definition~\vref{def:strict-prune}.\\
\hline
$\laxprune{\policy}{k}$ & A laxly truncated policy.  See Definition~\vref{def:lax-prune}.\\
\hline
$\append{\policy}{\policy'}$ & Policies $\policy$ and $\policy'$ concatenated together.  See Definition~\vref{def:policy-concatenation}.\\
\hline
$f$ & An objective function, of type $f:2^{\groundset} \times \outcomes^{\groundset} \to \NonNegativeReals$ unless stated otherwise.\\
\hline
$\avgf$ & Average benefit: $\avgf(\policy) := \expct{f(\played{\policy}{\rvrlz}, \rvrlz)}$.\\
\hline
$c$ & Item costs $c:\groundset \to \nats$.  Extended to sets via $c(S) := \sum_{e \in S} c(e)$. \\
\hline
$\acstsym$ &  Average cost of a policy: $\acst{\policy} := \expct{c(\played{\policy}{\rvrlz})}$.\\
\hline
$\awstsym$ &  Worst-case cost of a policy: $\wcst{\policy} := \max_{\rlz}{c(\played{\policy}{\rlz})}$.\\
\hline
$\costminsumsym$ &  Min-sum cost of a policy: $\costminsum{\policy} := \sum_{t =  0}^{\infty} \paren{\expct{f(\groundset, \rvrlz)} -
      \avgf(\strictprune{\policy}{t})}.$ \\
\hline
$\condcost{\prlz}{\policy}$ & Conditional average policy cost:
$\condcost{\prlz}{\policy} :=
\expct{c(\played{\policy}{\rvrlz})  \mid  \rvrlz \sim  \prlz}$. \ignore{\gimmespace} \\
\hline
$\alpha$ & Approximation factor for greedy optimization in an $\alpha$-approximate greedy policy.\\
\hline
$\quota$ & Benefit quota.  Often $\quota = \expct{f(\groundset, \rvrlz)}$.\\
\hline
$\eta$ & Coverage gap: $\eta = \sup \set{ \eta'  :  f(S, \rlz) > Q
  - \eta' \text{ implies }f(S, \rlz) \ge Q \text{ for all }S, \rlz }$.\\
\hline
$\indicator{P}$ & The indicator for proposition $P$, which equals one if $P$ is true and
zero if $P$ is false.\\
\hline    
\end{tabular}
 \caption{Important symbols and notations used in this article}  \label{table:symbol-table}
\end{table}

\subsection{Proof of Approximation Hardness in the Absence of \Term
  Submodularity}  \label{sec:proof-approx-hardness}

We now provide the proof of \thmref{thm:hardness} whose statement
appears on page~\pageref{thm:hardness} in~\secref{sec:hardness}.\\

\begin{proofof}{\thmref{thm:hardness}}
We construct a hard instance based on the following
intuition.  We make the algorithm go ``treasure hunting''.  There is a
set of $\ntreasure$ locations $\set{0, 1,, \ldots, \ntreasure-1}$, there is a treasure
at one of these locations, and the algorithm gets unit reward if it
finds it, and zero reward otherwise.  There are $\nmaps$ ``maps,''
each consisting of a cluster of $\mapsize$ bits, and each
purporting to indicate where the treasure is, and each map is stored
in a (weak) secret-sharing way, so that querying few bits of a map
reveals nothing about where it says the treasure is.
Moreover, all but one of the maps are \emph{fake}, and there is a
puzzle indicating which map is the correct one
indicating the treasure's location.  Formally, a fake map is one
which is probabilistically independent of the location of the treasure,
conditioned on the puzzle.

Our instance will have three types of items, $\groundset =
  \groundset_T \uplus \groundset_M \uplus \groundset_{P}$, where
  $|\groundset_{T}|=\ntreasure$ encodes where the treasure is,
  $|\groundset_{M}|=\nmaps \mapsize$ encodes the maps, and
  $|\groundset_{P}|=\nmatrix^{3}$ encodes the puzzle, where
 $\nmaps,\ntreasure,\mapsize$ and $\nmatrix$ are specified below.
  All outcomes are binary, so that $\outcomes=\{0,1\}$, and we
  identify items with bit indices.  Accordingly we say that
  $\rlz(\elem)$ is the value of bit $\elem$.
  For all $e\in
  \groundset_{M}\cup\groundset_{P}$, $\prob{\rvrlz(e)=1}=.5$
  independently. The conditional distribution of
  $\rvrlz(\groundset_{T})$ given $\rvrlz(\groundset_{M}\cup\groundset_{P})$
  will be deterministic as specified below. Our objective function $f$
  is linear, and defined as follows:
$$f(\groundsubset,\rlz)=|\{e\in \groundsubset\cap\groundset_{T}:\rlz(e)=1\}|.$$

\noindent We now describe the puzzle, which is to compute
$i(P) := (\perm(P) \mod p) \mod 2^\ell$ for a suitable random matrix $P$,
and suitable prime $p$ and integer $\ell$, where
$\perm(P) = \sum_{\sigma \in S_{\nmatrix}} \prod_{i=1}^n P_{i\sigma(i)}$ is the permanent of $P$.
We exploit Theorem~$1.9$
of~\citet{FeigeL97} in which they show that if there exist constants
$\eta, \delta > 0$ such that a randomized polynomial time algorithm can
compute $(\perm(P) \mod p) \mod 2^\ell$ correctly with probability
$2^{-\ell}(1 + 1/\nmatrix^\eta)$, where $P$ is drawn uniformly at random from
$\set{0,1,2, \ldots, p-1}^{\nmatrix \times \nmatrix}$, $p$ is any prime
superpolynomial in $\nmatrix$, and $\ell \le p\paren{\frac{1}{2} - \delta}$,
then $\class{PH} = \class{AM} = \Sigma^P_2$.
To encode the puzzle,
we fix a prime $p \in [2^{\nmatrix-2}, 2^{\nmatrix-1}]$ and use the $\nmatrix^3$ bits of
$\rlz(\groundset_P)$ to sample $P = P(\rlz)$ (nearly) uniformly at random from
$\set{0,1,2, \ldots, p-1}^{\nmatrix \times \nmatrix}$ as follows.
For a matrix $P \in \integers^{\nmatrix \times \nmatrix}$, we let $\numberof(P) := \sum_{ij}P_{ij}\cdot
p^{(i-1)\nmatrix+(j-1)}$ define a base $p$ representation of $P$.
Note $\numberof(\cdot)$ is one-to-one for $\nmatrix \times \nmatrix$ matrices with entries in
$\integers_{p}$, so we can define its inverse
$\invnumberof(\cdot)$.
The encoding $P(\rlz)$ interprets the bits $\rlz(\groundset_{P})$ as
an integer $x$ in $[2^{\nmatrix^3}]$, and computes $y = x \mod (p^{\nmatrix^2})$.
If $x \le \floor{2^{\nmatrix^3}/p^{\nmatrix^2}} p^{\nmatrix^2}$, then $P =
\invnumberof(y)$.
Otherwise, $P$ is the all zero matrix.
This latter event occurs with probability at most $p^{\nmatrix^2}/2^{\nmatrix^3} \le
2^{-\nmatrix^2}$, and in this case we simply suppose the
algorithm under consideration finds the treasure and so gets unit reward.
This adds $2^{-\nmatrix^2}$ to its expected reward.
So let us assume from now on that $P$ is drawn uniformly at random.

Next we consider the maps.
Partition $\groundset_{M}=\biguplus_{i=1}^{\nmaps}M_{i}$ into $\nmaps$
maps $M_{i}$, each consisting of $\mapsize$ items.
For each map $M_{i}$, partition
its items into $\mapsize/\log_{2} \ntreasure$ groups of $\log_{2}
\ntreasure$ bits each, so that the bits of each group encode a
$\log_{2} \ntreasure$ bit binary string.
Let $v_{i} \in \set{0,1,\ldots,\ntreasure-1}$ be the XOR of these $\mapsize/\log_{2} \ntreasure$
binary strings, interpreted as an integer (using any fixed encoding).
We say
$M_{i}$ \emph{points to} $v_{i}$ as the location of the treasure.
A priori, each $v_{i}$ is uniformly distributed in $\{0,...,\ntreasure-1\}$. For a
particular realization of $\rlz(\groundset_{P}\cup\groundset_{M})$,
define $v(\rlz) := v_{i(P(\rlz))}$.  We set $v(\rlz)$ to be the location of the
treasure under realization $\rlz$, i.e., we label $E_{T} = \set{e_0,e_1,\ldots, e_{\ntreasure-1}}$ and ensure
$\rlz(e_{j}) = 1$ if $j = v_{i(P(\rlz))}$, and $\rlz(e) = 0$ for
all other $e \in E_{T}$.
Note the random variable $v = v(\rlz)$ is distributed
uniformly at random in $\{0,1,\dots,\ntreasure-1\}$.  Note that this still holds
if we condition on the realizations of any set of $\mapsize/\log_{2} \ntreasure - 1$
items in a map, because in this case there is still at least one group
whose bits remain completely unobserved.

Now consider the optimal policy with a budget of $k=\nmatrix^{3}+\mapsize+1$ items to pick.
Clearly, its reward can be at most $1$.  However, given a budget of $k$,
a computationally unconstrained policy can exhaustively sample
$\groundset_{P}$, solve the puzzle (i.e., compute $i(P)$), read the
correct map (i.e., exhaustively sample $M_{i(P)}$), decode the map
(i.e., compute $v=v_{i(P)}$), and get the treasure (i.e., pick
$e_{v}$) thereby obtaining a reward of one.

Now we give an upper bound on the expected reward $R$ of any randomized
polynomial time algorithm $\Alg$ with a budget of $\beta k$ items,
assuming $\Sigma^P_2 \neq
\class{PH}$.
Fix a small constant $\gamma > 0$, and set $\mapsize = \nmatrix^3$ and $\nmaps = \ntreasure = \nmatrix^{1/\gamma}$.
We suppose we give $\Alg$ the realizations $\rlz(\groundset_M)$ for free.
We also replace its budget of $\beta k$ items with a
budget of $\beta k$ specifically for map items in $E_{M}$ and
an additional budget of $\beta k$ specifically for the treasure locations in
$E_{T}$.
Obviously, this can only help it.
As noted, if it selects less than
$\mapsize/\log_{2} \ntreasure$ bits from the map $M_{i(P)}$ indicated by $P$, the
distribution over $v_{i(P)}$
conditioned on those realizations is still uniform.  Of course,
knowledge of $v_{i}$ for $i \neq i(P)$ is useless for getting reward.
Hence $\Alg$ can try at most
$\beta k \log_{2}(\ntreasure) / \mapsize = o(\beta k)$
maps in an attempt to find $M_{i(P)}$.
Note that if we have a randomized algorithm which given a random $P$ drawn from
$\set{0,1,2,  \ldots, p-1}^{\nmatrix \times \nmatrix}$ always outputs a set $S$ of
integers of size $\alpha$ such that $\prob{i(P) \in S} \ge q$, then we can use it to
construct a randomized algorithm that, given $P$, outputs an integer $x$ such
that $\prob{i(P) = x} \ge q/\alpha$, simply by running the first algorithm
and then selecting a random item of $S$.
If $\Alg$ does not find $M_{i(P)}$, the distribution on the treasure's location
is uniform given its knowledge.  Hence it's budget of
$\beta k$ treasure locations can only earn it expected reward at most $\beta k/\ntreasure$.
Armed with these observations and
Theorem~$1.9$ in the work of~\citet{FeigeL97} and our complexity theoretic
assumptions, we infer
$\expct{R} \le o(\beta k) \cdot  2^{-\ell}(1 + 1/\nmatrix^{\eta}) +
\beta k/\ntreasure + 2^{-\nmatrix^2}$.
Since $\mapsize = \nmatrix^3$ and $\nmaps = \ntreasure = \nmatrix^{1/\gamma}$ and $\gamma = \Theta(1)$
and $\eta = 1$ and $\ell = \log_{2} \nmaps$ and $k = \nmatrix^3 +
\mapsize + 1 = 2 \nmatrix^3 + 1$,
we have
$$\expct{R} \le \frac{\beta k}{\ntreasure}\paren{1 + o(1)} =
2\beta\nmatrix^{3 - 1/\gamma}(1 + o(1))\mbox{.}$$

\noindent Next note that $|\groundset|=\ntreasure+\nmaps \mapsize
+\nmatrix^{3} =  \nmatrix^{3+1/\gamma}(1 + o(1))$.
Straightforward algebra shows that in order to ensure
$\expct{R}=o( \beta/|\groundset|^{1-\varepsilon})$,
it suffices to choose $\gamma \le \varepsilon/6$. Thus, under our
complexity theoretic assumptions, any polynomial time randomized
algorithm $\Alg$ with budget $\beta k$ achieves at most
$o(\beta / |\groundset|^{1-\varepsilon})$ of the value obtained by the
optimal policy with budget $k$, so the approximation ratio is $\omega(|\groundset|^{1-\varepsilon}/\beta)$.
\end{proofof}

\end{document}